\documentclass[11pt]{amsbook}
\usepackage{geometry}                
\usepackage{graphicx}
\usepackage{amssymb}
\usepackage{epstopdf}
\usepackage{graphics}
\usepackage{hhline}
\usepackage{amsmath}
\usepackage{amsfonts}
\def\argmax{\mathop{\rm argmax}}
\newcommand{\abs}[1]{\lvert#1\rvert}

\begin{document}
\begin{titlepage}
{\center
{\Huge
The University of Sheffield\\[18pt]
}

{\LARGE
T. E. Dunning \\[18pt]
}
{\LARGE
Finding Structure in Text, Genome and Other Symbolic Sequences \\[18pt]
}
{\large
Doctor of Philosophy \\
1998 \\
}
}
\end{titlepage}
\title{Finding Structure in Text, Genome and Other Symbolic Sequences}
\author{Ted Emerson Dunning\\
Doctor of Philosophy\\
Department of Computer Science}
\date{Submitted May 1998}
\maketitle
\begin{titlepage}
\begin{center}
{\large Acknowledgments}
\end{center}

I would like to particularly thank my advisor, Yorick Wilks.  Over the
years I have known him, Yorick has always been insightful, cheerful
and generous.  His ability to understand and constructively critique
research over an extraordinarily wide range of topics has been
extremely helpful, both professionally and academically.  Moreover,
his willingness to consider and even nurture points of view which run
counter to his own intuitions is extraordinary in a world filled with
entirely too many ideologues.  Even more extraordinary is his
willingness to adopt techniques that have proven useful regardless of
his initial assessment of them.  This flexibility is the mark of a true
scientist.

The members of my committee, Peter Willett, Robert Gaizauskis and
Steve Renals, also provided valuable advice and assistance and
oversight.  Peter has been especially helpful by sharing his wide
knowledge of and historical perspective on the field of information
retrieval.  In addition, he has been willing to carefully edit this
work to graciously help me to completely avoid the use of split
infinitives.  Robert has spent more time helping me and examining this
work than could ever have been expected.  

Outside of the university, Ellen Friedman also provided enormous
support in the form of advice and careful review of this thesis.  Her
guidance was critical to the quality of the chapters on genomic
sequence analysis.  Her advice and insight throughout this work have
helped me make this thesis far better than it otherwise could have
been.

Jamie Callan reviewed the chapter on document routing and provided
helpful comments.  At least as important, however, he supervised the
preparation of the InRoute comparison results which provided a
connection between my results in in chapter \ref{ir} and the IR
literature in general. 

Owen White extracted and provided data for the genomic sequence
analysis chapters.  Only those who have worked with the sequence
databases can know how helpful this was.

My current employer, Aptex Software, and the parent company, HNC
Software, are to be commended for their encouragement and support
during the preparation of this thesis.  They have provided financial
support, access to computational resources and the Convectis system.
This support has been crucial to the completion of this thesis.

Finally, my examiners, Keith van Rijsbergen and Steve Renals, are
taking a substantial amount of their time to judge this work.  I thank
them.

\end{titlepage}
\begin{titlepage}
\begin{center}
{\large Summary}
\end{center}
The statistical methods derived and described in this thesis provide
new ways to elucidate the structural properties of text and other
symbolic sequences.  Generically, these methods allow detection of a
difference in the frequency of a single feature, the detection of a
difference between the frequencies of an ensemble of features and the
attribution of the source of a text.  These three abstract tasks
suffice to solve problems in a wide variety of settings.  Furthermore,
the techniques described in this thesis can be extended to provide a
wide range of additional tests beyond the ones described here.

A variety of applications for these methods are examined in detail.
These applications are drawn from the area of text analysis and
genetic sequence analysis.  The textually oriented tasks include
finding interesting collocations and cooccurent phrases, language
identification, and information retrieval.  The biologically oriented
tasks include species identification and the discovery of previously
unreported long range structure in genes.  In the applications
reported here where direct comparison is possible, the performance of
these new methods substantially exceeds the state of the art.

Overall, the methods described here provide new and effective ways to
analyse text and other symbolic sequences.  Their particular strength
is that they deal well with situations where relatively little data are
available.   Since these methods are abstract in nature, they can be
applied in novel situations with relative ease.

\end{titlepage}

\tableofcontents
\listoftables
\listoffigures
\part{Introduction}
\chapter{Overview}
\label{overview}

Statistical methods for studying the properties of structured symbolic
sequences, such as human language, depend on the observation of
empirical data and the construction of models from these observed
data.  Unfortunately, in many cases, the enormous number of parameters
in the models being constructed makes almost any practically
available amount of data appear to be profoundly inadequate in size.
This problem can be mitigated to some degree by a clever choice of the
structure of the model or by collecting enormously more data.  A
different solution is to develop a more sophisticated statistical
analysis which makes better use of the available observations.
Preliminary results indicate that such an approach is not only
feasible, but that it can be used effectively in very diverse
applications.  The derivation of such techniques is the object of my
research.

The areas of research where the techniques developed so far have
proven effective include machine translation, parsing, information
extraction, name finding, information retrieval and genetic sequence
analysis.  For some applications, exactly the same computer programs
can be used to perform analogous tasks on data from radically diverse
origins.  For example, the same program that can identify in which
language a short text excerpt was written can also identify the
species of organism from which a short sequence of nucleotides is
taken.  In both cases, these statistical techniques perform
considerably better than any other known method in terms of accuracy,
amount of data required for classification and the amount of training
data required.

This dissertation describes the methods used for these analyses as
well as the results of research into these areas.  It contains four
main parts: introduction, methods, applications and discussion.  

The introductory part of this dissertation provides a rationale for the
research as well as a general background section.  This general background
establishes a context for this work within the larger framework of
computational linguistics.  This context is built through reference to
major works in the field, accompanied by brief discussion.

The second part of the dissertation, entitled ``Methods'', includes a more
detailed discussion of previous works which directly bear on the research
reported here.  This detailed discussion describes methods that others
have used to address similar problems.  It is followed by a detailed
exposition of the methods I have developed for addressing these
problems.

These new methods provide the abstract foundation for the applications
described in the next part of this dissertation which is called
``Applications''.  The applications described in this part demonstrate
the wide applicability of the relatively small set of methods
introduced here.  Each chapter in this part includes detailed
background material which is specific to the application described
there and which augments the more general background presented
earlier.  Of the five applications described in the third part of this
dissertation, three involve the analysis of human language, and two
describe the analysis of genetic sequences.  That the same methods can
be used in such divergent areas is indication of their strength.

The final part contains a discussion of ramifications of the research
described here.  The successful application of identical techniques in
diverse systems raises the interesting question of whether these
techniques capture some universal ess\-ence of human and genetic
languages.  This observation is not the same as simply noting that
statistical methods are general in scope.  Indeed, the methods
described here known limitations.  Also touched on in the final part
of this dissertation are the very interesting connections with minimum
description length methods and empirical risk minimization.  These
connections provide a fundamental basis for understanding how and why
the methods described here work so well under difficult practical
constraints.  The work reported in this dissertation does not attempt
to answer all the questions mentioned here, but it cannot help but
raise them.

The most significant feature at present of the work described here is
that has substantial practical impact; these methods are already being
incorporated into major commercial software systems.

\chapter{Rationale}

As was mentioned in chapter \ref{overview}, one of the primary
difficulties in the statistical analysis of text or other symbolic
sequences is the development of methods for dealing with very rare
events.  Indeed, while it is often possible to do interesting initial
work without facing these problems head on, producing a working system
which can deal with more than a few examples almost invariably
requires dealing with the problem of rare events.  In this context,
rare events can be considered to be observations which occur or are
expected to occur only a few times or not at all in a typical
experiment.  In symbolic sequences such as text, for example, the low
overall frequency of most words means that without further
information, we would expect nearly every six word sequence would
never be seen.  This is manifestly not the case.\footnote{The most
common English word {\em the} occurs roughly one out of every ten
words and roughly one-third of running English text consists of words
which occur less than once per 50,000 words.  Assuming that four words
out of a typical six word sequence are no more common than ``the'' and
the other two are relatively rare, we would expect to see each six
word sequence only once every ten trillion words or about a thousand
lifetimes.  And yet, in real text, this author has seen the seven word
sequence ``Staff writer of the Wall Street Journal'' literally
hundreds of times.}

The necessity of finding techniques which perform well in the face of
rare events is practically a signature of statistical natural language
processing.  Many researchers in statistical natural language
processing consider the problem of rare events to be a central issue.
It is so central, in fact, that critics often adduce it in the form of
an objection which they claim to prove the impossibility of
statistical natural language processing.  This objection is usually
phrased as some variation of the statement that ``no matter how much
text you examine, you will always see new words'' (or new grammatical
structures, or new retrieval topics, or ...).  This objection,
when stated so baldly, is clearly naive since it is a tenet of all
computational linguistics that there are observable regularities in
language which allow a finite program to process the infinite variety
of language.  Indeed, it is a central tenet of all the sciences that a finite set of observations will suffice to induce the basic structure of an infinite universe.  This objection does, however, contain the germ
of one of the central problems in statistical language processing.
This central problem is that many important phenomena will be observed
only very rarely.  In a moderate-sized sample, many of these rare
events may not be observed at all.

In recent times, the critics raising the rare event problem have
sometimes been answered by pointing to the huge amounts of electronic
text which are becoming available.  In natural language processing,
literally billions of words of English text are now available for
analysis.  It might be thought that this glut of raw data would make
moot the problem of rare events.

The simple fact of the matter is, however, that the problem is not
going away.  Rare events cause trouble even with such an enormous sea
of data.  There are several causes for this difficulty.  For example,
many statistical models which have been applied to natural language
processing have {\em enormous} numbers of parameters to be estimated.
As these models are made more detailed, the number of parameters
typically goes up exponentially or even faster.  Indeed, it is often
the availability of training data more than anything else which limits
the maximum complexity that is feasible for these models.  Handling
rare events well can sometimes substantially extend the maximum
complexity of the models which can be used for a given amount of
training data.

Another source of the problem is that many systems cannot use raw text
directly, but must instead use training text which has been annotated
by a human judge.  Appropriate data such as hand tagged or parsed text
may not be readily available in many cases.  In such a situation,
generating sufficiently large quantities of training data may simply
be too expensive an undertaking.  Reducing the need for such large
data sets by handling rare events in some other fashion is clearly
desirable in these situations.  Exploring methods to achieve this goal
is one of the major motivations for my research.  The methods that I
have explored all make use of the log-likelihood ratio tests that are
described in this dissertation.

Furthermore, the range of expression of human language is so enormous
that it is impossible to imagine that any finite corpus could begin to
explore a significant fraction of the total range.  Handling rare or
even missing data well can provide hints for a statistical system to
indicate where generalization is likely to be applicable, or where it
may be erroneous or simply not useful.

A final source of difficulties with rare events addressed by this
dissertation is that human language is not a static entity.  It
continually evolves and changes.  When a new phenomenon arises, be it a
new word, or other form of expression, a natural language processing
system will need to handle it competently long before any substantial
amount of data displaying the new phenomenon has been collected.

Given the near inevitability that a practical statistical natural
language processing system will have to deal with rare events, it is
natural to ask what the impact might be if rare events are not handled
well.  Examples are not hard to find in which rare events cause
practical problems.  Speech recognition systems, statistical machine
translation systems, part of speech taggers and information retrieval
systems all need to deal well with rare events.  The first three of
these types of systems often incorporate language models with hundreds
of thousands to tens of millions of free parameters.  In the case of
speech recognition and statistical machine translation systems,
performance suffers when simpler language models are used, but when
the higher order models are used, many possible events may be assigned
zero probabilities unless considerable efforts are taken to deal with
rare events.  These zero probabilities can lead to nonsensical
results.  An example occurs when past observations are used and some
event is judged utterly impossible.  If the event is subsequently
observed the system is faced with the actuality of an
``impossibility''.

In the case of part of speech taggers, if missing data and rare events
are handled well, the required amount of training data can be
decreased by several orders of magnitude with little if any loss in
performance.

Information retrieval systems have traditionally suffered from these
effects although the cause has not often been well identified.  As
early as the mid-70's, Robertson and Sparck Jones had introduced term
weighting methods which implicitly dealt to some degree with these
problems.  More recently, some of the most effective prototype
document routing systems have finessed the problem by creating
enormous queries with hundreds of search terms.  Both of these
approaches were attempts to avoid the problem of dealing with small
numbers of observed events.  The latter problem of query size
explosion, in particular, can be handled better with methods which
deal well with rare events.  These and other approaches as well are
discussed in more detail in chapter \ref{old_methods}.






To some degree, the problems described so far can be handled by
collecting more data.  This approach is the obvious brute force
solution to the problem, and may well be the most appropriate in some
situations.  In others, it may simply not be practical. 
 
Another option might be to modify the model being used in such a way
that relatively independent phenomena can be handled separately.  For
instance, vast amounts of text might have to be analysed before all
forms of a rare verb are observed, because a system which always
considers each verb form completely independently from every other
form of the same verb will need to observe a number of instances of
each form before being able to deal with all of the forms.  If, on the
other hand, an analysis is done ahead of time to analyse verb forms
into their stem and suffix, then it may be possible for the system to
learn independently about the verb stem and about verb forms in
general.  This process of factoring independent phenomena is likely to
work well for very regular situations such as have been conjectured to
pertain to rare words.  For irregular, or highly idiomatic usages,
factoring is likely to work much less well, but these cases are likely
to be much more common.

Even better, however, is to use a system which can effectively handle
both situations.  For common forms, enough data can be gathered to
handle idiomatic usages, while for rare forms generalization from the
factored form of the model where stem and ending are considered
independently can be done.  With appropriate methods of analysis,
common and rare events can be handled uniformly and effectively.  The
overall result of using these methods well can be substantial
improvements in efficiency both in terms of the amount of training
text and computer resources required.

It is an extraordinary property of the methods which are described in
this dissertation that not only do they facilitate the processing of human
written and spoken language, but these methods also can be applied to
genetic sequences.  This is provocative not only because of the
interdisciplinary nature of this application, but also because the
most commonly used language by far is not English or Mandarin but is
instead the language of the genome.  Cells in every living organism on
this planet make use of this language, and yet we have only the barest
beginning of an understanding of how it works.

One major commonality between the processing of natural language and
the language of genetic sequence data is that the problem of rare
events appears in each field.  The commonality, however, goes much
deeper than mere shared agony.  Instead, many of the causes of the
problem are shared, and many of the methods for dealing with the
problem are applicable in both areas.  One example of several of the
phenomena common to both human language and genomes is the fact that
the observed structure in each tends to be positive structure.  In
English, for example, when two words appear near or next to each other
for some structural reason, they tend to appear much more often than
would be expected based on the isolated frequencies of the words.  The
same tends to be true of genomes where we must replace ``word'' by
``sub-sequence''.  This statistical property may be due to the fact
that there are very many more relatively rare words than there are
common words.  Thus, any particular collocation would be expected to
be quadratically rare so any systematic occurrence must tend to be
more common than expected.  Any statistical test which is good at
finding this kind of over-represented coincidence would, by nature, be
good for both text and genomes since they both exhibit this kind of
structure.  This trait of over-representation is not universal to all
sequences, so the fact that both kinds of sequence share this trait is
non-trivial.

Furthermore, genomes and natural language are very different in at
least one important aspect.  Genetic sequences make heavy use of
palindromes, while reversible sequences or words or letters seem to
play no part in natural language other than as curiousities.
Likewise, the duality between one strand of DNA and its complementary
strand has no corrolary in natural language.  In spite of these and
many other dramatic differences between genetic sequences and written
language, statistical tools which were originally devised to process
written language can be used to analyse genetic sequences very
effectively without any modifications whatsoever.

A potential criticism of methods which work well in describing
extraordinarily diverse phenomena is that they necessarily lack depth
in any single area.  It is my contention that the close mathematical
connection that exists between the methods described here and the
minimum description methods of Rissanen (and ultimately to Occam and
Aristotle) indicates that the methods described here are actually
capturing a deep essence of symbolic sequences.  The mathematical
basis for this claim is described in section \ref{methods:mdl}.  That
these new methods can capture this essence by highly tractable
computational means is one of their chief attractions.

In summary, the novel application of several statistical methods which
can help deal with rare events in processing natural language and
genetic sequences are described in this dissertation.  In addition,
these methods have successfully been applied to significant problems
and the results are described here.  These methods are straightforward
enough that they can also be applied to other problems or as
components of other methods.  In order to illustrate the wide scope of
these methods, the sample applications described here deal with both
written language as well as the chemical language of genomes.

The novelty in this work is the introduction of the techniques based
on generalized log-likelihood ratio tests to the analysis of symbolic
sequences such as human language and genetic sequences.  These methods
had never been applied to these applications or in these areas of
inquiry before.  Generalized log-likelihood ratios were studied by
Wilks in the early part of the 20th century, but they had been applied
to neither natural language processing nor to genomic sequence
analysis before the work described here.  The work presented in this
dissertation demonstrates that these techniques can be very effective
and yet are simple to implement and are computationally very
efficient.  These methods represent a significant contribution to the
field based on their novelty and the range of their application and
effectiveness.
\chapter{General Background}

\section{Overview}

Linguistics as a science has an inherent problem shared only by a few
other fields such as psychology and philosophy in that language is a
phenomenon which is internal to the minds of the researchers examining
it.  Cognitive psychologists have long recognized that such a
situation can lead to introspective analysis which appears valid, but
which is falsified by careful experiment.  Human language is one of
the most complex aspects of human cognition and there is little reason
to suspect that the analysis of human language should be more
susceptible to introspection than other, apparently simpler cognitive
phenomena have proven to be.

One way to do experiments on language would be to follow the example
of cognitive psychology and give some large number of people a
linguistic task and measure their performance on the task.  Each
subject might be presented with a sentence and asked to press a button
according to their judgement of the sentence based on a specified
criterion.  Other tasks, both simpler and more complex, can easily be
constructed, but the effort required to get reliable and repeatable
results from such an experiment can be substantial.  In spite of the
difficulty, important results have been achieved by these direct
experimental methods and experiments very much like the sentence
judgement task just mentioned have actually been done \cite{schvan76,vanDijk/Kintsch:1978, vanDijk/Kintsch:1983}.  

Unfortunately for the cognitive psychologists, the general population
does not go to the trouble of recording their reaction times to
stimuli in their everyday life.  They do, however, emit language
nearly continuously, and a very large proportion of the population do
us the courtesy of spending a large fraction of their lives recording
this language as machine readable text.  This text can be used to
address questions about the basic structure of language.

Since these large bodies of text are produced without reference to the
experimenter, they inherently avoid the problems of introspective
analysis.  Newspapers, books, recorded speech can all serve as our
experimental corpus and there are enormous numbers of questions about
language which can be answered using such a corpus.  For example, once
such a corpus becomes the focus of research, it becomes natural to
start talking about the frequency of words, or whether a particular
form of expression is common or rare.

The idea of applying statistical techniques to language is not a
particularly novel one.  There has been, however, a sort of conspiracy
of circumstances which has until recently made statistical analysis
exceedingly difficult to apply profitably to natural languages.  One
major limitation has been the lack of real data in the form of large
bodies of machine readable text with which to work.  Another
limitation has been the lack of computers powerful enough to do the
necessary analysis.  Over the last three decades, the size of
available corpora has increased by three or more orders of magnitude
and the price/performance ratio of computers has improved by nearly
six orders of magnitude.  Together, these factors alone have allowed
qualitative differences in the sorts of statistical analysis which are
practical.

One simple example of this progress can be had by projecting what
Kucera and Francis would have had to do to simply count the words in a
corpus with $10^9$ words in it instead of the $10^6$ word corpus that
they actually analysed \cite{kucera-francis-67}.  Instead of the 25
foot stack of computer cards that they used, they would have needed a
5 mile deep stack of cards.  Reading these cards into a computer with
a high speed card reader would have literally taken months of time.

Just as important as the progress in terms of machine speed, cost and
corpus size have been, but less obvious in the linguistic community is
the fact that the body of statistical techniques has itself been
dramatically extended during this same period.  The availability of
inexpensive computational power has fundamentally changed the sorts of
statistical analyses which can be used.  Today, more computation can
be applied to an undergraduate homework problem than could be applied
to virtually any statistical problem in the 1920's.  This has
fundamentally changed what sorts of statistical algorithms are
practical and has thus changed the sorts of statistical algorithms
which have been the subject of research.  The overall result has been
a fundamental change in the character of statistical analysis which
has been as much of a qualitative difference as the change in the
availability of data and computer power.  It is easy to dismiss the
magnitude of these changes, but they have been absolutely fundamental.
Examples of new techniques which were developed during this period
include bootstrap estimates, minimum description length techniques
\cite{rissanen, wallace}, the EM algorithm \cite{em_alg}, structural
risk minimization \cite{vapnik-srm}, regularization techniques for
ill-posed statistical problems \cite{vapnik} and many other
developments.

Together the three factors of machine performance, corpus size and new
statistical techniques have made approaches to natural language which
were completely impractical three or four decades ago highly feasible
today.  For example, it was once required months of effort and
considerable organizational ingenuity simply to count the words used
by Shakespeare.  This exercise can now be done in less than a minute
and is suitable as a casual assignment for undergraduates.  It is not
unusual to find that variations on approaches which were suggested and
discarded decades ago as impractical are now highly profitable avenues
for research.

\section[History]{A short (and biased) history of statistical language processing}

The threads of development which have led to the current state in
natural language processing by statistical means can be classified
into a few distinct areas.  The very early work by Zipf \cite{Zipf49}
and others before him \cite{Markov:1913} was distinctive by its
descriptive nature and lack of practical applications.  Much of this
lack can easily be attributed to the utter impracticality of applying
statistical methods to text on a wide scale using manual methods.
Another thread of development includes researchers in the Speech
Recognition Group at IBM \cite{jelinek} and the numerous researchers
in text retrieval leading to the present \cite{Salton91}.  This work
on the problem of information retrieval ultimately has produced a wide
range of highly practical systems which are now seeing wide usage, but
this practical focus has largely precluded the use of anything much
more advanced than counting words and accumulating scores based on
those counts.

One fundamental difference between the text retrieval efforts starting
in the late 50's and the research into methods for speech recognition
is that the speech recognition work has required a considerably more
focus on mathematical methods since raw speech input is, of necessity,
something that requires considerable signal processing.  With text,
especially in western European languages such as English, essentially
all of this signal processing is done by the person who transcribes
the text.  The requirement for fundamentally numerical computation
even to begin work on recognizing speech has meant that researchers in
the area are much more comfortable with mathematical formalisms than
with linguistic theories.  The result of this difference in basic
focus ultimately has been systems which use numerical methods to
process language in non-trivial ways.

The work on the mathematical properties of random processes relative
to model formation has proceeded over a period of centuries with roots
as deep as Western philosophy.  Over the last few decades, this work
has been particularly fruitful and the results have seen rapid
reduction to practice by the mathematically and statistically oriented
speech recognition community.  Practical application of the
mathematical machinery to other problems is also possible, however,
and many areas of natural language processing are particularly ripe
targets for these methods.

\subsection{Zipf and before}

Numerical approaches to language are hardly new.  As early as the
mid-nin\-eteenth century, the mathematician de Morgan had suggested that
numerical techniques could be used to distinguish authors.  His
suggested technique of looking at the distribution of word lengths was
taken up by the American physicist Mendenhall who tried without
success to resolve whether Bacon was the actual author of the words
attributed to Shakespeare.  These efforts had virtually no effect on
the modern state of the art in statistical natural language
processing, however.

Of much greater import was the work by Markov who mathematically
analysed text as a sequence of letters \cite{Markov:1913}.  The use of
Markov processes is now a fixture in much of statistical text
analysis, but it is often forgotten that Markov not only invented the
necessary mathematical machinery, but he actually used text as a
fundamental example of how his methods could be used.

\begin{figure}[htb]
\begin{center}
\includegraphics[]{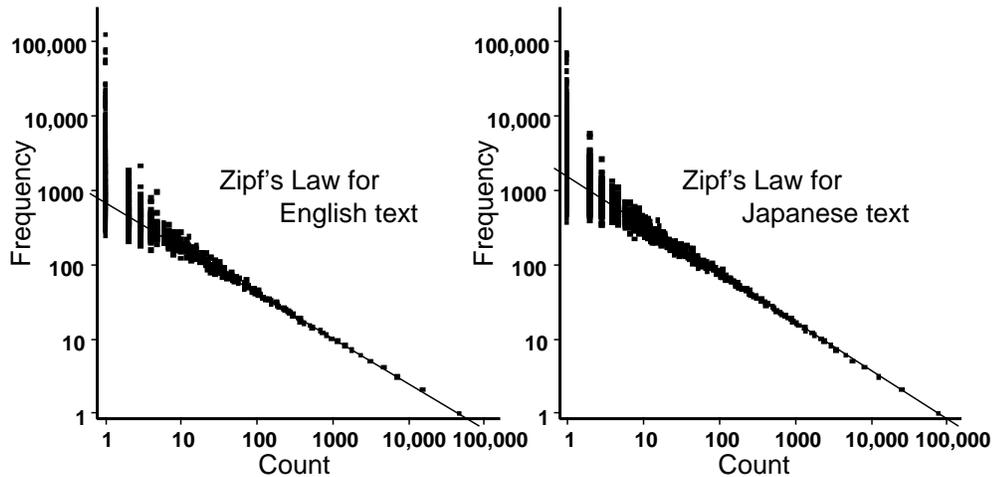}
\caption[Zipf's ``Law'' relates the frequency of a word and the word count]{Zipf's ``law''
captures the extraordinary log-linearity of the relationship between
the frequency of a word and the number of words with exactly the same
frequency as that word.  For instance, there are many words which
occur just once, and all of these words are represented by a single
dot in the lower right part of the graph.  On the other hand, with
words with very high frequency, typically only a few have the same
frequency.  The overlaid lines have very slightly different slope, but
the similarity of the two curves is striking. Very similar plots are
produced when the frequency of each word is plotted against the rank
of the word.  The rank of a word is computed by ordering words by
frequency so that the most common word has rank 1 and all of the words
which have frequency 1 have the highest rank (that is, appear last in
the rank-ordered list).  \label{fig:zipf}}
\end{center}
\end{figure}

Much more widely known by name is Zipf's work on the distributional
properties of language.  Zipf originally claimed that there were at
least three properties of word frequencies in text, the most prominent
of which was a relationship between the frequency of a word and the
number of words with exactly that frequency.  Ultimately, this
relationship was rephrased to claim that the frequency of the $n$th
most frequent word was proportional to $1/n$ \cite{Knuth_v3}.  In many
ways, this relationship is one of the most striking and universal to
be found in language to date.  For instance, if we plot the frequency
of words versus the number of words with the same frequency for
English and Japanese corpora, we get the results shown in Figure
\ref{fig:zipf}.  The similarity between the two graphs and their
linearity is astounding to see in any experiment which analyses a
human cognitive activity.  It should be noted that these plots
demonstrate Zipf's original law \cite{Zipf49}, not the rephrased
version which involved plotting frequency versus rank.

Unfortunately, Zipf's observation has proved to be relatively sterile
in terms of linguistic insight.  The original claims that there was a
universal property of least effort represented in Zipf's law have
largely remained unfulfilled and have been relegated to obscurity.
Zipf's law has been used to advantage in the design of algorithms to
look up words \cite{knuth_hash}, but little has come from it in terms of
the understanding of language.  Part of the problem has undoubtedly
been that the predictions made by Zipf's law are so general and
non-specific that it is almost impossible to use them for any particular
application.

\subsection{Text Retrieval}

The idea of applying statistical methods to natural language
processing was not completely dormant during the 1950's.  Hans Peter
Luhn was ahead of his time as witnessed by his work on text retrieval
\cite{Luhn:1957:SAM} and automatic document summarization
\cite{Luhn:1958:ACL}.  He also was the apparent inventor of
hashing \cite{knuth_hash} and KWIC indexing methods.  Most of this
work appears to be a collection of {\em ad hoc} methods by present
standards, but the idea that numerical techniques could be applied to
text is key to all of this work and is relatively novel for the time
although Fairthorne reported work along similar lines not long after
\cite{fairthorne}.  Numerical approaches had already been used in the study of
language by this time, but the goal had typically been the analysis of
observational data about the relative frequency of various phenomena.
Luhn was not even the first to recommend that numerical techniques
such as applying weights to words or to stylistic features.  The
American physicist Mendenhall had recommended something similar during
the nineteenth century for distinguishing authorship.  Luhn's key
contribution that he was among the first people who had the
opportunity to apply sufficient computational resources to implement
large scale numerical techniques for language processing.  This
opportunity allowed Luhn to gain real experience working with real
problems; for the first time, recommendations could be based on actual
results rather than conjecture.

Not so much later Sparck Jones \cite{ksj64} found herself in a
similarly fortuitous situation.  She was able to bring to bear newly
available computational power with newly developed clustering
algorithms on the problem of word context analysis.  Although unable
to fully implement her system and although she used highly
idiosyncratic text, she was able to show that computers could indeed
be used to provide useful analyses of the coocurrence patterns of
words and to show that there were indeed useful connections between
similarity of the contexts in which words were used and the meanings
of the words that could be exploited using computers.

The idea that symbols which are fully equivalent can be substituted
freely for each other was not new with the work of Sparck Jones; it is
a basic tenet of very long standing in mathematics.  An early example
of the use of this is Euclid's axiom that if two equal elements are
equal to a third, then they are equal to each other \cite{euclid}.  It
is clear that the Greeks considered this axiom in a very general sense
rather than in the specific sense limited to geometric figures or
numbers.  Indeed Appollonius' ``proof'' of this axiom by the example of
three lines which occupy the same space was not generally adopted
presumably because it lacked the generality of the statement as given
by Euclid.  

Substitution of equivalents is also a fundamental process in algebra.
If $x=y$, then in any formula containing $y$, $x$ can be substituted.
In mathematical logic, a symbol can be defined to be equivalent to
some logical form and then can be used interchangeably with that form.
Frege, for example, uses an unstated rule of inference based on such
substitution, but does not even bother to carefully define when how it
can be used \cite{frege}.  Indeed, the idea that there might be any
controversy about substitution as a form of inference did not emerge
until nearly the 20th century.  From such examples of the use of
equivalence, it is not unreasonable to infer analogically that nearly
equivalent words in human language ought to appear in similar
contexts, and when they do appear similarly, that the meanings of the
expressions in which they appear are changed at most slightly.  It is
also quite reasonable to examine the contrapositive induction which
starts with the observation of two words appearing only in similar
contexts and in similar proportion and which then leads to the
presumption that the words must have similar meaning.

This logical reversal was not novel in the mid 1960's when Sparck
Jones did her work.  Quine even showed by famous {\em gavgai} example
that the reversed leap of logic was fallacious \cite{quine}.
Regardless, however, of whether the contrapositive followed logically,
there is still the {\em empirical} question of whether or not it might
just hold {\em practically} or {\em approximately}.  There was also
the further question of whether computers could be programmed to make
practical use of the putative relationship.  Sparck Jones examined
this latter question using empirical means and claimed to find support
for the conjecture that context similarity often indicates synonymy.
Even though the question of the connection between context of words
and their meaning had been around for some time, Sparck Jones' work
was essentially the first which dealt with the problem using a
computerized analysis of a corpus of text.  It is unfortunate that her
work is so little recognized outside the community in which it
originated since this use of similar context as a surrogate for
similar meaning is so thoroughly established in contemporary work.
Even without wide recognition, Sparck Jones' work stands as a
significant step towards the methods used in more recent efforts at
computerized analysis of text.

Subsequent to experimenting with the nature of synonymy, Sparck Jones
made substantial contributions to the mainstream of text retrieval
efforts by being one of the first to quantify the advantages of term
weighting in text retrieval \cite{robertson}.  Her work was
facilitated by the emergence of corpora intended specifically for the
evaluation of information retrieval methods \cite{SPARCKJONES76}.  The
Cranfield collection developed by Cleverdon \cite{cleverdon} was
prominent among the collections of the time.  These smaller test
collections have been largely superceded by the TREC corpus
\cite{harman95}.

The dominant figure in the development of so-called vector retrieval
systems was Gerard Salton, who was the driving force behind the SMART
system from the 60's until the mid 90's \cite{CORNELLCS//TR69-36,
Salton91}.  Virtually all of the successful text retrieval systems
during this time were, however, based on extremely shallow analysis of
the documents being retrieved and very simple combination of weights
applied to individual retrieval terms.  Efforts at more sophisticated
statistical analysis of the patterns of term occurrence led to what
was called probabilistic retrieval.  In practice, systems which claim
to be probabilistic are subject to serious resource limitations which,
in turn, places bounds on the sophistication of the method which can
be applied.  This means that most ``probabilistic'' systems use linear
weighting of terms in documents \cite{callan_private}.  Furthermore,
it has been shown that the weights used in the so-called vector space
methods are reasonably good approximations for the weights which would
be used by probabilistic systems \cite{prob_weights}.  The practical
result is that systems based on vector space methods and those based
on probabilistic methods often have nearly identical implementations.

Probabilistic methods are perhaps most applicable when evidence is
available about the relevance of sample documents.  Robertson and
Sparck Jones dealt with this problem early on \cite{robertson}.  A
conventional retrieval system which is enhanced by allowing the system
to use relevance judgements made by the user is said to use relevance
feedback.  The extreme case of relevance feedback is document
routing.  In document routing, a large number of relevance judgements
are assumed to be available.  Indeed, there may not even be a good
statement of what makes a document relevant other than the user
judgements.  The system described in chapter \ref{ir} is a document
routing system.

There was early work by researchers at the National Bureau of
Standards in the United States on building automated thesauri for
query expansion based on various associational measures
\cite{association}, followed on by Sparck Jones \cite{ksj71}, but this
work was hampered by many factors, not least of which was the scale of
the problem relative to the computational resources available at the
time.  Furthermore, although Robertson and Sparck Jones were clearly
aware in their work on term weighting \cite{robertson} of the problems
caused by small numbers of observations, for the most part, methods
were not widely available to deal with these problems in a principled
and comprehensive manner.  The methods available at until recently
tended either to smooth too much and thus not make good use of
information available in the data or they gave too much credence to
coincidental happenstance.  

As an example of over-smoothing, Robertson and Sparck Jones added
$1/2$ to all counts to smooth the resulting probabilities.  This is
equivalent to using a uniform Dirichlet prior with $\alpha=0.5$.  An
early version of the the language identification system described in
Chapter \ref{lingdet} used a similar approach to smoothing to avoid
brittleness, but the results were too poor to report.  At the other
end of the smoothing spectrum is the work of Buckley \cite{buckley95}
in which words appearing in relevant documents are added to routing
queries in descending frequency order.  If enough terms are added in
order to include some important, but relatively rare, terms, then far
too many terms are added, which hurts performance.  As a result,
Buckley et al. had to limit the number of terms considered quite
severely.  Chapter \ref{ir} provides an alternative which avoids both
over-smoothing as well as over-expansion.




At the present time, work in information retrieval has several basic
fronts.  There are a number of researchers who continue to develop new
weighting schemes, phrase finders and query expansion techniques
\cite{harman95}.  
As well, there are those who work on building highly efficient search
engines which combine the ability to work with very large, dynamic
collections with modern term weighted retrieval in order to search the
World Wide Web \cite{chang, cutting, Lesk97}.  The Web has not only brought
modern retrieval systems to popular cognizance, but it has also
fostered the development of collaborative filtering schemes.  In these
schemes the traditional paradigm of a single user system is
dramatically altered to one where there is a potentially very large
community of users with related interests.  Feedback from one user can
be applied to improve the ability of the system to find information
for another user.  A classic work in this area was the work done by
Patti Maes and others at the MIT Media Lab
\cite{maes_chi95}.   

Other lines of development with origins in the early days of retrieval
are also being followed.  For instance, in spite of its essentially
independent development, the Yahoo directory is essentially an
embodiment of cluster-based retrieval
\cite{Article:73:Rijsbergen:ClustHyp}.  Such a system hypothesizes
that a high quality clustering of documents can facilitate retrieval
because if any document in a cluster is relevant, then other members
of the cluster are likely to be relevant as well.  The novelty of
Yahoo is that when a system has tens of millions of users, clustering
can be done by human intellectual effort.  This is particularly true
when many of the users have a personal stake in augmenting the system
with information about their particular area of expertise.  Without
the free labor that users of the Web expend on populating Yahoo's
directory, it would be impossible to maintain such a comprehensive
directory to given the pragmatic constraints of running a business.
This change in paradigm to a largely user-supported system mirrors in
some important aspects the design of collaborative filtering systems.
Both recognize that the web is not so much millions of computers each
with a single user but rather is more like a single composite computer
with millions of users.

\subsection{Speech Recognition}

The speech recognition community has contributed strongly to the
technology available for statistical text processing.  This has been a
natural offshoot of the techniques which have proven successful for
automatically recognizing speech.  Very early techniques attempted to
use forms of acoustic template matching to recognize individual
words.  Unfortunately, these techniques, even with the use of dynamic
time warping and other aggressive data normalization techniques, appear
to be unable to provide low error rate recognition, especially in the
case of connected speech.

A fundamental advance in voice recognition came when statistical
language models were incorporated into speech recognition systems.
This work followed in the footsteps of Hockett's introduction of
information theoretic techniques to linguistic problems
\cite{hockett}.  Subsequently, Damerau provided a key advance by
demonstrating how Markov models for language were not, in fact,
subject to the theoretic limitations as claimed by Chomsky
\cite{damerau}.  It is not clear if later researchers in the speech
recognition community drew directly from Damerau's demonstration, but
the methods that ultimately provided key advances were quite similar
to those he demonstrated.  Damerau appears to have subsequently worked
with members of the speech recognition group at IBM, so it seems
reasonable to presume that he had an influence on their later work.

As early as the early 1970's, the potential contribution of the
techniques based on noisy channel decoding techniques was recognized.
A highly coun\-ter-intuitive finding from this period was that purely
statistically based systems consistently and significantly
outperformed systems which incorporated components based on
traditional linguistically inspired techniques \cite{Jelinek:1975:DLS}.
This pragmatic superiority of statistical methods paralleled the same
observation in the information retrieval community \cite{Salton91}.
The success of statistical techniques inspired considerable additional
development in techniques for statistically analysing human language.
Perhaps as important, the intellectual heritage of the signal
processing aspect of speech recognition made the application of
information theoretic techniques to text almost inevitable.

During the 1980's, this confluence resulted in a number of
applications which were designed to process text without reference to
speech per se.  For instance, researchers in the IBM speech
recognition group developed statistical methods for part-of-speech
tagging \cite{bahl1976}.  Variations on these methods were later
published by Ken Church \cite{church_tag}.  In addition, methods for
aligning textual translations were also developed at IBM and later
popularized and adapted by Gale and Church \cite{gale93}.  These
systems were developed at IBM as components of a prototype translation
system which originally used purely statistical methods to perform
translation from French into English at nearly the level of other more
traditional systems which had seen man-millenia of development effort
\cite{stat_mt}.  Ultimately, the researchers at IBM were able to
develop methods to incorporate several aspects of the conventional
machinery from the natural language processing such as morphological
reduction and even some basic syntactic methods.  In a sense, the
methods used by the IBM group implemented the very early statement of
Gilbert King regarding the role of probability in a translation
system, ``to guess at a sequence of word which constitute the best
estimate of the meaning of the sentence in the foreign language''.
The methods used by the IBM group go far beyond what King was able to
articulate.  King's reported work \cite{king} described nothing like
the data-directed approach that the IBM system used.  Instead, King
described how presumably hand-crafted rules could probabilistically
select alternative readings for individual words.  Later experience
has shown that this is only a very small part of the entire problem.

Substantial developments which have stemmed from the IBM research
effort include the development of decision tree based parsing systems
\cite{magerman94, magerman_acl} as well as trigger-based Markov
language models \cite{Black+al:1992, CMU//CS-94-188}.  While the
translation systems built by this group never did outperform systems
built using more traditional methods, the overall impact of this work
on the field of corpus linguistics has been substantial.  Not only did
many of the innovations pioneered by the IBM work have a strong impact
on what was possible for others to do, but the use of statistical
methods and parallel corpora are now a staple even in very traditional
machine translation development efforts.

The primary techniques which were pioneered by the IBM group include
the application of held out smoothing to language modeling (see {\em
smoothing} in the glossary), the wide-spread use of Markov language
models, and the extensive use of mutual information and relative
entropy as optimization criteria.  Many of these techniques were known
before this work, but the successful application of these techniques
to text processing was novel work by the IBM group.  As described in
chapter \ref{new_methods}, there are close and deep mathematical ties
between many of these methods and the work described in this thesis.

\subsection{Algorithmic Information Theory}

The origins of algorithmic information theory go back somewhat further
than the origins of information retrieval.  For instance, Aristotle
argues at least twice \cite{aristotle1, aristotle2} that when the
consequences of two lines of reasoning are the same, then the more
limited antecedent is preferable.  Specifically, Aristotle considered
two theories of the nature of matter, one of which presupposed a small
number of composable qualities and the other which presupposed an
infinitude of qualities.  He stated without further argument that the
explanation with the smaller number of qualities was preferable.  This
sort of intuitive preference for economy is the fundamental basis of
what has grown into algorithmic information theory and the school of
statistical inference based variously on the principle of Minimum
Message Length (MML) or Minimum Description Length (MDL).

Slightly closer to the present is the often misquoted \cite{occam}
Franciscan William of Ockham who apparently did say {\em ``pluralitas
non est ponenda sine necessitas}'' (a plurality [of entities] should
not be posited without necessity).  This preference for the simple and
spare is entirely consistent with the Franciscan philosophy and was
not original with William.  It was, in fact, a rather widely used
philosophical principle at the time especially among those who aspired
to the parsimony of Saint Francis.

The early observations of Ockham and others have been extremely
important and have had substantial impact on science.  Unfortunately
without some precise way to measure the relative complexity of
alternative theories, these principles can only stand as philosophical
guides rather than precise computational tools.  As a philosophical
principle, Ockham's razor can be quite sharp, but in practical
application it dulls quickly.

The beginnings of the mathematical machinery needed to frame Occam's
principle more precisely came much later from the work in statistical
mechanics by Boltzmann, Gibbs and others.  This work introduced the
concept of entropy in the context of thermodynamics as a measure of
the disorder in a physical system.  Nearly a century later, Shannon
and Weaver \cite{shannon49} introduced the equivalence between entropy
and information and showed that entropy was the only function (to
within a constant factor) which met a set of key intuitive
requirements for a definition of information in communication systems.
Entropy was also shown to give a lower bound for the minimum size for
an encoding of a sequence of symbols, given some reasonable
statistical assumptions.  Shannon's efforts led to an enormous body of
work within the context of signal processing, communications and
coding theory, but it neglected the potential link from
entropy to computational systems.

This link from Shannon's entropy to computation and algorithms was
made first by Kol\-mo\-go\-rov \cite{Kolmogorov65}.  This work was
seminal in that it showed there was an absolutely fundamental
connection between the concept of entropy as formulated by Gibbs and
the concept of universal computing machines as formulated by Church,
Turing, Markov \cite{Turing37,Markov:47} and others.  Moreover,
Kolmogorov's concept of complexity is provably universal.  Where the
limitations imposed by Shannon's framework apply, Kolmogorov's
complexity is also equivalent to Shannon's information or to
topologically defined measures such as metric entropy \cite{vitanyi,
vapnik}.  In terms of actually using the definitions proposed by
Kolmogorov, however, there were still some difficulties.  Minor issues
such as maximization over all possible partitions of the real numbers
and over all possible universal computing systems stand in the way of
practical application.  In the realm of theoretical investigation of
the properties of computation, these difficulties are of less
importance, and Chaitin was able to build algorithmic information
theory into a powerful and concise tool for proving mathematical
theorems and investigating what the concept of random might mean
\cite{Chaitin82,Chaitin87}.  In spite of the philosophical attraction
of this work, however, real applications were still elusive.

Mathematical statisticians such as Akaike \cite{akaike73} and Rissanen
\cite{rissanen} were able to fashion algorithmic information theory into
a method for statistical inference.  Others, such as Wallace and
Boulton \cite{wallace}, combined Bayesian methods with Shannon's
entropy to come to similar conclusions.  Initially, these methods had
the form of heuristic corrections to statistical tests based on the
maximum likelihood principle \cite{Fisher:1922}, but these soon gave
way to statistical tests which were based directly on the number of
bits required to describe the outcome of an experiment in terms of a
probabilistic model.  In Wallace's work, this optimization was taken
as the maximization of posterior likelihood in a Bayesian analogy with
Fisher's maximum likelihood principle.  Rissanen's work and the work
inspired by it ultimately made heavy use of the Levin ``universal''
prior \cite{vitanyi} which provided a very direct connection back to
Kolmogorov's work.  The results based on Wallace's work owed more to
Fisher and Bayes and were ultimately more conventional and perhaps
more approachable.

Simultaneously with the development of MDL-based statistical analysis,
the asymptotic properties of nested models and log-likelihood ratio
test statistics were being described \cite{Chernoff1954,mood74}.
These tests are the primary basis for the work reported here, but
their close connection to MDL-based methods provides much of their
philosophical interest.  This connection is explored in chapter
\ref{new_methods}.  The fact that these tests are simple to implement
and can be computed very efficiently makes them very interesting from
the standpoint of practicality.  The use of these tests in the context
of computational linguistics was first described in \cite{dunning93}.
The work presented here is much more significant in that it describes
how the general technique of log-likelihood ratio tests can be used in
a wide variety of applications.  These methods can also be used to
develop a number of additional new applications.

More recently, there has been a spate of work which has attempted to
apply MDL principles to the analysis of language.  Prominent in this
recent work has been Carl de Marcken's work \cite{demarcken} in which
it was shown that segmenters for text and speech signals could be
built based directly on the MDL principle.  De Marcken also did some
preliminary work on inducing grammars for language based on these
principles.  Others have taken up this work \cite{french_stuff}, but
it is not yet clear where these methods will lead.

\part{Methods}

\chapter[Previous Methods]{Previous Methods for Dealing with Small Counts \label{old_methods}}

This chapter describes a variety of statistical techniques for
modeling symbolic sequences such as natural language that provide a
context and basis for the new methods described in chapter
\ref{new_methods}.  A common thread for all of these methods is that
they are intended to model sequences of symbols without making
appealing to traditional syntactic structural analysis.  Since these
methods generally deal best with short range structure and often have
no explicit mechanism to express the long range structure found in
text, they are sometimes described as working at a level of analysis
below the level of grammatical structure.  In fact, though, such a
comparison is somewhat inappropriate since the information captured by
statistical methods is due to structure which cannot be isolated to a
single level given the choices of pragmatic, semantic, syntactic,
syntagmatic or lexical levels.  Since so much of the linguistic
literature has focussed on grammar as the essential property that
makes human language actually be language, it is understandable that
modeling language without having identifiable and separable mechanisms
for handling syntactic, semantic and lexical aspects of language might
seem misguided.  Indeed, the very concept of the probability of a
sentence can appear nonsensical from the perspective of traditional
linguistics
\cite{teeter}.

Restricting the level of analysis is, however, not as onerous as might
be thought.  Practical and useful systems can be built using these
techniques.  Indeed, these methods allow systems to be built for
which no other feasible solution has ever been demonstrated.  For
instance, most information retrieval engines, as well as most speech
recognition systems, fundamentally rely on elaborations of the
statistical techniques described here.

Machine translation systems which perform at a non-trivial level of
quality can be built using essentially these same techniques.  Indeed,
these ``low-level'' techniques can be used to create decision tree
based parsing systems which are among the best parsing systems ever
created.  It is a mild paradox that methods which appear at first
fundamentally to lack the necessary power to describe the syntactic
structure of language are among the most effective known methods for
creating systems to analyse this syntactic structure.

This chapter starts by introducing the general notion of probability
models to describe symbolic sequences and then progresses through a
catalog of statistical model types of increasing complexity.  The
problem of parameter estimation and model approximation is then
addressed followed by a section describing statistical tests and
measures which have been used in analysing symbolic sequences, or
which have been described in the statistical literature.

This chapter presents the state of the art that led to the novel
developments described in this thesis.  Chapter \ref{new_methods}
describes the theoretical contributions I have made in terms of
applying log-likelihood ratio tests to the problems which arise in the
analysis of symbolic sequences.  Subsequent chapters describe, in
detail, how these tests can be applied to real-world problems to
achieve results.  The work described in these chapters represents the
first application of these tests to problems of this nature.

\section{Models for Text and Other Sequences}

In the context of statistical analysis, there is conventionally an
assumption that probabilities can be assigned to empirically observed
events.  This assumption is one which is completely non-controversial
in a field such as physics, but which can generate heated discussion
when applied to linguistics.  Part of the problem is, no doubt, a
collision of traditions, but a more fundamental problem has to do with
the confusion that inevitably occurs in linguistics between the
observer and the experiment.  Since language is not only the object of
study but is also the means of reporting results, it is very easy for
a confusing meta-circularity to appear.  It is as if a physicist could
conjure up a new particle to analyse with a simple utterance.  If such
were possible, the concept of the probability of a particle appearing
in the world at large would be difficult to reconcile with the
probability of a particle appearing as the result of a scientific
publication.

Interestingly, the other kind of language that humans use
ubiquitously, i.e. genetic language, is not subject to this sort of
circularity.  Sequences of human nucleotides are sequences of
nucleotides because they occur in human cells, not because they occur
in scientific articles.  There is a clear distinction between genetic
sequences and text describing genetic sequences.

In order to avoid this confusion, it is customary in the statistical
analysis of language to emulate this division by performing
statistical tests based on a corpus of text which was produced without
any possible influence from the person doing the analysis.  This
corpus analysis allows models to be constructed which accurately
reflect the distributional properties of the original corpus, and
which, ideally, can predict the properties of previously
unseen material.

This prediction is done using functions which map events to
probabilities.  In order to make statistical analysis feasible, these
functions are not completely general but rather must meet a number of
abstract constraints, the most familiar of which is that the sum of
the probabilities of all possible events must be one.  In order to
make mathematical analysis more convenient, it is typical to restrict
one's attention to families of probability functions.  These families
of functions have desirable mathematical properties.  The selection of a
particular probability function from the family of functions is done
by choosing specific values of the parameters of the model. 

These probabilistic models therefore are functions with two kinds of
arguments, the parameters and the event (which is a specific value of
what is called a random variable).  To make this distinction clear in
this dissertation, I write the general form for a family of
probability functions as $p( \theta ; x )$ where $\theta$ represents
the parameters and $x$ represents the event.  The parameters can be
manipulated to try to make the model fit the real world, while the
event is some observation from the real world.  An alternative
notation which is used elsewhere in this dissertation is $p(x \mid
\theta)$.  Expressing the dependence of the distribution of $x$ on the
parameters $\theta$ as a conditional probability can be used to
highlight situations where a Bayesian formulation is being used and
the parameters themselves are being considered a random variable.

Generally, the value of $\theta$ is constrained to be a member of a
set $\Omega$ to ensure that the resulting function $p(\theta ; x)$ is
a valid probability density function.  Generally, these constraints
are that $\forall_{\theta\in\Omega} \forall_x p(\theta ; x) \ge 0$ and
$\forall_{\theta\in\Omega} \sum_x p(\theta ; x) = 1$.  The specific
form of the function $p(\theta ; x)$ and the particulars of the
constraints on $\theta$ together specify a probabilistic model.
Hypotheses can be expressed as further constraints of the form $\theta
\in \Omega_0 \subset \Omega$.

This chapter describes three kinds of models for symbolic sequences
such as text or genomic sequences.  These three types of models, the
binomial, multinomial and Markov models, are described roughly in
order of complexity.  The first two models are special cases of Markov
models.  These three kinds of models are among the simplest to be used
for analysing language and are the ones on which the statistical tests
described in chapter \ref{new_methods} are based.

Several other kinds of models for symbolic sequences, including hidden
Mar\-kov models, interpolated $n$-gram models and exponential models,
are also described here for the sake of completeness.  These models
can be used to develop statistical tests in the same way that Markov
models are used, but because there is no closed form for the maximum
likelihood estimators for these models, their use in generalized
log-likelihood ratio tests can be prohibitively expensive in terms of
computational resources.  Interestingly, the statistical tests
described in chapter \ref{new_methods} for the Markov and simpler
models can be put to good use in estimating the parameters for the
exponential models.  These tests can also be used to build mixed-order
Markov models which can take the place of the interpolated $n$-gram
models.  These models use lower-order models as empirical priors in
order to build higher order models.  Such applications of the
likelihood ratio tests are outlined in chapter \ref{future}.

\subsection{Binomial Distributions \label{methods:binomials}}

Binomial distributions arise commonly in statistical analysis when the
data to be analysed are derived by counting the number of positive
outcomes of repeated identical and independent experiments.  Flipping
a coin is the prototypical experiment of this sort.  Testing such a
coin for fairness is a typical experiment that might be done
statistically. 

The task of counting a particular word can be cast into the form of a
repeated sequence of such binary trials by comparing each word in a
text with the word being counted.  These comparisons can be viewed as
a sequence of binary experiments similar to coin flipping except that
the probability of a successful outcome would be much lower.  In text,
each comparison is clearly not independent of all others, but the
dependency falls off rapidly enough with distance between words that
the assumption of independence is good enough to preserve the utility
of the model.  

Another assumption that works relatively well in practice is that the
probability of seeing a particular word does not vary.  Of course,
this assumption is not really true, since changes in topic may cause
the frequency of a word to vary considerably more than could be
accounted for by an assumption of constant probability.  Indeed, it is
the failure of this assumption that makes most information retrieval
techniques possible at all; such techniques work precisely because the
observed frequencies of content bearing words vary between documents
dealing with different topics.  The analysis of binomial experiments
can reveal which words vary significantly in frequency and, as is
shown in chapter \ref{ir}, this analysis can substantially improve the
performance of information retrieval systems.  Harter took a similar
tack in \cite{harter1} in which he assumed that the occurrence of
content bearing words could generally be described by a mixture of
Poisson distributions while the occurrence of non-content bearing
words could be described by a single Poisson distribution.  Since the
Poisson is simply the continuous limit of the multinomial, Harter's
work with the two Poisson model (or Poisson mixture model) bears
directly on the work described in this thesis.  The major differences
between Harter's and the work reported here are first that even though
Harter made use of the Poisson mixture model, he had only a heuristic
measure of significance to find content bearing words.  This heuristic
measure has difficulty with small counts because it is based on the
same assumptions as Pearson's $\chi^2$ test.  Another problem is that
the distribution of the Harter's test statistic vary dramatically just
as the distribution of Pearson's statistic does.  The log-likelihood
ratio test provides the measure that Harter needed to analyze the
behavior of low frequency words as well as to compare low and high
frequency words.  The current work also extends Harter's work in that
it provides a framework for analyzing both the unlabelled case that
Harter examined, but also the labeled case.  It is the labeled case
that is mostly highlighted in this thesis, but the analysis of the
unlabelled case proceeds analogously.

To the extent that the assumptions of independence and stationarity
are valid, we can switch to an abstract discourse concerning Bernoulli
trials instead of words in text, and a number of standard results can
be used.  A Bernoulli trial is the statistical idealization of a coin
flip in which there is a fixed probability of a successful outcome
that does not vary from flip to flip.  When Bernoulli trials are
repeated, the number of positive outcomes is distributed according to
the binomial distribution.  Binomial distributions can also be used more
generally to model any sort of decision that can be made on a word by
word basis.  

To be specific, if the actual probability that the next word in a text
matches a particular word is $p$, and assuming that this probability
does not depend on the other words in the text, then the number of
matches generated in the next $n$ words is a random variable ($K$)
with binomial distribution whose mean is $np$ and whose variance is
$np (1-p)$.  The probability that the number of successful outcomes
will be exactly $k$ is given by
\begin{equation*}
p(p; K=k) = p^k (1-p) ^ {n-k} {\binom n  k}
\end{equation*}
Here, the single parameter of the model is the probability of positive
outcome $p$, while the event is $K=k$.  If $np (1-p) >5$, then the
distribution of the random variable $K$ will be approximately normal,
and as $np~(1-p)$ increases beyond that point, the distribution
becomes more and more like a normal distribution.  Since the binomial
is a discrete distribution and the normal distribution is continuous,
this convergence is expressed in terms of uniform convergence of the
cumulative distributions rather than in terms of the density functions
themselves.  This progression to the normal distribution is a
consequence of the law of large numbers.  The use of this convergence
for statistical purposes is described in section
\ref{methods:standard_tests}.

The binomial distribution is useful precisely because it is such a
simple model of text.  Using tests based on the binomial distribution
allows an analysis to focus very carefully on a very specific
phenomenon.  In addition, tests based on the binomial distribution are
typically very cheap computationally.  The binomial distribution
therefore trades expressivity for simplicity more aggressively than do
any of the other distributions described here.

\subsection{Multinomial Distributions \label{methods:multinomials}}

Multinomial distributions are a generalization of binomial
distributions in which the binary outcome of the Bernoulli trial is
replaced with a choice from a finite set of symbols called an
alphabet.  The assumptions of independence and stationarity of the
process are the same as for the binomial distribution so that the
parameters of the multinomial distribution consist of the
probabilities of occurrence associated with each symbol in the
alphabet.  In essence, the multinomial model describes the generalized
rolling of a die which may have an arbitrary number of faces and which
may be fair or loaded.  These dice are assumed not to change over
time, nor does one roll affect any other roll.

With a multinomial model, the probability of observing a particular
sequence of symbols $W = w_1 \ldots w_n$ is $p(W) = \prod_{i = 1
\ldots n} p(w_i)$.  Here the parameters of the model are the
probabilities for each of the symbols in the alphabet
(i.e. $p_\sigma$ for $\sigma \in \Sigma$).  If we consider all
rearrangements of a sequence as equiprobable, then these sets of equiprobable sequences can be
characterised by the number of times each symbol occurs.  Using
the function $T$ to denote counting so that $T(\sigma, W)$ represents the
number of times that $\sigma$ occurs in the string $W$, the
probability of a particular vector of symbol counts $K$ where
$k_\sigma = T(\sigma, W)$ is given by the following formula.
\begin{equation*}
p(K) = n! \prod_{\sigma \in \Sigma}
\frac {p_\sigma^{T(\sigma,W)}} {T(\sigma,W)!}
\end{equation*}

Analogous to the case of the binomial distribution, the expected value
of $T(\sigma, W)$ is $n p_\sigma$, and the variance is $n p_\sigma
(1-p_\sigma)$.  If $n p_\sigma (1-p_\sigma) > 5$ for all $\sigma$,
then the multinomial distribution is well approximated by
$\abs{\Sigma}-1$ independent normal distributions.  This approximation
is very similar to the way that the binomial distribution can be
approximated by a single normal distribution under similar conditions.
However, when any of the values $n p_\sigma (1-p_\sigma)$ are less
than the rough limit of 5, then this approximation can lead to severe
overstatement of significance in both the binomial and multinomial
cases.  These approximation and overstatement issues are described in
more detail in section \ref{methods:chi2_comparison}.  These problems
can be largely avoided using the likelihood ratio tests described in
section \ref{methods:lr_tests}.

Using a multinomial model for text is relatively straightforward if the
text can be reliably separated into individual symbols or tokens.  For
many languages, space separated words form natural tokens; for others,
more sophisticated segmentation is necessary.  Once tokenization is
done, the text can be further processed to reduce each word to a
canonical root form in a process generally called stemming, or by
separating out all morphological indicators into special tokens.  When
stemming, {\em glasses} might be reduced to {\em glass} and {\em
causing} to {\em caus}.  When separating morphological indicators, these
might instead be reduced to the token pairs {\em glass -es} and {\em
cause -ing}.  Since a multinomial model requires that there be a
finite alphabet, the vocabulary of the language being analysed may
need to be truncated.  This truncation is usually done by defining a
reference vocabulary of known words and replacing all words outside
this vocabulary with a special unknown word token.  Nothing, of
course, prevents morphological processing from being done on these
unknown words.  For example, {\em asphyxiating} might well be reduced
to the pair of tokens {\em unknown-word} and {\em -ing}.
The extreme case of vocabulary truncation is the reduction of a
multinomial model to the binomial case.

The multinomial model expresses the some of the properties of language
such as the appearance of words at differing average frequencies.  The
multinomial model makes two assumptions which are known to be false.
First, the multinomial assumes that the frequency of each word is
constant and secondly that the each word is independent of all
others.  This second assumption implies that ordering is irrelevant to
the probability of a string.

Considering all permutations to be identical as is done in the
multinomial model is clearly invalid when either human language or
genetic sequences are considered.  Not making this assumption,
however, requires the use of a more advanced model, such as a Markov
model.  Pretending that the assumption holds, at least under limited
circumstances, is a way of trading fidelity for tractability.  Since
many useful results can be had by making the assumption of order
independence (as shown in some of the applications described in other
chapters) making this trade can be justified in some cases on
pragmatic grounds.  In any case, the multinomial model makes a less
severe tradeoff than the binomial model since more of the distinction
between words can be retained.

\subsection{Markov Models \label{methods:markov}}

Markov models are a generalization of multinomial models much as
multinomial models are a generalization of the simpler binomial
models.  Markov models add a notion of sequentiality to multinomial
models at the cost of a considerable increase in the number of model
parameters.  This gain comes without the loss of many of the very
desirable mathematical properties of the multinomial models.  In a
Markov model, both time and the current state can be either discrete
or continuous.  Normally the only Markov model used in text processing
is the discrete state/discrete time form.  This choice is so
ubiquitous that the general term Markov model is often used as a
synonym for the more restricted discrete case.

Simply stated, a Markov model consists of a set of states and a
probabilistic state transition function.  The probability distribution
of the next state depends only on the current state.  More formally,
if $\Sigma$ is the set of states and $p$ is the state transition
function, the probability of state $\sigma^\prime$ at time (or position) $i$
is 
\begin{equation*}
Pr(s_i = \sigma^\prime \mid s_{i-1} = \sigma) = p(\sigma^\prime \mid \sigma)
\end{equation*}

This definition limits the degree to which the distribution of the
current state can depend on past states.  This property of limited
history is the key to much of the mathematical simplicity of the
Markov models.  In particular, this property allows many problems to
be solved using very efficient dynamic programming algorithms.
Limited history is also the key to what makes these models somewhat
unsatisfactory as models of language.  Asymptotically speaking, a
Markov model has sufficient power to describe potentially infinite
sets of symbol sequences arbitrarily well if a few reasonable
assumptions hold, but this is only in the limiting case of arbitrarily
many states.  With any discrete Markov model which has a finite number
of states, the mutual correlation between tokens must fall off
exponentially with the separation between the tokens.  Wentian Li
\cite{wli89} demonstrated empirically that exponential correlation
decay was not observed for human language as did Peng \cite{peng} for
genetic sequences.  These observations mean that a finite Markov model
can only be an approximate model of the process which produced these
sequences.

On the other hand, this approximation can be quite good.  For example,
the interpolated trigram model produced by the IBM speech recognition
group produced the lowest measured entropy relative to English of any
contemporaneous language model of any sort (see \cite{brown92a}).
Thus, while finite Markov models are observably not correct models of
English and are not the most economical model in terms of number of
free parameters, not so long ago they were the best approximation (on
information theoretic grounds) of English ever produced.  

When using discrete Markov models with language, there are two basic
options.  The first is to use fixed length token strings taken from
the observed text to represent the Markov model state.  This option
allows the Markov state to be directly observed.  The second option is
to hide the Markov state, and assume that the observed text is a
probabilistic function of the state sequence.  The first option is
sometimes called a visible Markov model (or more commonly, just Markov
model).  The second option is the hidden Markov.  The hidden Markov
model is used extensively in speech recognition.  Hidden Markov models
are described in section \ref{methods:mm}.

With visible Markov models, the use of token strings to represent the
Markov model state allows a great simplification of the state
transition function.  This simplification is possible since successive
states share most of their labels; the only difference is the addition
of the newest token and deletion of the oldest token.  For instance,
if strings $m$ long are used to label states, then the label on the
$(i-1)$-th state in the text $w_1 \ldots w_n$ would be $w_{i-m} \ldots
w_{i-1}$.  The label on the next state would be $w_{i-m+1} \ldots
w_{i}$.  For brevity, it is conventional to write these as
$w_{i-m}^{i-1}$ and $w_{i-m+1}^{i}$ respectively.

These two labels are almost identical.  This redundancy in
state labels allows us to write the state transition function $p$ as
\begin{equation*}
Pr(w_i = w_{i-m+1}^i \mid w_{i-1} = w_{i-m}^{i-1}) =
p(w_{i} \mid w_{i-m}^{i-1})
\end{equation*}

These conditional probabilities are the parameters of the Markov
model.  This notation is used by almost all authors using visible
Markov models to analyse text, and the formal definition of the Markov
model is almost never used in the literature.  To make clear how many
symbols are being used as state labels, this Markov model is often
referred to as a Markov model of order $m$.  Another common term used
for such a Markov model is an $n$-gram model (where $n$ is $m+1$).

To simplify various derivations, it is convenient to assume that all
strings produced by a Markov model are padded with $m$ initial tokens
$\phi$ in positions $-m+1$ through $0$.  The state transition function
can be constrained so that these pad characters can never appear after
the beginning of strings produced by a Markov model.  Some authors do
not take this step and as a result must explicitly represent the
probabilities of every possible $m$ long prefix.  The use of padding
characters, on the other hand, allows the likelihood of producing a
particular string $w_1^n$ to be written as
\begin{equation*}
p(w_1^n) = \prod_{i=1}^n p(w_i \mid w_{i-m}^{i-1})
\end{equation*}

It should be noted that an order $m$ Markov model is not defined here
in terms of the joint probabilities of the various $m+1$ and $m$ long
subsequences (the so-called $n$-grams in linguistic circles or
$n$-mers in biochemistry).  This definition of the model uses
the conditional probabilities instead.  The model could, in fact, be
defined in terms of the joint probabilities of $m+1$ and $m$ long
subsequences.  In this definition, the probability of an $n$ long
string would be
\begin{equation*}
p(w_1^n ) = 
\prod_{i=1}^n \frac {p(w_{i-m}^i)} {p(w_{i-m}^{i-1})}
\end{equation*}

This assumes that the same padding trick is used as with the
definition using conditional probabilities.  There are three basic
problems with this definition:

\begin{enumerate}
\item{More parameters are required to specify the model.}
\item{The constraints on the parameters are more complex
than in the definition using conditional probabilities.}
\item{The expression involves division, which along with the second
problem, makes the derivation of the Bayesian estimators of the 
parameters {\em much} more difficult (the maximum likelihood
estimators are still fairly simple).}
\end{enumerate}

As $m$ goes to $0$, Markov models reduce to the simplest case of a
multinomial model.  This means that Markov models add only the ability
to model sequential inter-symbol dependencies to the multinomial model.
This greater expressivity comes at a cost, however.  Regardless of the
parameterization used, the set of parameters for a Markov model
increases in size dramatically as $m$ increases.  For example, the
trigram ($m = 2$) language model used by the IBM speech group for
their machine translation efforts included tens of millions of free
parameters.

\subsection{Hidden Markov Models \label{methods:mm}}

With hidden Markov models, the state of the system is not directly
observable.  Instead, what is observed is a probabilistic function of
the state sequence.  Thus, in addition to the state transition
function, there is a probabilistic output function.  In essence the
observed text is a shadow of the actual sequence of states; we
cannot observe the actual sequence, but we can draw inferences, much
as we can draw inferences by looking at shadows.

Two sorts of hidden Markov models are commonly used.  In the simpler
of the two kinds, the output depends solely on the current state.  In
the more complex, the output depends on which state transition is
made.  The difference between these two kinds of models can be likened
to the difference between Mealy and Moore state machines.

For the simpler kind of hidden Markov model, the probability of
observing a particular sequence of output symbols given that the
system goes through a particular sequence of states is
\begin{equation*}
p(w_1^n \mid s_1^n ) = \prod_{i=1}^n p(w_i \mid s_i)
\end{equation*}
and the probability that the system goes through a particular sequence
of states is
\begin{equation*}
p(s_1^n ) = \prod_{i=1}^n p( s_i \mid s_{i-1} )
\end{equation*}

In this equation, we assume that the $s_0$ is a special padding state
$\phi$.  This assumption avoids having to show the probability
distribution of the initial state explicitly.

Taken together, these give the probability of observing a particular
output sequence:
\begin{equation*}
p(w_1^n ) = \sum_{s_1^n \in \Sigma^n}
\prod_{i=1}^n p(w_i \mid s_i) \, p( s_i \mid s_{i-1} )
\end{equation*}
where the sum ranges over all possible sequences of states.  As
stated, this sum would take time exponential in $n$ to evaluate.  The
probability of a particular sequence can be computed without such an
exponential sum by successively computing the distribution of states
for each time step and noting that
\begin{align*}
p(s_i) &= \sum_{s_{i-1}\in\Sigma} p(s_i \mid s_{i-1}) \, p( s_{i-1} ) \\
p(w_i) &= \sum_{s_i\in\Sigma} p(w_i \mid s_i) \, p( s_i )
\end{align*}

This allows the computation of the probability of an output sequence
in $O(n \abs{\Sigma})$ time and $O(\abs{\Sigma})$ space.  Furthermore, given an
output sequence, the sequence of states most likely to have given rise
to that output sequence can be computed using dynamic programming in
$O(n \abs{\Sigma})$ time and space.  The $k$ most likely sequences of
states can also be computed efficiently.

The corresponding expressions for the hidden Markov model where the
output is a function of transitions rather than single states are
quite similar to the expressions given above, except for the fact that
the output function has the current and next state as arguments.  This
means that the sequence of output symbols is one shorter than the
sequence of states, but it also allows somewhat more information to be
used in determining the output symbol.  The simpler type of hidden
Markov model can emulate the more complex one at the possible cost of
introducing a great many more states.  Depending on the application,
this tradeoff may or may not be helpful.

The hidden Markov model is clearly an extension of the visible Markov
model since the output function can be trivial.  In general, however,
the internal state of the hidden Markov model cannot be observed,
which means that iterative methods such as the EM algorithm (see
section \ref{methods:mle}) must be used to estimate the values of the
parameters of model.  For very large models such as a model of all of
English, this estimation step is simply not feasible.  In many other
cases, however, hidden Markov models provide substantial value over
the simpler models described earlier.  This is particularly true when
the ability of the state of the hidden Markov model to represent some
abstraction can improve performance.

\subsection{Exponential Models}

Exponential models are a very general class of probabilistic models
which include many other models as special cases.  These special cases
include decision trees, Markov models and others.  The key to this
generality is that exponential models are defined in terms of
arbitrary feature functions which represent any function of the input
sequence.  Stated so generally, however, exponential models are worse
than useless since searching over all combinations of partial
functions is not even a computable task, much less a feasible one.

The trick with exponential models, therefore, is to define sets of
feature functions and a search algorithm which selects the feature
functions to use and then derives weights for these functions.
Hopefully the search algorithm will lead to a good model at least, if
not an absolutely optimum one.

The probability of a sequence of symbols for an exponential model is
defined by
\begin{equation*}
p(w_1^n ) =
\frac {1} {Z} {\rm exp} \left( \sum_i \lambda_i f_i (w_1^n ) \right)
\end{equation*}
where the $\lambda_i$ are the weights, the $f_i$ are the feature
functions and $Z$ is a normalizing factor which guarantees that the
sum of the probabilities of all sequences is unity.  In some cases, it
is convenient for the feature functions to be binary functions with a
range containing only the values $0$ and $1$.  Using binary functions
simplifies the computation of the weights substantially, although
computing $Z$ is still difficult.

The major problem with exponential functions is that $Z$ is a
non-linear function of all of the $\lambda_i$ and is generally quite
expensive to compute.  Since most algorithms for estimating the values
for $\lambda_i$ or selecting sets of feature functions require that
the probabilities be normalized, this computation of $Z$ must be done
repeatedly.  A family of algorithms known as iterative scaling
algorithms are usually used to estimate the values of $\lambda_i$.
Results in \cite{exponential} demonstrate the utility of an improved
iterative scaling algorithm for computing the probability of English
words given their spelling.

Exponential models have very attractive properties in that they are
able to make use of essentially any sort of computable feature of a
string.  This means that exponential models could conceivably make use
of partial parsers, semantic agreement testers or any of many other
potential sources of non-local information.  It is not clear if the
full power of exponential models is useful, however.  Specializations
such as decision models (a variation on decision trees), or
self-organized approaches as advocated by Kohonen (such as in
\cite{kohonen}) may be more practical.  Another limitation to
exponential models which may make them tractable is to require that
they be ``causal'' models.  A causal model is one which assumes that
the likelihood of text to the right of some point can be predicted
entirely by text to the left.  This limitation to causal models turns
exponential models into a very sophisticated version of the visible
Markov model and may make parameters relatively easy to estimate.  The
selection of which decision functions to use is still a very difficult
problem, however.

\section{Parameter Estimation for Text Models}

In all of the models presented to this point, there are a number of
parameters which determine the detailed shape of the distribution of
the probability for various strings of symbols.  These parameters are,
in fact, of no interest to us in and of themselves.  Rather, what is
of interest is the use of these parameters to help us predict the
likelihood of strings yet unseen or to determine how compatible a
string is with some hypothesis or other.  Speaking somewhat loosely, a
particular value for the parameters of a model represents a theory
taken from a somewhat circumscribed set of theories.  Selection of a
single theory, or reasonably small set of theories, is done by
selecting a particular value or set of values for the parameters.  

It is important to remember, however, that while the parameters in
most of the models discussed so far have been called
``probabilities'', and in spite of the intuitive concepts that such a
name calls up for many, these parameters must be distinguished from
the frequency at which certain events have been observed to occur in
some limited set of experiments.  There are two concepts here which
must be kept distinct.  First, there is our {\em observation} which
often takes the form of some sort of count of the number of times some
phenomenon of interest has been observed.  In this dissertation counts
such as these are indicated by the letter $k$ or by the counting
function $T$.  Second, there is our {\em estimate} of the parameters.
We generally base this estimate on our observations as well as our
prior knowledge.  In this dissertation, a circumflex is used to mark a
quantity being estimated.  Thus $\hat p$ is the estimate for the
parameter $p$.

Now it so happens that a very useful estimator for the parameters in
the models of interest here is the one which, {\em post facto},
maximizes the likelihood of our observation.  For many quantities,
this estimator is exactly equal to the corresponding relative
frequency.  Other estimators used for other purposes may not have this
fortuitous form, so the distinction between estimate and observation
is important to preserve.

In an ideal world, there would be a true value for the parameters of
our model.  Our intent in making an estimate of a parameter is to
divine this true value, but we must always realize that our estimate
and the true value are not the same thing.  Similarly, we must
remember that even when the observations on which we base our estimates of
parameters are entirely accurate, they are not in themselves the true
value we seek.  Our estimates of parameters are the particular theory
we select to reflect the true mechanism.  Paradoxically, in the
real world as opposed to the ideal world, our model may not even be a
reasonable reflection of the true mechanism, and as such, the ``true''
value of the parameters cannot even be said to exist.  Even in this
case, judiciously chosen estimates of parameters should make our
models reflect the true mechanism with maximal accuracy.

\subsection{Maximum Likelihood Estimators \label{methods:mle}}

Since the visible Markov model is a generalization of both the
multinomial and binomial models, deriving various estimators for the
visible Markov model gives the corresponding estimators for the less
general models.

The likelihood of a particular string being generated by a visible
Markov model can be rearranged to make explicit the number of times
that various $m+1$ long sub-strings appear in a particular observed
string $w_1^n$:
\begin{equation*}
p(w_1^n ) = n! \prod_{\sigma_0^m \in \Sigma^{m+1}}
\frac 
{p(\sigma_m \mid \sigma_0^{m-1}) ^{T( \sigma_0^m , w_1^n )}}
{T( \sigma_0^m , w_1^n )!}
\end{equation*}

We can just as well maximize the log of this expression,
\begin{equation*}
\log p(w_1^n ) = \sum_{\sigma_0^m \in \Sigma^{m+1}}
{T( \sigma_0^m , w_1^n )}
\log p(\sigma_m \mid \sigma_0^{m-1}) + C
\end{equation*}
where $C$ is a value which depends only on the counts and is thus not
subject to maximization.  This maximization is done by introducing the
multiple constraints on the conditional probabilities,
\begin{equation*}
\sum_{\sigma_m} p(\sigma_m \mid \sigma_0^{m-1}) = 1
\end{equation*}
and then using as many Lagrangian multipliers as we have constraints.

This leads to the maximum likelihood estimator for the parameters of a
visible Markov model:
\begin{equation}
\hat{p}(\sigma_m \mid \sigma_0^{m-1}) =
\frac {T( \sigma_0^m , w_1^n )} {T( \sigma_0^{m-1} , w_1^n )}
\end{equation}
 
The maximum likelihood estimator for the multinomial can be derived
by restricting $m$ to be 0,
\begin{equation}
\hat{p}(\sigma) =
\frac {T( \sigma , w_1^n )} {n}
\end{equation}

If the alphabet $\Sigma$ is restricted to two symbols, then this same form
is the maximum likelihood estimator for the binomial model.

For the case of the hidden Markov model, obtaining the maximum
likelihood estimate of the parameters given only the output symbols as
evidence is considerably more difficult than in the case of the
visible Markov model.  This difficulty occurs because the counts used
in the expressions above must be inferred rather than determined by
counting observable events.  If the state sequence were known, then
the maximum likelihood estimators would be the same as for a visible
Markov model, but since the state sequence is not known, then it must
be estimated.  The algorithm used to determine the maximum likelihood
estimators for hidden Markov models is a special case of the
Expectation Maximization (EM) algorithm called the Forward-Backward
algorithm \cite{Baum67, rabi1986}.  

In this algorithm, an initial definition of the state transition and
output functions is used to estimate distributions for the state at
each time step given the state distribution for the previous time step
and the current output.  This is known as the forward step.  Then
the final state distribution (as defined by boundary conditions or by
the final state from the forward step) is used to compute another
estimate of the state distribution at each time step given the state
distribution for the next time step.  This is known as the backward
step.

In the more complex hidden Markov model, in which the output depends
on the current and next state, these estimated distributions are then
used to estimate a new value for the output function, with the
distribution from the forward step being used for the current state
and with the distribution for the backward step being used for the
next state.  Then the data from both the forward and backward steps
can be combined to derive the output distribution.  In any case, the
estimated distribution of states is used in lieu of direct
observations of the state transitions in order to update the state
transition probabilities.

This forward-backward algorithm is repeated until convergence, which
can take a considerable amount of time for long training sequences.
The result of the algorithm is a maximum likelihood estimate for both
the state transition and output functions.

The procedure for iteratively finding the maximum likelihood estimator
of the parameters for an exponential model is computationally even
more expensive.  A training procedure was recently proposed in
\cite{exponential} based in part on the EM algorithm.  In this
training procedure, a simulated annealing method is used to sample
from a proposed exponential distribution.  This sample is used to find
features which are over- or under- represented in the proposed
distribution.  One of these features is then selected, and a scaling
parameter for this feature is estimated using an iterative scaling
algorithm.  The distribution is then normalized using another Monte
Carlo step.  This algorithm is computationally very expensive and has
not yet been applied to derive models for problems more complex than
estimating a model for English spelling.  A number of researchers have
advocated using exponential models for more ambitious tasks
(\cite{maxent95, berger95, lau93}), but the results of these efforts
have not yet been demonstrably more effective than the much simpler
methods.  Neal has described methods for accelerating Monte Carlo
estimates which may improve the feasibility of exponential models and
related methods \cite{neal}
 
\subsection{Bayesian Estimators}

The maximum likelihood estimators for the Markov models and their
specializations have such an intuitively appealing form that many make
the mistake of forgetting that $\hat p$ is only an estimate of the true
underlying probability, $p$.  This distinction is very important when
dealing with rare events since it is quite possible for events to have
non-zero probability, but still not be observed in a particular
experimental sample.  It is, however, dangerous to infer a probability
of zero for any event, no matter how many times it is not observed.
Since computing probabilities of strings so often involves products, a
zero probability for one small part of a string can propagate so that
the estimated probability for the entire string is also zero.  A
probability of zero is singularly uninformative so the effect is that
a localized mishap causes global failure.  By mechanical analogy, a
model which exhibits this globalized breakage due to local problems is
called brittle.

Brittleness can clearly lead to poor performance.  As the amount of
training data increases, the probability of encountering such a global
failure typically becomes smaller.  Unfortunately, in real
applications with strictly limited amounts of training data (or with
very complex models which have the same effect) encountering
situations not seen in the training data is almost inevitable.

This problem of zero probability estimates can be dealt with in any
number of ad hoc ways.  One such method is to set a lower bound on the
probability assigned to any event.  In many applications it is
preferable to define more carefully what the consequences of an error
in an estimate might be.  Incorporating this idea of potential loss
for particular kinds of error in parameter estimation leads to
Bayesian estimators.

With Bayesian estimators, the goal is to estimate the value of some
function of the parameters of a model in such a way as to minimize the
expected value of some loss function.  This expected value can, of
course, only be computed if we have some idea of a probability
distribution, so with Bayesian estimators, the idea of the prior
distribution of the model parameters is introduced.  When there is no
obvious prior distribution, the uniform distribution on a finite
domain is often used since it captures the intuitive concept of
maximum ignorance.  Other priors with desirable mathematical
properties can also used and in many cases are preferable to the
uniform prior.  Finding a useable closed form solution for a Bayesian
estimator is often not possible except in the special cases such as
the uniform prior distribution or other distinguished prior
distributions which lead to simple mathematical results.  In these
cases where closed form solutions are not possible, the general form
of the results obtained using the uniform prior distribution or some
class of prior distributions such as the Dirichlet distribution may be
helpful in the construction of heuristic methods.  For each particular
class of distributions, there is also a special class of prior
distributions called a conjugate distribution.  The salient
characteristic of a conjugate prior is that the form of the posterior
distribution is known.  These conjugate priors often provide a useful
alternative to the uniform prior.  Some commonly used conjugate priors
include the normal distribution, which is conjugate to itself, and the
Dirichlet distribution which is conjugate to the multinomial
distribution.  The beta distribution is a special case of the
Dirichlet distribution and is conjugate to the binomial distribution
which is the corresponding special case of the multinomial
distribution.

In other cases where there is no convenient prior.  In cases where the
uniform prior will not do and there is no handy conjugate prior to
provide a closed form solution, numerical techniques can sometimes be
used to calculate values for Bayesian estimators.

Stated formally, the Bayesian estimator for a function $\phi(\theta)$
given the prior distribution of the parameters $\theta$ is
\begin{equation}
{\mathcal B}[\phi] =
\frac 
{\int \phi(\theta) \, p(\theta) \, p(x \mid \theta) \, d \theta}
{\int p(\theta) \, p(x \mid \theta) \, d \theta}
\end{equation}

The expression $p(x \mid \theta)$ is equivalent to what earlier was
called $p(\theta ; x)$.  The reason for the change in notation is due
to the fact that $\theta$ is no longer a parameter which can be varied
at will, but rather is the value of a random variable whose expected
value must be calculated.  Thus the change in notation reflects a
change in the role of $\theta$.

Detailed discussions of the mechanics of applying Bayesian estimators
can be found in standard texts on statistics such as \cite{mood74},
\cite{degroot86} or \cite{papoulis}.  Useful discussions of the
philosophical ramifications of the Bayesian approach can be found in
the works of Good \cite{good}.  A more extreme view is taken by de
Finetti \cite{finetti}.  In any case, the detailed philosophical
ramifications of a Bayesian or frequentist point of view need not
concern us here.  What is important here is that the mathematics works
for our purposes.  Besides, Good's description of his type I, II and
III models provides an account of Bayesian inference that should be
acceptable in a frequentist frame of reference.  What is important
about Bayesian estimators is that for many commonly used loss
functions and priors, the resulting estimators spread the available
probability mass so that even events which are not observed generally
are ascribed non-zero probabilities.  The form of these estimators has
interesting parallels with the form of the log likelihood ratio test
described later.

In the case of the visible Markov model (and its specialized
variants), some useful Bayesian estimators for various functions of
the parameters are given here without derivation (which in many cases
is quite involved).  Detailed derivations are available in
\cite{wolf}.

If we assume the uniform prior distribution, then the estimator for
the parameters of the Markov model is
\begin{equation}
{\mathcal B}[p(\sigma_m \mid \sigma_0^{m-1})] =
\frac {T( \sigma_0^m , w_1^n )+1}
{T( \sigma_0^{m-1} , w_1^n )+\abs{\Sigma}}
\end{equation}
where $\abs{\Sigma}$ is the number of symbols in the alphabet, and $w_1^n$
is the training text.  In the multinomial case, $m=0$, so this becomes
Laplace's population size correction:
\begin{equation}
{\mathcal B}[p_\sigma] =
\frac {T( \sigma , w_1^n )+1} {n+\abs{\Sigma}}
\end{equation}
If we assume that the prior distribution for the probability of each
possible symbol, ${\mathbf p} = p_{\sigma_1} ... p_{\sigma_n}$ is the
generalized Dirichlet distribution,
\begin{equation*}
p({\mathbf p}) = \prod_\sigma p_\sigma^{\alpha {\mathbf m}_\sigma -1}
\end{equation*}
where $\alpha >= 0$ and $\sum_\sigma {\mathbf m}_\sigma = 1$, then we
find that the Bayesian estimator for the probability of $\sigma$ is
\begin{equation}
{\mathcal B}[p_\sigma] =
\frac {T( \sigma , w_1^n )+\alpha m_\sigma} {n+\alpha \abs{\Sigma}}
\end{equation}

It is important in many cases to estimate the value of
$\log(p(\sigma_m \mid \sigma_0^{m-1})$.  Unfortunately, due to the
non-linearity of the logarithm, the Bayesian estimator for
$\log(p(\sigma_m \mid \sigma_0^{m-1})$ is not just the log
of the estimator for $p(\sigma_m \mid \sigma_0^{m-1})$.  Instead, it is
\begin{equation*}
{\mathcal B}[\log p(\sigma_m \mid \sigma_0^{m-1})] =
{\psi (T( \sigma_m , w_1^n )+2)} - 
\psi( T( \sigma_0^{m-1} , w_1^n ) +\abs{\Sigma}+1)
\end{equation*}
where $\psi(x) = \Gamma^\prime (n) / \Gamma (n)$.  For many purposes,
the approximation $\psi(x+1) \approx \log(x) + 1/2x$ (derived using
Stirling's formula) can be used.  This approximation indicates that
the log of the estimator of the parameters to estimate the log of the
parameters may be acceptable.  In the case of the uniform prior, this
gives
\begin{equation*}
{\mathcal B}[\log p(\sigma_m \mid \sigma_0^{m-1})] \approx
\log 
\frac
  {T( \sigma_0^m , w_1^n )+1} 
  {T( \sigma_0^{m-1} , w_1^n )+\abs{\Sigma}}
\end{equation*}


\subsection{Held Out Smoothing}

The motivation given in the previous section for using Bayesian
estimates was that maximum likelihood estimators tend to behave poorly
with limited training data.  The symptom of this poor behaviour is
that when the model derived using maximum likelihood estimators is
presented with novel data from the same source as the training data,
the estimated likelihood of the novel data will have a good chance of
being zero.  This brittle behaviour is a result of what is sometimes
called over-training or over-fitting.  The root of the problem is that a
model has been built with the specific peculiarities of the training
data too much in mind.

The key to avoiding over-training is to realize when the training data
provide sufficient evidence for an involved model to be used and when
insufficient evidence has been provided so that a weaker, less
specific form of the model should be used.  Bayesian estimators
perform this function in a very general mathematical way, and they
have a good philosophical pedigree in that they describe an appealing
progression from preconceived notions to more refined estimates.

Unfortunately, Bayesian estimators are sometimes difficult to derive,
and often this difficulty leads to the use of unjustified
simplifications such as the assumption of the uniform prior
distribution.  When presented with highly structured data such as
human language, the uniform prior Bayesian estimators used often fail
to capitalize fully on the data available.  Unfortunately, estimators
which are based on more interesting priors are difficult to derive in
many cases.

Another method which achieves many of the same goals as the Bayesian
estimators, but which is better able to characterize on peculiarities
found in training data, is a method known as held out smoothing.  In
this method, the training data are divided into two parts, one
typically much larger than the other.  The larger portion of the
training data is used to derive highly simplified as well as highly
specific models while the smaller portion of the training data is held
out.  The lesser, held out, portion is then used to determine when the
more specific models should be used and when the less specific, but
less brittle models are more appropriate.  Over-training is avoided
since the highly specific models are only used when justified by the
held out data.  Furthermore, the overly cautious tendencies of the
Bayesian estimators are also avoided since real data (the held out
portion) are used to choose between specific and general models rather
than prior distributions whose only real virture is that they can be
manipulated mathematically.

The best known example of the application of held out smoothing to
text processing is the interpolated trigram model developed by the IBM
speech recognition group working on statistical methods for machine
translation \cite{stat_mt}.  In this work, a second order Markov model
was used to estimate the probability of strings.  The parameters used
to describe the model were the conditional probabilities of seeing a
particular word given the previous two words.

Since the number of possible word trigrams in English is so large,
even the very large amount of training data (nearly a billion words)
used in this model was insufficient to allow the use of maximum
likelihood estimators.  Instead, the conditional probabilities
making up the model are estimated using a linear interpolation of the
maximum likelihood estimators for the unigram, bigram and trigram
models.  That is,
\begin{equation}
\tilde{p}(w_3 \mid w_1^2) =
\lambda_{w_1^2} p(w_3 \mid w_1^2) +
\mu_{w_1^2} p(w_3 \mid w_2) +
\nu_{w_1^2} p(w_3)
\end{equation}

Here, the probabilities on the right hand side are estimated directly
from the training corpus, while the values of the $\lambda$, $\mu$,
and $\nu$ interpolation parameters are estimated using a held out
portion of the training corpus.  These parameters are subject to the
constraint that
\begin{equation*}
\lambda_{\sigma_1^3} +
\mu_{\sigma_1^2} +
\nu_{\sigma_1^2} = 1
\end{equation*}
where $\sigma_1^2 \in \Sigma^2$, the set of all bigrams.

This model has achieved very good performance based on evaluations
which use average perplexity of the model relative to the Brown
corpus.  The construction of this language model is described in
\cite{brown93}.  

One difficulty in using held out smoothing is that the final step, in
which the held out data are used to determine the interpolating
constants, is a multi-variate optimization process with very high
dimensionality.  In the IBM work, there were in excess of 100,000
smoothing parameters which had to be optimized.  Even though this
optimization problem has substantial regularities which can be
exploited to make it more efficient, this is still a very large
numerical optimization problem.  Furthermore, extending such a model
to higher than second order Markov models radically increases the size
of the optimization problem so that even though there are cases of
very long essentially fixed phrases in English in particular domains
(``staff writer of the Wall Street Journal'', for instance), it is
difficult to imagine directly extending the basic technique of held out
smoothing much further than the cases where it has already been
applied. 

Held out smoothing can be of considerable use in the source
identification problem to determine good models for the various
sources with only a small amount of training data.  In many cases,
very high quality models are not needed and much simpler methods such
as modified log-likelihood ratio techniques give very good overall
results, as is described in chapter \ref{lingdet}.

It is also possible to avoid actually solving for the best values of
the mixing parameters $\lambda_{\sigma_1^2}$, $\mu_{\sigma_1^2}$, and
$\nu_{\sigma_1^2}$ by using heuristic considerations such as unigram,
bigram and trigram frequencies to set all but one of the parameters to
zero.  The resulting model is known as a backoff model and was
described in \cite{katz}.

A general purpose and novel alternative to held out smoothing which
avoids the multi-variate optimization step is described in section
\ref{methods:iterative_models}.   

\section{Statistical tests \label{methods:standard_tests}}

Statistical models for language can be used to develop sensitive
statistical tests which can highlight the linguistic structure found
in a text corpus.  There is always the danger, however, that any
observed structure is purely coincidental.  If we take as an example
the problem of determining whether two words tend to occur in
proximity, the simple fact that these two words appear near each other
may strongly indicate some association if they occur a large number of
times together and never apart.  On the other hand, this coincident
appearance may only very weakly indicate a possible association if
they occur once together (and never separately).  In either case the
evidence for association is unanimous, but in the second case we
remain unconvinced that the association necessarily exists.

Conversely, after a large number of observations, we may become nearly
certain that two words have a very subtle association which is not due
to chance.  The observed strength of an association can be subtle or
readily apparent.  We can also be more or less certain that the
observed association is not accidental.  These two characteristics of
subtlety and accidental nature can thus appear independently.

Statistical measures of association can, unfortunately, not measure
both characteristics simultaneously.  They can measure how unsubtle an
association is, or they can measure how unlikely the observed
association is due to chance.  There is, by nature, a conflict between
tests which indicate that two words are associated and tests which
indicate that any particular association may have occurred by chance.
In this dissertation, the strongest focus is on tests which, among
other things, are useful for detecting anomalous association rather
than measuring the strength of that observed association.  

There are a number of measures of strength of association which have
been proposed in the computational linguistics literature.  The most
commonly suggested measures are described in chapter 3 of van
Rijsbergen's book \cite{rijsbergen}.  Most measures of association
strength are heuristic in nature.  There have been far fewer tests
proposed to detect anomalous association.  Essentially the only
measures which are used widely are the $Z$-score derived from
single-cell mutual information (or log association ratio) proposed by
Gale and Church \cite{church89} and the likelihood ratio test (or
$G^2$) as proposed by this author \cite{dunning93}.  The preponderance
of measures of association in the computational linguistics literature
indicates clearly that the question of whether an observed association
is due to random fluctuation has not been well addressed.  This lack
of depth of investigation is an interesting contrast to the case in
other fields where this question of significance has been a central
issue.

All of these tests are described in this section, except for the
generalized likelihood ratio test, a new method first applied to
problems in computational linguistics by this author.  The generalized
likelihood ratio test is described in chapter \ref{new_methods}.  An
outgrowth of the new methods described in chapter \ref{new_methods} is
the work using Fisher's Exact Test; this work is described in chapter
\ref{future}. 

\subsection{Tests for Strength of Association}

Among measures of association which appear in the literature, two
heuristic measures which commonly appear are the Dice coefficient and
the Jaccard coefficient.  They are closely related in that they are
computed by dividing the number of times two features appear together
by some normalizing factor.  For the Dice coefficient, this
denominator is the sum of number of times the each feature occurs (i.e
$2 {T(A \land B)}/{( T(A) + T(B) )} $ where $T(x)$ is the function
which returns the number of times the event $x$ occurs).  For the
Jaccard coefficient, this denominator is the number of times either
feature occurs ($ { T(A \land B) }/{T(A \lor B) } $).  The Jaccard
coefficient is often reinvented under a name something like
``intersection over union'' as in \cite{pathfinder}.

Both the Dice and Jaccard coefficients are well behaved in that they
have easily determined and interpreted maximum and minimum values, but
neither is well behaved in the face of small numbers of observations.
Under such conditions, the reliability of these tests suffer.
Proportional changes in the observed counts does not affect either the
Dice or Jaccard coefficients.  Indeed, proportional changes in $T(A)$,
$T(B)$ and $T(A \land B)$ leave both coefficients unchanged, regardless
of the total number of observations.  This means that if two words
appear once adjacent to each other in a corpus and never apart, then
Jaccard (and Dice) coefficient will be exactly $1$ regardless of
whether the corpus contains two words or a million.

In the extreme case of a two word corpus, this value is inevitable and
is thus unremarkable, but in the case of the million word corpus, such
a coincidence of adjacency is worthy of at least some note.  The Dice
and Jaccard coefficients provide no insight into the difference
between these situations.

\subsubsection{Mutual Information}

Mutual information is another measure of association which has
received attention in the literature.  There is some confusion,
however, since what is called mutual information in information
theoretic circles is actually average mutual information, while the
statistic called mutual information in computational linguistics is a
single component of average mutual information which might better
described as a log association ratio.  In this work, I follow the
convention from the information theoretic literature and use the term
mutual information to mean average mutual information unless confusion
seems likely.  For a random variable $X$, average mutual information
is defined in terms of entropy, $H$ where
\begin{equation}
H(X) = \sum_x p(x) \log p(x)
\end{equation}

Analogously to joint probabilities, joint entropy between two random
variables $X$ and $Y$ is defined as $H(X,Y) = \sum_{xy} p(x,y)
\log p(x,y)$.  

Mutual information between two random variables is defined as the
joint entropy less the entropy of each of the variables considered
separately.
\begin{equation}
MI(X,Y) = H(X,Y) - H(X) - H(Y)
\end{equation}

This can lead to an intuitively appealing form for mutual
information.  Starting with the expansion of the maximum likelihood
estimate for mutual information, we get the following.
\begin{equation*}
MI(X,Y) = \sum_{xy} p(x,y) \log p(x,y) - \sum_x p(x) \log p(x) - \sum_y p(y) \log p(y)
\end{equation*}
Then, noting that $\sum_x p(x,y) = p(y)$ and $\sum_y p(x,y) = p(x)$,
\begin{equation*}
MI(X,Y) = \sum_{xy} p(x,y) \left[ \log p(x,y) - \log p(x) - \log p(y) \right]
\end{equation*}
This can be simplified to the following form.
\begin{equation}
MI(X,Y) = \sum_{xy} p(x,y) \log \frac {p(x,y)} { p(x) p(y)}
\end{equation}

This last form is useful because it describes mutual information in
terms of the deviation from independence between $x$ and $y$.  If $x$
and $y$ were independent, then we would have $p(x,y) = p(x) p(y)$ and
the mutual information would clearly be zero.  By averaging the
deviation from independence, we get a measure of how independent $x$
and $y$ are.  By using a well known inequality generally attributed to
Gibbs, mutual information can be shown to have a minimum value of zero
and a maximum equal to the lesser of $H(X)$ and $H(Y)$.

The simplest way to estimate entropy and related measures from observed
data is to use maximum likelihood estimates for $p(x)$ and $p(x,y)$.
If the random variable $X$ is sampled a number of times and the
various symbols are each observed $T(x)$ times in an experiment and
$T(*) = \sum_x T(x)$, then the maximum likelihood estimate of entropy is
\begin{equation}
\hat H(X) = \sum_x \frac {T(x)} {T(*)} \log \frac {T(x)} {T(*)}
\end{equation}

The maximum likelihood estimator for joint entropy is based on the
joint counts.  If $X$ and $Y$ are observed jointly then the number of
trials where $X = x$ and $Y = y$ can be written as $T(xy)$.  To avoid
ambiguity, the number of times that $X = x$ without constraining $y$
is written as $T(x*)$, while the number of times that $Y=y$ is written
as $T(*y)$.  Analogously, the total number of trials is $T(**)$.  The
maximum likelihood estimator for joint entropy in terms of these joint
counts is
\begin{equation}
\hat H(X,Y) = \sum_{xy} \frac {T(xy)} {T(**)} \log {\frac {T(xy)} {T(**)}}
\end{equation}

The maximum likelihood estimator for mutual information based on
observed counts is
\begin{equation}
\hat {MI}(X,Y) = \sum_{xy} {\frac {T(xy)} {T(**)}}
\log {\frac {T(xy) T(**)} {T(x*) T(*y)}}
\end{equation}

One particular virtue of mutual information is that it is symmetrical
in that there are no values of $x$ and $y$ which are special in any
way.  Another way to express this symmetry is to consider mutual
information to be a function of a matrix of the counts in the
contingency table.  The rows and columns can be permuted arbitrarily
and the matrix can be transposed without changing the value of the
mutual information of the matrix.

This invariance is a characteristic clearly not shared by the Dice and
Jaccard coefficients each of which designate a special cell in
the contingency table.  If such a cell exists, perhaps because we are
trying to confirm a previously suspected relationship, then the
symmetry of mutual information may not be a particular advantage.  If,
on the other hand, the goal is to search for structure, then such
symmetry can be particularly helpful.

Mutual information as just described poses serious problems, however,
when used to analyse rare events.  There are two basic problems.
First, mutual information is a measure of the strength of association
which gives no clue about how reliable an observed association is.  An
association may be strong because two events are only ever observed
coincidently, but this may just be a coincidence if the number of
observations is too small.  This is a defect shared by all statistics
which only measure the observed strength of an association.  A second
problem is that when the phenomena of interest occur only rarely, {\em
average} mutual information can be dominated by the most common
(i.e. uninteresting) case.  This domination is not an artificact of
sampling, but rather is due to the fact that each logarithm in the
formula for average mutual information is multiplied by a probability
of occurrence.  

As an example, if two words, $A$ and $B$ each appear twice and always
appear consecutively in a corpus of ten words, the mutual information
between the appearance of these words is about $0.76$ bits out of a
maximum of $1$ bit\footnote{For a symmetric diagonal $2 \times 2$
contingency table such as this, mutual information can be simplified
to $-p \log_2 p -(1-p) \log_2 (1-p)$.  Since there are only 9 pairs in
this 10 word corpus, the maximum likelihood estimator of mutual
information for the ten word corpus is $- 0.22 \log_2 0.22 - 0.77
\log_2 0.77 = 0.7642$.  Note how the second logarithm is moderately dominant.
In the larger corpus, the coefficient on the second logarithm
increases to $0.998$, which makes the second term completely
dominant.}.  If the same coincidence is observed in a corpus of a
thousand words, then the mutual information decreases to $0.021$ bits
due to a much larger dilution of the effect of the case of interest
(where $A$ coocurs with $B$).  This decrease is in spite of the fact
that the second case is intuitively a stronger indicator of
association than the first.  This decrease means that mutual
information is best suited to comparing the strength of association
between random variables when the total number of observations is held
constant.

One example of the use of mutual information for detecting terms which
coocur anomalously often was described by van Rijsbergen
\cite{vanr77}.  In this work, van Rijsbergen used mutual information
to estimate how much the coocurrence deviated from the amount that
would have been expected given an independence assumption.  In order
to make this estimate useful with small samples, van Rijsbergen
focussed on methods for improving the estimates of the various
probabilities in question.  No effort was made, however, to compensate
for the dilution effect noted above.  The overall result in this work
was a set of term pairs whose simultaneous presence could presumably
be used to improve the performance of a retrieval system.  Harper
demonstrated in related work \cite{harper-vanr78} that finding
significant coocurrences was useful in a relevance feedback setting
and that this method could result in significant improvements in the
performance of a retrieval system relative to a reference system which
used the same weighting scheme, but did not make use of coocurring
word pairs.

The Luduan system described in chapter \ref{ir} does not use word
pairs at all so the work of Harper and van Rijsbergen does not apply
directly.  Extending the Luduan system to use word pairs derived by
the methods described by Harper and van Rijsbergen seems likely to
further improve the performance of beyond the performance levels
described there.  The Luduan system can, however, make use of features
such as word-pairs without the mechanism proposed by van Rijsbergen so
that mechanism may not be needed.

More recently, Ken Church has recommended \cite{church89} that a
version of mutual information be used which makes use of the
contribution to average mutual information from only one term of the
summation form above.  The definition Church used for this single-cell
mutual information is:
\begin{equation}
SCMI(X,Y) = \log {\frac {p(x,y)} { p(x) p(y)}}
\end{equation}

This value could also reasonably be referred to as the logarithm of
the association ratio.

For the same case of $A$ and $B$ appearing consecutively twice in ten-
and thousand-word corpora, the single cell mutual information is $2.17$
bits for the ten word corpus and $8.96$ bits for the thousand word
corpus.  Single cell mutual information incorporates a notion of
corpus size, but this awareness comes at the cost of losing the
symmetry with respect to row and column permutation that average
mutual information exhibits.

A second problem with single-cell mutual information is that it is
difficult to use it to compare the significance of cases with
differing numbers of observations.  Church attempted to avoid this
problem by using several approximations to obtain a
$z$-score\footnote{A $z$ score is a normalized score which assumes a
normal distribution with known or estimated mean ($\mu$) and variance
($\sigma^2$).  The $z$ score for a test statistic $x$ which should
have distribution $N(\mu, \sigma)$ if the null hypothesis holds is $z
= (x-\mu)/\sigma$.}  and by only considering cases with relatively
high rates of occurrence.

The difficulties with single-cell mutual information make it less
useful than other options.  A better solution than using either
average or single-cell mutual information is to use the generalized
likelihood ratio test as described in section
\ref{methods:lr_binomial}.

\subsection{Tests for Anomaly}

An alternative to testing for strength of association is to use a test
which indicates how anomalous an observed association might be.
Typically such tests for anomaly are constructed by measuring the
degree to which the observed data are inconsistent with a particular
model.

\subsubsection{Pearson's $\chi^2$ Test}

The most commonly used test to detect differences between discrete
distributions is Pearson's $\chi^2$ test.  The reference to the
inventor of the test is often dropped, and this test is usually called
simply a $\chi^2$ test after the asymptotic distribution of the test
statistic.  Since several of the test statistics described here are
asymptotically $\chi^2$ distributed, using the shortened name could
easily lead to confusion; the longer name is used here exclusively.

Pearson's $\chi^2$ test can be used to detect anomalous association.
One form of anomalous association is the dependency of word frequency
on genre; this issue was examined using Pearson's test by Kucera
and Francis \cite{kucera-francis-67}.  Another sort of anomalous
association is when two words are more likely to appear near each
other than independently.  

Pearson's test is designed to analyse a table of categorical
observations.  When looking for anomalous association between words
$A$ and $B$, the table contains the number of times where word $A$
appears (or doesn't) compared with the number times that word $B$
appears (or doesn't).  This is illustrated in the table below in which
$T$ indicates counting.

\begin{table}[ht]
\begin{center}
\begin{tabular}[h]{|l|l|}
\hline
$T(A B)$ & $T( \lnot A B)$ \\
\hline
$T(A \lnot B)$ & $T( \lnot A \lnot B)$ \\
\hline
\end{tabular}
\end{center}
\caption[The arrangement of bigram counts in a contingency table]{
The arrangement of bigram counts in a contingency table.  $T(X)$ is
the counting function whose value is the number of times $X$ occurs.
The counts are arranged so that the columns summarize the presence or
absence of $A$ and rows summarize the presence or absence of $B$.
Words other than $A$ are represented as $\lnot A$.}
\end{table}

One definition of the $\chi^2$ statistic is
\begin{equation}
{PK}(X,Y) = \sum_{xy} {\frac {\left[ T(x y) - T(x *) T(* y) / T(**) \right] ^2} {T(x *) T(* y) / T(**)}}
\end{equation}
where $T(x y)$ is the number of observations in the contingency table
cell labeled $x y$, and $T(x *) = \sum_y T(x y)$ and $T(* y) = \sum_x
T(x y)$ are the row and column sums respectively and $T(**) = \sum_{xy}
T(x y)$ is the total number of entries in the table.  This definition
can be transformed into a more traditional form by a relatively
straightforward expansion.

Pearson's $\chi^2$ statistic can be very useful for the analysis of
phenomena which can be cast into the form of contingency tables.  It
has many of the advantages of mutual information such as invariance
in the face of rearranging the rows or columns of the contingency
table.  Furthermore, $\chi^2$ tests can compare observations from
different sized samples and can be used to determine the significance
of a set of observations.

The major disadvantage of the $\chi^2$ test is that it cannot be used
reliably when the expected value of any cell in the contingency table
(which is just $T(x*) T(*y)/T(**)$) is less than approximately 5.
This problem is due to the fact that the $\chi^2$ test is ultimately
based on the assumption that the binomial distribution can be well
approximated by the normal distribution.  This is accurate enough if
the mean of the binomial is greater than $5$, but becomes radically
incorrect when the mean is considerably smaller.  In linguistic
applications, this necessary condition is generally not satisfied
except in the simplest analyses of the behavior of relatively common
words in very large corpora.  The approximation issues are described
in more detail in section \ref{methods:chi2_comparison}.

Ultimately, a major source of the virtues of the $\chi^2$ test is due
to the fact that it is explicitly based on mathematical models of
textual structure.  Unfortunately, the $\chi^2$ test is derived by
approximating these models using the normal distribution.  This
approximation introduces new assumptions about expected frequencies
which are not usually valid in the analysis of text.  The
log-likelihood ratio test described in section \ref{methods:lr_tests}
has all of the advantages of the $\chi^2$ test, but many of the
problems in analysing text are avoided with that test.  In particular,
since the log-likelihood ratio test does not depend on approximating
the multinomial distribution with normal distributions, the most
prominent failure modes of the $\chi^2$ test in language analysis can
be avoided.
\chapter[New Methods]{New Methods for Dealing with Small Counts \label{new_methods}}

\section{Overview}

This chapter describes the analysis of symbolic sequences by the use
of generalized log-likelihood ratio tests.  The fundamental
statistical methods underlying generalized log-likelihood ratio tests
are known.  This dissertation describes the first application of these
methods to text and genomic analysis.  My contribution has been to
recognize and demonstrate the applicability of generalized
log-likelihood ratio tests to the area of text and genome analysis.
My earliest report of this work appeared in \cite{dunning93, white},
but this dissertation considerably extends that work.

Likelihood ratio tests have been known, although not widely, for some
time in the field of mathematical statistics; the fundamental
innovation here is the use of generalized likelihood ratio tests for
natural language processing and other symbol sequence analyses.  While
the original outline of the theory of generalized likelihood ratio
tests was done early in the 20th century \cite{wilks}, some of the
most critical aspects in the mathematical development of likelihood
ratios were only published as late as the 1970's.  Likelihood ratio
tests have a reputation of not being as effective or as easy to apply
as tests such as Pearson's $\chi^2$ test\cite{agresti}.  In fact,
though, the analysis of language very often involves situations where
likelihood ratio tests are enormously more effective than Pearson's
test.  Examples of this effectiveness are given later in this chapter.
Computationally, likelihood ratio tests are no more difficult to apply
than Pearson's test unless one is limited to hand calculations.  An
interesting sidelight is that Pearson's test can be derived as an
approximation of the likelihood ratio test for multinomially
distributed random variables.

This chapter starts with a mathematical description of the likelihood
ratio test and an indication of why it is highly preferable to
$\chi^2$ tests in natural language processing.  Subsequently, the
practical issues involved in applying likelihood ratio tests to
natural language are examined.  This examination provides the basic
infrastructure used in later chapters where examples of real
applications for these techniques are described.

Likelihood ratio methods can also be used in many other applications.
Outlines of how these tests can be used to build variable order
Markov language models, to create minimum description length lexicons
for segmenting Asian languages, as part of the process of building
decision trees as well as other applications are given in chapter
\ref{future}.

\section{Likelihood Ratio Tests \label{methods:lr_tests}}

Statistical tests are generally used to provide a numerical measure of
how much an experimental observation contradicts some abstract
assumption.  This assumption is called the null hypothesis.  If the
distribution of the test statistic is known given that the null
hypothesis is true, then experimental observations which result in
extreme values of the test statistic can be said to contradict the
null hypothesis.  Extreme here means that the value of the test
statistic falls into a predefined range of values which has a suitably
small total probability under the null hypothesis.

The null hypothesis is not in itself of particular interest.  Rather,
the cases of interest are those in which the null hypothesis is
violated.

A null hypothesis can often be formulated so that it is represented by
a restricted form of model.  For models which are described in terms
of continuous-valued parameters, this restriction can sometimes be
expressed by placing an additional constraint on the model parameters,
such as declaring that some parameters must be equal to each other.

Generalized likelihood ratio tests are exactly the comparison that we
need.  These tests examine how well the more restrictive models
corresponding to the null hypothesis fit the observed data compared to
how well the more general models fit the data.  When the more general
models fit the observed data much better than do the restricted
models, the log-likelihood ratio test statistic is large.  When the
more general models do not fit much better than the restricted models,
then the log-likelihood ratio test statistic is relatively small.  As
such, the log-likelihood ratio test serves as a mathematical
expression of Occam's razor\cite{occam}.

Further background on the mathematical theory underlying generalized
likelihood ratio tests can be found in texts on theoretical
statistics\cite{mood74}.

\subsection{The Likelihood Ratio \label{methods:lr_def}}

The likelihood ratio itself is just the ratio of the maximum
likelihood of the observed data for all models where the null
hypothesis holds to the maximum likelihood of the observed data for
all models where the null hypothesis may or may not hold.

If we write the observed data as $X$, the parameters of the models as
$\omega$ and the likelihood of $X$ for particular values of $\omega$
as $p(X \mid \omega)$, then the likelihood ratio $\lambda$ is
\begin{equation}
\lambda =
\frac 
  {\max_{\omega \in \Omega_0 } p( X \mid  \omega)}
  {\max_{\omega \in \Omega } p( X \mid  \omega)}
\end{equation}
where $\Omega_0$ is the set of values of $\omega$ where
the null hypothesis holds and $\Omega$ is the set of all permissible
values for $\omega$.

Since $\Omega_0 \subset \Omega$, the unrestricted case will always fit
at least as well as the restricted case where the null hypothesis
holds.  This means that $\lambda \le 1$.

Regardless of whether the null hypothesis holds, the unrestricted case
is likely to allow a better fit to observed data than the restricted
case, if only due to slight random variations from the ideal.  There
is, however, a difference between the situation when the null
hypothesis holds and when it doesn't.  If the null hypothesis holds,
the unrestricted case (when $\omega \in \Omega$) will probably only
allow a better fit than the restricted case (when $\omega \in
\Omega_0$) due purely to random variation in the observations.  Such
improvement is likely to be quite small if the null hypothesis is
true.  If, on the other hand, the null hypothesis is actually false,
then the unrestricted case is liable to fit the data {\em
considerably} better than the restricted case.  Somewhat astoundingly,
the asymptotic distribution of $\lambda$ when the null hypothesis
holds is the same for a very wide class of models given some easily
met limitations on the form of $\Omega_0$ relative to
$\Omega$\cite{Chernoff1954,mood74,agresti}.  Thus we can state a
criterion of the form $\lambda \le \alpha$ and know just how likely
this criterion is to happen if the null hypothesis is true.  If we
have enough data so that the distribution of $\lambda$ is close to the
asymptotic distribution, then this critical value $\alpha$ is
independent of the structure of our model.

In practice, the log-likelihood ratio, $-2 \log \lambda$, is a more
useful test statistic than the raw likelihood ratio.  One strong
practical incentive for this preference is that using the
log-likelihood ratio avoids problems with numerical overflow.  At
least as important, however, is that using twice the natural log of
$\lambda$ transforms the asymptotic distribution into the well known
and extensively studied $\chi^2$
distribution\cite{Chernoff1954}.  This distribution is the
same as the asymptotic distribution of Pearson's $\chi^2$ statistic
\cite{edm400K} (the similarity in name is not coincidental).  This
similarity of asymptotic distribution means that results from a
likelihood ratio test and from Pearson's $\chi^2$ can be compared
directly.
\begin{table}[ht]
\begin{center}
\begin{tabular}[ht]{| l l l | l l l |}
\hline
\multicolumn{3}{|c}{Sample A} & \multicolumn{3}{c|}{Sample B} \\
\hline
-1.403026 & -0.144247 &  0.357771 &  -2.378491 &  -0.183818 &   0.560698 \\
-1.274699 & -0.128707 &  0.421952 &  -2.012306 &  -0.048246 &   0.863809 \\
-1.263579 & -0.127009 &  0.682820 &  -1.205998 &   0.011868 &   0.938207 \\
-1.084593 & -0.095151 &  0.706464 &  -1.170100 &   0.071464 &   0.969654 \\
-0.902815 & -0.038444 &  0.744047 &  -1.156853 &   0.190475 &   0.988129 \\
-0.692947 & -0.032206 &  0.818873 &  -1.147635 &   0.212006 &   1.130579 \\
-0.664504 &  0.046642 &  0.934158 &  -0.808311 &   0.278394 &   1.137118 \\
-0.455702 &  0.106907 &  1.046340 &  -0.699006 &   0.404206 &   1.220452 \\
-0.356289 &  0.231216 &  2.200705 &  -0.507104 &   0.497376 &   1.720593 \\
-0.266931 &  0.257050 &  2.532218 &  -0.487091 &   0.520730 &   1.799110 \\
\hline
\end{tabular}
\end{center}
\caption[Samples from unit normal distribution]{Two sets of 30 independent samples from the normal distribution with zero mean and unit variance.  The samples have been sorted to ease comparison. \label{methods:sample1}}
\end{table}

\begin{figure}[htb]
\begin{center}
\includegraphics[]{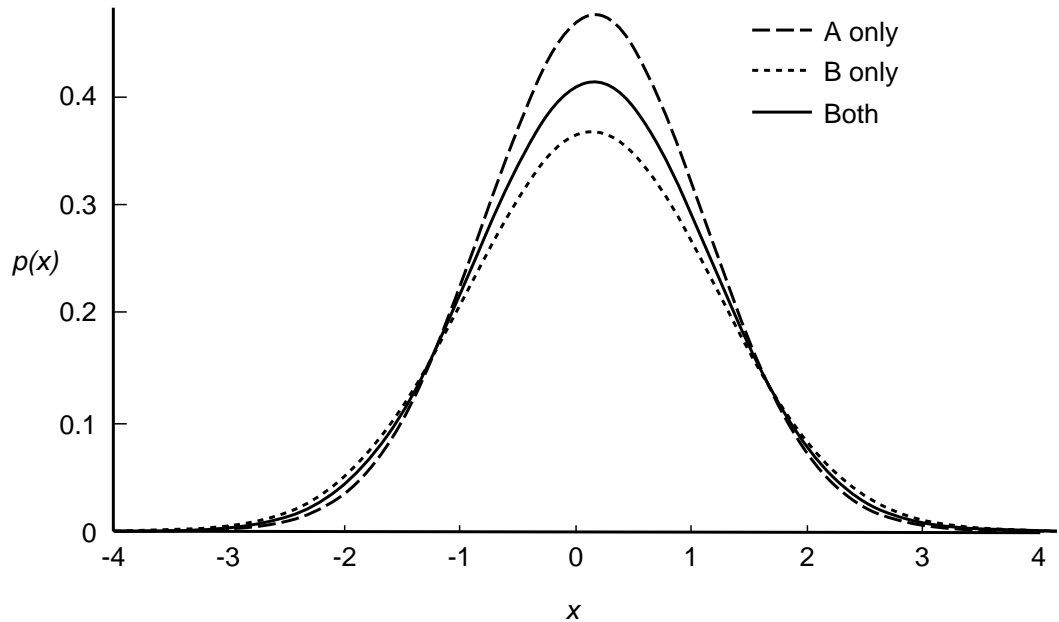}
\caption[Maximum Likelihood Distributions]{Maximum likelihood
distributions for separate and pooled data.  When the data come from
the same distribution, then fitting each group separately has no particular
advantage since the estimate for the pool is nearly the same.
\label{fig:sample1}}
\end{center}
\end{figure}

As an example of how the likelihood ratio test works, two sets of
numbers, each sampled from the same normal distribution were used in
comparative tests.  The numbers themselves are shown in table
\ref{methods:sample1}.
As can be seen in figure \ref{fig:sample1}, the
maximum likelihood normal distributions for each of the datasets are nearly
indistinguishable from each other and from the maximum likelihood
distribution for the two datasets taken together.  In this case, the
log likelihood ratio test comparing the means of the two sample distributions against each other, assuming equal variances, has a value of $0.00355$ which indicates that
the hypothesis that the two datasets are from normally distributed
random variables with the same mean and variance cannot be rejected.

\begin{figure}[htb]
\begin{center}
\includegraphics[]{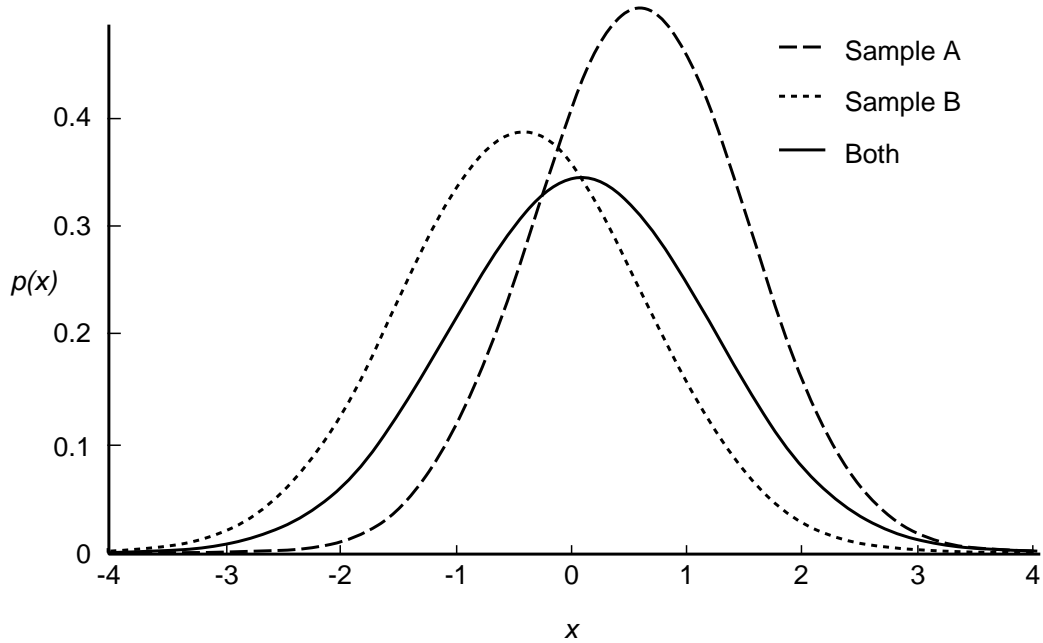}
\caption[Maximum Likelihood Distributions]{Maximum likelihood 
distributions for the shifted datasets.  When the data comes from
sources with different means, then fitting each group separately
provides a much better fit to the data.  In such a case, the estimate
derived from the pooled data cannot fit either sample very well.
\label{fig:sample2}}
\end{center}
\end{figure}
Figure \ref{fig:sample2} shows what happens when the mean for the
distribution for sample A is changed to $0.5$ while the mean for the
distribution for sample B is changed to $-0.5$.  This change makes
these maximum likelihood distributions easily distinguishable.  Now,
the maximum likelihood distribution estimated from the pooled samples
cannot fit either sample terribly well.  The distributions estimated on
each sample separately, however, fit the data as well as ever.  After
the shift, the log-likelihood ratio test has a value of $14.16$
indicating that the hypothesis that the two samples are from different
distributions is a much better explanation than the hypothesis that
they from the same distribution.  This value of the log-likelihood ratio test indicates a significance of $p < 0.000168$ which is highly significant.

Normally, of course, Student's $t$-test would be used to examine the
difference between two samples like this.  When applied to the
unshifted dataset, an unpaired, two-tailed $t$-test assuming equal
variance gives a significance of $p < 0.95$ which indicates that there
is essentially no indication of difference between the two samples.
With the shifted dataset, the same $t$-test gives a significance of $p
< 0.000181$ which is a slightly weaker indication than the
log-likelihood ratio test gave.  The log-likelihood ratio test
provides no special advantages over Student's $t$-test in this
example, other than perhaps some conceptual simplicity.  In other
situations, however, where the assumptions required by the $t$-test
cannot be justified, log-likelihood ratio tests can often be used to
great advantage.  In fact, many well-known statistical tests such as
the $t$-test can be derived using generalized log-likelihood ratio
methods
\cite{edmappa,papoulis,agresti}.

\subsection{Likelihood Ratio for Binomials and Multinomials \label{methods:lr_binomial}}

The likelihood ratio statistic for binomials and multinomials can be
derived directly from the definition of the likelihood ratio test.
As described in section \ref{methods:multinomials}, with a multinomial
model, the probability of a particular sequence of symbols $S = s_1
\ldots s_n$ is $p(S) = \prod_{k = 1}^n p(s_k)$.  If this outcome is
expressed instead in terms of the symbol counts $T(\sigma, S)$, then
the probability will be
\begin{equation*}
p \left(T(\sigma_1, S), \ldots, T(\sigma_m, S) \mid p_1 \ldots p_m \right) = n! \prod_{i=1}^m \frac {p_i^{T(\sigma_i, S)}} {T(\sigma_i, S)!}
\end{equation*}

The parameter space for these models is the set of all values for the
various $p_i$ for which the constraints $p_i >= 0$ and $\sum_i p_i =
1$ are satisfied.  For brevity, the model parameters can be collected
into a single parameter as can the observed counts.  Using this
notation, the likelihood function is written as
\begin{equation*}
p( k  \mid  \theta) = n! \prod_{i=1}^m \frac {p_i^{T(\sigma_i, S)}} {T(\sigma_i, S)!}
\end{equation*}
where $\theta$ is considered to be a point in the parameter space $P$ for
multinomial models, and $k$ a point in the space of observations $K$.

One very important null hypothesis is that two independent sets of
observed count vectors $k_1$ and $k_2$ for the two symbol sequences
$S_1$ and $S_2$ generated by multinomial random processes were
generated by identical processes.  The likelihood function for the two
sets of observations together is $p( k_1 \mid \theta_1) \, p(k_2 \mid
\theta_2)$.  The unrestricted parameter space would be $\Omega = P
\times P$ while the null hypothesis is the subset of the unrestricted
space where all parameters are identical, $\Omega_0 = \{(\theta_1,
\theta_2) \in \Omega \mid \theta_1 = \theta_2 = \theta \}$.  For the
null hypothesis, the likelihood of the observed data is $p(k_1 \mid
\theta) \, p(k_2 \mid \theta)$.

The maximum likelihood for the case when the null hypothesis holds is
given by 
\begin{equation*}
\hat p_{1i} = \hat p_{2i} = \mu_i = 
\frac {T(\sigma_i, S_1) + T(\sigma_i, S_2)} {T(*, S_1+S_2)}
\end{equation*}
In the unrestricted case the maximum likelihood is achieved when
\begin{align*}
\hat p_{1i} &= \pi_{i1} = \frac {T(\sigma_i, S_1)} {\sum_u T(\sigma_u, S_1)} \\
\intertext{and}
\hat p_{2i} &= \pi_{i2} = \frac {T(\sigma_i, S_2)} { \sum_u T(\sigma_u, S_2) } 
\end{align*}
The derivation of maxima such as these by the use of the method of
Lagrangian multipliers is outlined in section \ref{methods:mle}.

Using these maximum likelihood estimators, the value of the
log-like\-li\-hood ratio statistic is
\begin{equation}
-2 \log \lambda = 2 \sum_{ij} T(\sigma_i,S_j) \log {\pi_{ij} / \mu_i}
\end{equation}

It is also convenient to express the log-likelihood ratio in terms of
the row and column sums:
\begin{equation}
- 2 \log \lambda = 2 \sum_{ij} T(\sigma_i,S_j)
\log {\frac
    {T(\sigma_i,S_j) \left[ \sum_{uv} T(\sigma_u, S_v) \right]}
    {\left[\sum_v T(\sigma_i, S_v) \right] \left[ \sum_u T(\sigma_u, S_j) \right]}
}
\end{equation}

This log-likelihood ratio statistic is asymptotically
$\chi^2(\abs{\Sigma}-1)$ distributed where $\Sigma$ is the alphabet from
which the $\sigma_i$ are drawn.  If $N$ strings are compared, instead
of just two, then this statistic is asymptotically
$\chi^2$-distributed with $(N-1) (\abs{\Sigma}-1)$ degrees of freedom.  It
is claimed in \cite{agresti} that the log-likelihood ratio statistic
does not converge to the asymptotic distribution as quickly as does
Pearson's $\chi^2$ statistic, but the log-likelihood ratio test is
actually far superior to Pearson's $\chi^2$ in situations similar to
those found in the analysis of text.  This superiority is shown
graphically in section \ref{methods:chi2_comparison}.

Other null hypotheses can be used to derive log-likelihood ratio tests
for multinomially distributed random variables, but most of the
questions that seem to come up in the analysis of word frequency can
be cast into a framework which permits the use of the log-likelihood
ratio test just described.  Several applications of the log-likelihood ratio
test for multinomials are described in chapters \ref{colloc},
\ref{ir}, and \ref{intron}.

\subsection{Likelihood Ratio for Markov Models \label{methods:lr_markov}}

As described in section \ref{methods:markov}, Markov models are a
generalization of multinomial models which allows a limited notion of
word order to be represented.  As such, Markov models can describe
more interesting phenomena than multinomial models.  Markov models
also provide a correspondingly larger variety of interesting null
hypotheses than do multinomial models.  Interestingly, these null
hypotheses often lead to log-likelihood ratio tests which are
identical in form to the test derived in section
\ref{methods:lr_binomial}.

\subsubsection{Are two strings from the same source? \label{methods:source}}

One useful null hypothesis is that the Markov models which
generated two strings are in fact identical.  This
hypothesis is quite similar to the one used in section
\ref{methods:lr_binomial}, except that many more parameters need to be
equated.  For Markov models of order $k$, the likelihood function for
a particular string $S$ is
\begin{equation*}
p( S \mid \theta) = \prod_{i=0}^{\abs{S}}p(s_i \mid s_{i-k}^{i-1})
\end{equation*}
When all rearrangements with the same $k+1$-gram counts are considered
identical, then this can be rearranged to be
\begin{equation*}
p( S \mid \theta) =
\abs{S}\,!
\prod_{\sigma_0^k \in \Sigma^{k+1}}
\frac {p(\sigma_k \mid \sigma_0^{k-1})^{T(\sigma_0^k, S)}} {T(\sigma_0^k, S)!}
\end{equation*}

The null hypothesis that the two strings $S_1$ and $S_2$ are generated
by the same Markov models is expressed by setting all of the
parameters for each of the two models to be equal.  This is expressed
by
\begin{equation*}
p_1(\sigma_k \mid \sigma_0^{k-1}) =
p_2(\sigma_k \mid \sigma_0^{k-1})
\end{equation*}
where $\sigma_0^k \in \Sigma^{k+1}$.

Following essentially the same chain of derivation as for
multinomials, the generalized log-likelihood ratio statistic for this
null hypothesis is
\begin{equation}
-2 \log \lambda =
2 \sum_{\sigma_0^k \in \Sigma^{k+1}}
T(\sigma_0^k, S_1) \log \frac {\pi_1(\sigma_k \mid \sigma_0^{k-1})} {\mu(\sigma_k \mid \sigma_0^{k-1})} +
T(\sigma_0^k, S_2) \log \frac {\pi_2(\sigma_k \mid \sigma_0^{k-1})} {\mu(\sigma_k \mid \sigma_0^{k-1})} 
\end{equation}
where
\begin{align*}
\pi_i(\sigma_k \mid \sigma_0^{k-1}) &=
{\frac
  {T(\sigma_0^k, S_i)}
  {T(\sigma_0^{k-1}, S_i)}
} \\
\mu(\sigma_k \mid \sigma_0^{k-1}) &=
{\frac
  {T(\sigma_0^k, S_1) + T(\sigma_0^k, S_2)}
  {T(\sigma_0^{k-1}, S_1) + T(\sigma_0^{k-1}, S_2)}
}
\end{align*}

By referring back to section \ref{methods:lr_binomial}, it can be seen
that this formula is the sum of a number of log-likelihood ratios, one for
each possible $k$ long prefix $\sigma_0 \ldots \sigma_{k-1}$.  Thus,
software which is used to implement the test described in section
\ref{methods:lr_binomial} can be used to compare strings using Markov
models.  This log-likelihood ratio statistic for Markov models is
asymptotically distributed with $(\abs{\Sigma}-1)(\abs{\Sigma^k}-1)$ degrees of
freedom.

This log-likelihood ratio test statistic for Markov models can be
expressed as a set of tables.  As an example, suppose that our
alphabet has three symbols so that $\Sigma = \{a,b,c\}$.  Then, for
each possible $k$ long prefix of symbols, $\sigma_0 \ldots
\sigma_{k-1}$ (written below as $\sigma_0^{k-1}$ for brevity), we would
need to construct a contingency table which compares the frequency at
which various symbols follow the prefix in the two observed strings
$S_1$ and $S_2$.  Such a contingency table is shown in Table
\ref{methods:markov_table}.  In this table, each row represents the
counts for an observed string ($S_1$ or $S_2$) and each column
represents the counts for a particular symbol $\sigma_k$.

\begin{table}[ht]
\begin{center}
\begin{tabular}[ht]{r r@{} l c c c c c l@{} l}
$\sigma_0 \ldots \sigma_{k-1}$ &&  & $\sigma_k = a$ && $\sigma_k = b$ && $\sigma_k = c$ \\
\hhline{~~-------}
$S_1$&& \vline& $T(\sigma_0^{k-1} a, S_1)$ &\vline& $T(\sigma_0^{k-1} b, S_1)$ &\vline& $T(\sigma_0^{k-1} c, S_1)$ &&\vline\\
\hhline{~~-------}
$S_2$&& \vline& $T(\sigma_0^{k-1} a, S_2)$ &\vline& $T(\sigma_0^{k-1} b, S_2)$ &\vline& $T(\sigma_0^{k-1} c, S_2)$ &&\vline\\
\hhline{~~-------}
\end{tabular}
\end{center}
\caption{The arrangement of $k+1$-gram counts in a contingency table \label{methods:markov_table}}
\end{table}

It should be noted that the summation in this formula need only run
over values of $\sigma_0^k$ such that $T(\sigma_0^k, S_1) > 0$ and
$T(\sigma_0^k, S_2) > 0$.  Whenever either count is $0$, the
corresponding term in the summation is zero.  In addition, if $S_2$ is
much larger than $S_1$, then $T(\sigma_0^k, S_1) \ll T(\sigma_0^k,
S_2)$ will generally hold even if $\pi_1(\sigma_0^k) >
\pi_2(\sigma_0^k)$.  In this special case, this log-likelihood ratio
can be approximated by the form
\begin{equation}
-2 \log \lambda \approx 2
\sum_{\sigma_0^k \in S_1}
T(\sigma_0^k, S_1) \log
{\frac
  {T(\sigma_0^k, S_1) \left( T(\sigma_0^{k-1}, S_1)+T(\sigma_0^{k-1}, S_2) \right)}
  {T(\sigma_0^{k-1}, S_1) \left( T(\sigma_0^k, S_1)+T(\sigma_0^k, S_2) \right) }
}
\end{equation}

This simpler form and variants of it are used in chapters
\ref{lingdet} and \ref{bio_id}.  In these applications (language
identification and species identification), the training data is
always much larger than the string being identified.  Potential
applications of the unsimplified form are described in chapter
\ref{future}. 

This form can also be used to determine which of several training
strings are most compatible with a test string.  In practice, it is
helpful to combine the information from Markov models of different
orders so that information from various scales can be used in the
decision.  This can be done nicely by noting that the probability of a
symbol sequence as estimated by a Markov model of order $k$ is 
\begin{align*}
p_k(\sigma_1^n) &= \prod_{i=1}^n p(\sigma_i \mid \sigma_{i-k}^{i-1}) \\
p_k(\sigma_1^n) &= \prod_{i=1}^n 
\frac {p(\sigma_{i-k}^i)} {p(\sigma_{i-k}^{i-1})}
\end{align*}
This means that the product of the probabilities estimated by each of
the Markov models of order from $1$ to $k$ is
\begin{equation*}
\prod_k p_k(\sigma_1^n) = \prod_{i=1}^n p(\sigma_{i-k}^i)
\end{equation*}

This same construction can be used to show that the sum of the
log-likelihood ratios for Markov models of order $1$ to $k$ as
approximated using the simplified formula above can be written as
\begin{equation}
-2 \log \lambda \approx 2
\sum_{\sigma_0^k \in S_1}
T(\sigma_0^k, S_1) \log
{\frac
  {T(\sigma_0^k, S_1) \left( T(*^k, S_1)+T(*^k, S_2) \right)}
  {T(*^k, S_1) \left( T(\sigma_0^k, S_1)+T(\sigma_0^k, S_2) \right) }
}
\end{equation}

This last form is very provocative because it is essentially saying
that smoothing of the model as derived from the training data can be
done by using the observations in the test data.  Conceptually
speaking, if the training and test data are from the same source, then
combining them to form an estimate of the model is justified, while if
they are not from the same source, then the log-likelihood ratio
should tell us that the composite model formed from training and test
data together does not fit the test data very well.  Considerable
earlier work in smoothing distributions (such as in \cite{Good53a,
katz, brown92a} and many others) has focussed entirely on
smoothing based only on the training data, possibly by holding out
some of the training data for smoothing.  The approach indicated here
by the log-likelihood ratio can, in contrast, make use of all of the
training data as well as the test data for smoothing.

This last form is very similar to the Bayesian estimator derived using
a uniform prior.  The difference is that the Bayesian estimator
smooths over all possible data while the log-likelihood ratio test
smooths only over tuples which are actually observed in the test data.
Bayesian smoothing with a uniform prior quickly becomes useless
because the number of possible tuples is exponential in size and thus
the smoothing quickly swamps any observed structure in the training
data.  Since we are assuming that the training data are considerably
larger than the test data, we can safely assume that smoothing which
operates only on tuples observed in the test data cannot easily swamp
the signal observed in the training data.

As is the case with Bayesian smoothing, it is often true that even
the limited smoothing indicated by the log-likelihood ratio is more
than is optimal.  The amount of smoothing can be decreased easily by
multiplying the counts from the test data by a heuristic reduction
factor. 

\subsubsection{Can nested models be justified?}

Another class of null hypotheses for Markov models of text can be used
to compare different models applied to a single observed string rather
than comparing two observed strings.  This sort of comparison can be
used to determine whether a higher order Markov model fits a set of
training data sufficiently better than a lower order model in order to
warrant using the more complex model.  For example, to compare an
order $k$ model with an order $k-1$ model, the null hypothesis would
be formulated by setting the conditional probabilities which define
the order $k$ model to be identical conditional probabilities which
define the order $k-1$ model.

Of perhaps more interest on the general theme of log-likelihood
ratios and Markov models is their use to build variable order Markov
models.  To do this, we need to compare two order $k$ models where one
of the models is only slightly simpler than the other.  There have
been tentative steps in the direction of constructing mixed order
Markov models as in \cite{samuelsson96,ristad}, but having a simple
statistical test available to guide the process of creating these
mixed order models should be very helpful.  Further discussion of the
implications of this issue is found in chapter \ref{future}.

\section[Comparison With Other Methods]{Comparison with Other Methods \label{methods:chi2_comparison}}

The mathematical connections between log-likelihood ratio tests for
multinomials, Markov models and other methods are deep and
substantial.  There are, however, substantial elements which
differentiate log-likelihood ratio tests from other methods and which
provide advantages.

\subsection{Comparison with Pearson's Chi-Squared Test}

When contingency tables of the sort which have been discussed here are
analysed, the test which is most often used is Pearson's $\chi^2$.  In
fact, Pearson's $\chi^2$ test is an approximation of the
log-likelihood ratio test which is described in section
\ref{methods:lr_binomial}.  This approximation can be derived by
using the normal distribution to approximate the binomial or
multinomial distribution on which the log-likelihood ratio test is
based.

As is shown in this section, however, this approximation leads to
serious error in exactly the circumstances which are of most interest
in the analysis of text.

\subsubsection{The normal distribution as an approximation of the binomial}

\begin{figure}[htb]
\begin{center}
\includegraphics[]{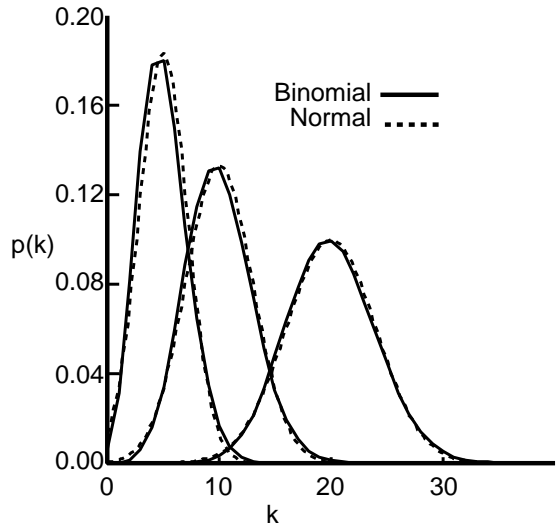}
\caption[Binomial and normal distributions compared]{Probability density functions for binomial and normal distributions compared.  As the
variance of the binomial distribution increases beyond roughly $5$, it
becomes more and more closely approximated by the normal distribution
with the same mean and variance.
\label{fig:normal}}
\end{center}
\end{figure}

One important characteristic of the binomial distribution is that when
the variance exceeds roughly $5$, the continuous normal distribution
becomes an increasingly good approximation of the discrete binomial
distribution.  This convergence is illustrated in Figure
\ref{fig:normal}.

On the other hand, when the variance of the binomial distribution is
small, the use of the normal approximation leads to greater and
greater discrepancy.  This is hardly surprising; for $n = 100$ samples
with a probability of success of $p = 0.001$, the mean of the binomial
distribution is $np = 0.1$ and the standard deviation is $\sqrt {n p (1-p) }
\approx 0.3$.  A normal distribution with similar mean and variance
would have considerable mass to the left of zero, and thus the normal
distribution indicates that the chance of a negative number of
successful binomial trials is nearly one in two.  Since negative
counts are impossible, this observation alone makes the normal
distribution untenable.  Just as disturbing as the poor behavior on
the left is the fact that the normal approximation predicts a much too
low chance for counts above one.  This problem is illustrated in Table
\ref{table:normal_approx} where the normal distribution is shown to
overestimate the significance of observing one or more successful
trials by over $200$ orders of magnitude.

\begin{table}[ht]
\begin{center}
\begin{tabular}[h]{|l|l|l|l|}
\hline
$p$ & $n$ & $p_{actual}(k \ge 1)$ & $p_{normal}(k \ge 1)$ \\
\hline
$0.01   $ & $100$ & $6.340 \times 10^{-1}$ & $7.005 \times 10^{-01} $ \\
$0.001  $ & $100$ & $9.521 \times 10^{-2}$ & $1.787 \times 10^{-02} $ \\
$0.0001 $ & $100$ & $9.951 \times 10^{-3}$ & $5.217 \times 10^{-22} $ \\
$0.00001$ & $100$ & $9.995 \times 10^{-4}$ & $1.927 \times 10^{-217}$ \\
%
\hline
\end{tabular}
\end{center}
\caption[The normal approximation leads to serious error]{The normal approximation leads to serious error for low expected counts. \label{table:normal_approx}}
\end{table}

These results indicate just how serious the difference between the
actual probability and the normal distribution can be when assessing
the significance of observing rare events.

\subsubsection{Pearson's $\chi^2$ as a log-likelihood ratio}

If in the derivation of the likelihood ratio test for the multinomial, we
had initially approximated the binomial distribution with a normal
distribution with mean $np$ and variance $np (1-p)$ then we would have
arrived at another form which is a good approximation of $- 2 \log
\lambda$ when $np (1-p)$ is more than roughly 5.  This form is
\begin{align*}
-2 \log \lambda &= 2 \sum_{ij} T(\sigma_i,S_j) \log {\mu_i / \pi_{ij}}
\approx \sum_{ij} \frac {(T(\sigma_i, S_j) - \mu_i n_j )^2} {n_j \mu_i (1 - \mu_i ) } \\
\intertext{where}
\mu_i &= \frac {\sum_j T(\sigma_i, S_j)} {\sum_{ij} T(\sigma_i, S_j)}
\intertext{as in the multinomial case above and}
n_j &= \sum_i T(\sigma_i, S_j) 
\end{align*}
are the row sums.

\begin{figure}[htb]
\begin{center}
\includegraphics[]{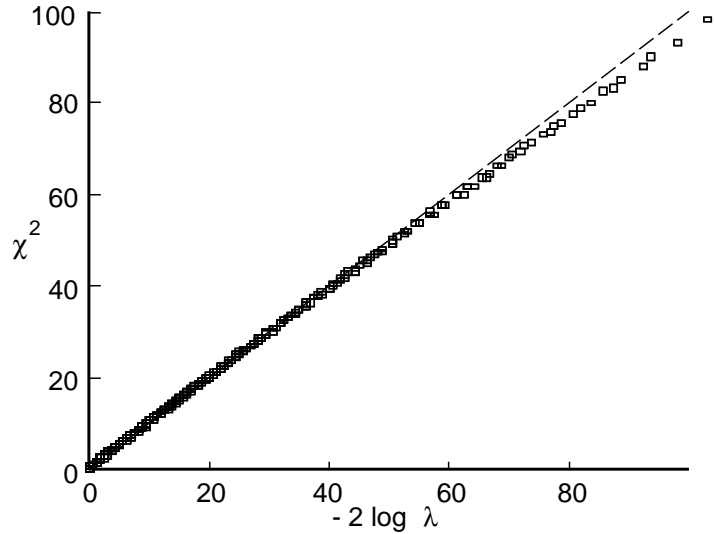}
\caption[Pearson's $\chi^2$ versus log-likelihood ratio statistics]{Pearson's $\chi^2$ versus log-likelihood ratio statistics.
Here, $p=0.1$, $n_1 = n_2 = 1000$ which means that Pearson's test is
applicable.  The plot was made by sampling from two independent
binomial distributions and testing the samples for independence using
Pearson's and the log-likelihood ratio tests. \label{fig:exact}}
\end{center}
\end{figure}

Interestingly, this expression is exactly the test statistic for
Pearson's $\chi^2$ test, although the form shown is not quite the
customary one.  Figure \ref{fig:exact} shows the reasonably good
agreement between this expression and the exact binomial
log-likelihood ratio derived earlier.  In this figure, $p = 0.1$ and
$n_1 = n_2 = 1000$ for various values of $T(\sigma_1,S_1)$ and
$T(\sigma_1,S_2)$.  Note that $np = 100$ so that the normal
approximation is justified.

\begin{figure}[htb]
\begin{center}
\includegraphics[]{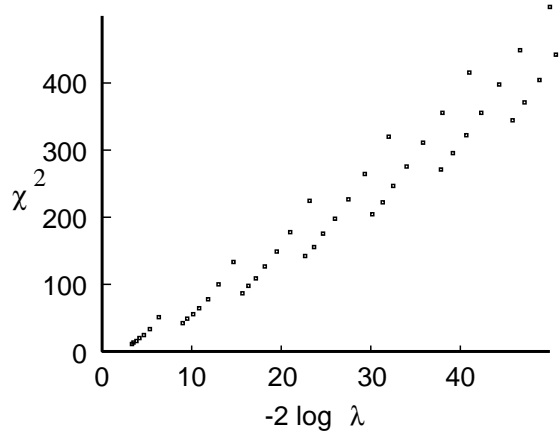}
\caption[Pearson's $\chi^2$ vastly overstates significance of rare events]{Pearson's $\chi^2$ vastly overstates significance of rare events.
Here, $p=0.01$, $n_1=100$, and $n_2=10,000$.  Note the change of
vertical scale with respect to previous
figure. \label{fig:overestimate}}
\end{center}
\end{figure}

Figure \ref{fig:overestimate}, on the other hand, shows the divergence
between Pearson's statistic and the log-likelihood ratio when $p =
0.01$, $n_1 = 100$ and $n_2 = 10,000$.  Here, the normal approximation
is not justified.  Note the large change of scale on the vertical axis
relative to the previous figure.  The pronounced disparity occurs when
$T(\sigma_1, S_1)$ is larger than the value expected based on the
observed value of $T(\sigma_1, S_2)$.  Unfortunately, the case where
$n_1 < n_2$ and ${T(\sigma_1, S_1) / n_1} > {T(\sigma_1, S_1) / n_2}$
is exactly the case of most interest in many text analyses.

\begin{figure}[htb]
\begin{center}
\includegraphics[]{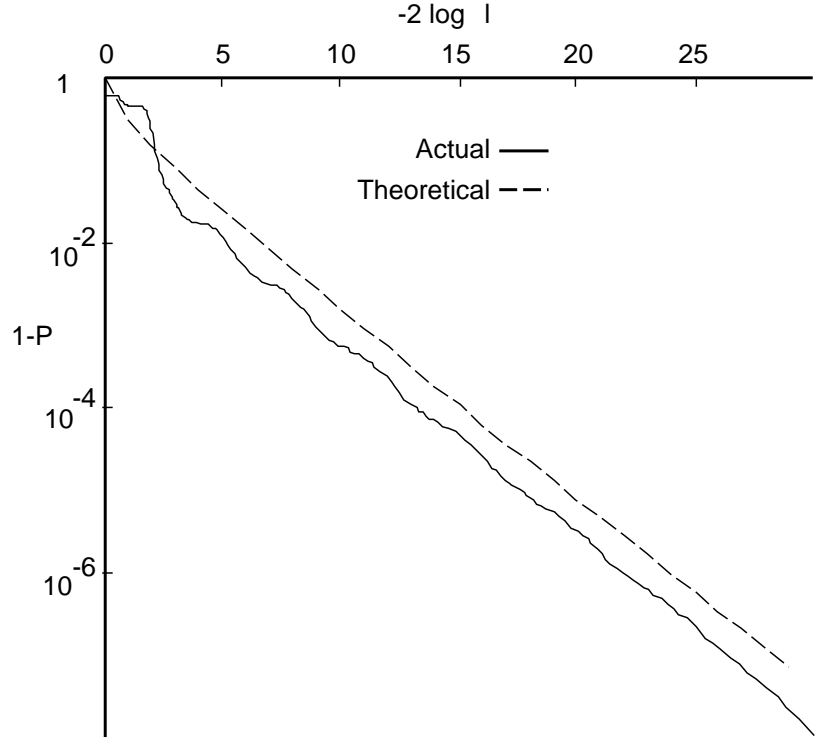}
\caption[Theoretical versus actual residual distribution of log-likelihood
ratio]{Theoretical versus actual residual distribution of log-likelihood
ratio.  The dashed line represents the theoretical distribution.  The
solid line was approximated by directly computing the probability of a
large sampling of possible outcomes.\label{fig:agreement}}
\end{center}
\end{figure}

The convergence of the log of the likelihood ratio to the asymptotic
distribution is demonstrated dramatically in Figure
\ref{fig:agreement}.  In this figure, the straighter line was computed
using numerical software taken from \cite{recipes} and represents the
idealized one degree of freedom cumulative $\chi^2$ distribution.  The
rougher curve was computed by a numerical experiment in which all
possible results were tested and their probability of occurrence
computed assuming binomial distributions defined by $p = 0.01$, $n_1 =
100$ and $n_2 = 10,000$.  This is the same case as was illustrated in
Figure \ref{fig:overestimate}.  The close agreement shows that the
likelihood ratio measure produces accurate results over 6 decades of
significance.  Accurate results are produced well into the range where
the Pearson's $\chi^2$ statistic diverges radically from the ideal.
It should be noted that in general, the log-likelihood ratio test is
conservative in that it {\em underestimates} the significance of all
rare outcomes.  The way that the log-likelihood ratio test statistic
scales relative to significance is also a strong indication that this
statistic can be used as a measure of dependency as well as a test.
Another reason for the reliability of this use is the fact that the
log-likelihood ratio test statistic is proportional to average mutual
information which has known desirable properties as a measure of
dependency \cite{cover+thomas}.

\subsection{Comparison with Mutual Information}

The log-likelihood ratio statistic for multinomials and average mutual
information are closely related mathematically.  If we return to the
expression for the log-likelihood ratio given earlier,
\begin{equation*}
- 2 \log \lambda = 2 \sum_{ij} T(\sigma_i,S_j)
\log {\frac
    {T(\sigma_i,S_j) \left[ \sum_{uv} T(\sigma_u, S_v) \right]}
    {\left[\sum_v T(\sigma_i, S_v) \right] \left[ \sum_u T(\sigma_u, S_j) \right]}
}
\end{equation*}
We can replace the row and column sums with equivalent expressions in
terms of the total count and maximum likelihood estimators for the
cell, row and column probabilities: $N = {\sum_{uv} T(\sigma_u,
S_v)}$, $\zeta_{ij} = T(\sigma_i,S_j)/N$, $\mu_j = \sum_u
T(\sigma_u,S_j)/N$ and $\nu_i = \sum_v T(\sigma_i,S_v) / N$.  This gives
\begin{equation*}
- 2 \log \lambda = 2 \sum_{ij} T(\sigma_i,S_j)
\log {\frac
    {T(\sigma_i,S_j) N}
    {\mu_j \nu_i N^2}
}
\end{equation*}
After cancellation and regrouping, this becomes
\begin{equation}
- 2 \log \lambda = 2 N \left[
\sum_{ij} \zeta_{ij} \log \zeta_{ij}-
\sum_{j} \mu_j \log \mu_j -
\sum_{i} \nu_i \log \nu_i
\right]
\end{equation}
This last is just $2N$ times the average mutual information.  Average
mutual information is commonly computed using $\log_2$ to give results
in bits while the log-likelihood ratio test is computed using $\log_e$
to give results which are $\chi^2$ distributed, but this difference
merely results in a constant factor $\log_2 e$.

It is important to note that the relationship between the
log-likelihood ratio test and single cell mutual information is not
nearly so simple.  This is because the log-likelihood ratio score has
symmetry properties which are not shared by single-cell mutual
information.

In practice, the log-likelihood ratio test is most appropriate for use
as a selection filter which selects multiple cases to find those which
show significant relationships.  Since selection is the central
difficulty in many situations, average mutual information is of
somewhat lower general utility.

\subsection{Comparison with Minimum Description Length \label{methods:mdl}}

Minimum description length (MDL) methods are essentially a statistical
formalization of Occam's razor.  Succinctly stated, the MDL principle
states that given a set of observed data and a set of models to choose
from, the model which provides the minimum encoded size for both
observed data and the model together is the best description of the
data.  More colloquially, the theory that allows us to explain our
observations together with the theory most succinctly is the best
theory.  The statistical theory of MDL methods owe much of their
underpinnings to algorithmic complexity theory as developed by Chaitin
\cite{Chaitin87}, Kolmogorov \cite{Kolmogorov65} and others.  One of
the first statistical applications similar to MDL was found in the
work of Akaike \cite{akaike73}, but until Rissanen's work
\cite{rissanen} the implications of the link between information
theory and MDL methods were not fully developed.  Wallace and others
developed the very closely related Minimum Message Length (MML) method
considerably earlier \cite{wallace} apparently independent of the inspiring influence of Solomonoff \cite{solomonoff}.

Interpreting the basic statement of the MDL principle mathematically
requires some additional machinery for linking the encoding of
observed data and model with probabilistic concepts.  This machinery
is provided by information theory and arithmetic coding which together
provide lower bounds on the most compact encoding of symbolic
sequences as well as an existence proof which shows that we can reach
those bounds.  Conventionally, it is also assumed that the observed
data are only known to a finite accuracy so that discrete
distributions can be used.  In most applications, we need not actually
encode the data and model since compression is not really our aim.
Instead, we merely need to find the model and parameters which would
provide the most compact encoding were the encoding ever to be done.

The minimum description length principle is very similar to Fisher's
maximum likelihood principle \cite{Fisher:1922}.  The maximum
likelihood principle states that the best values for the parameters of
a model are those which give the highest likelihood for the observed
data.  As an example, let us assume that the observed data consists of
a set $X = \left\{ x_i \mid i=1..n \right\}$ of successive independent
values of a discrete random variable which has distribution $p_\sigma$
where $\sigma$ and $x_i$ are from the alphabet $\Sigma$.  We wish to
find the particular values of model parameters $\omega$ which makes
$P(\sigma \mid \omega)$ most like $p_\sigma$.  By the maximum
likelihood principle, we would choose the values of $\omega$ in order
to maximize the likelihood of the observed values.  This maximum
likelihood estimator for the model parameters $\omega$ is
\begin{align*}
\hat \omega = \argmax_\omega \prod_i p(x_i \mid \omega)
\intertext{Because the $\log$ function is monotonic, this is equivalent to}
\hat \omega = \argmax_\omega \sum_i \log p(x_i \mid \omega)
\end{align*}

The expected value of the last sum is $n \sum_{\sigma \in \Sigma}
p_\sigma \log p(\sigma \mid \omega)$ and by the law of large numbers,
as $n$ grows, the value of the sum above will approach the expected
value.

There is a well-worn inequality attributed to Gibbs which can help
here.  If both $p_i$ and $q_i$ are distributions then $\sum_i p_i \log
q_i$ is maximized exactly and uniquely where $p_i = q_i$ for all $i$.
Being a distribution means that $\sum_i p_i = \sum q_i = 1$, $p_i \ge 0$
and $q_i \ge 0$.

Without loss of generality, we can assume $p_i > 0$.  Now, note that $\log x
\le x-1$ and so also
\begin{align*}
\sum_i p_i \log q_i - \sum_i p_i \log p_i &= \sum_i p_i \log \frac {q_i} {p_i} \\
& \le \sum_i p_i \left( \frac {q_i} {p_i} - 1 \right)  \\
& \le \sum_i q_i - \sum_i p_i = 0 \\
\end{align*}
The maximum is attained only when $p_i = q_i$ because $\log x = x-1$
only where $x = 1$.  If $p_i \ne q_i$ were ever true for any $i$, then
equality could never be regained.

This means that, in the limit of large samples, the maximum likelihood
principle allows us to find the model parameters $\hat \omega$ which
make our model most like reality.  Note, however, that by the
Shannon's Noiseless Channel coding theorem\cite{shannon48,shannon49},
\begin{equation*}
-\sum_i p(x_i) \log p(x_i \mid \omega)
\end{equation*}
is the expected encoded length of the sequence $\left< x_1 \ldots x_n
\right>$ if we use an arithmetic coder with model $p(x_i \mid
\omega)$.  Furthermore, this is the best expected length that can be
achieved by any coding system using this model.

This means that the maximum likelihood principle is equivalent to
finding the model which lets us encode observed data most compactly.
This is guaranteed to be the optimal model in the limit of large
samples both in terms of minimum encoding length as well as in terms
of fit to the actual distribution.  With smaller samples, however,
there is a possibility that over-fitting can occur; the apparently
optimal model may model the observed data perfectly, but then fail to
model new data from the same distribution.  This problem is
particularly acute when the family of models under consideration vary
in complexity.  In such a case, complex models can often exactly
reproduce the training data, but then utterly fail to generalize to
novel data.

The MDL principle extends the maximum likelihood principle by adding
another term in addition to the cost of encoding the observations.
This additional term is the cost of encoding the model.  This is
equivalent to following the same chain of reasoning as with the
maximum likelihood principle, but taking as a starting point the
maximization of the posterior likelihood instead of the conditional
likelihood.  This means we start with
\begin{align*}
\hat \omega &= \argmax_\omega \prod_i p(x_i \mid \omega) \,\, p(\omega) \\
\intertext{then after following the same chain of reasoning, the result is}
\hat \omega &= \argmax_\omega \log \prod_i p(x_i \mid \omega) \,\, p(\omega) \\
&\approx \argmax_\omega \left[ 
\sum_{\sigma \in \Sigma} p_\sigma \log p(\sigma \mid \omega) + \log p(\omega) 
\right]
\end{align*}

Maximizing this last expression is exactly the same as minimizing the
expected minimum encoded length of the observed data as well as the
minimum encoded length of the model parameters themselves.  It is
important to note that this minimum encoded length cannot be found by
minimizing the encoded length separately from the encoded length of
the model.  Instead, the tradeoff between data length and model length
must be considered.  It is common in the work on MDL to use the
so-called Levin prior to compute the model length.  This prior is
based on the Kolmogorov complexity of the model parameters
\cite{vitanyi,vapnik}.

This is the point where the deep connection between the log-likelihood
ratio and MDL methods can be found.  Referring back to the beginning
of section \ref{methods:lr_def}, we effectively had two models based
on the two different allowable domains for $\omega$.  We restate these
two models slightly by defining $p_0(\omega)$ to be the prior
distribution in the case of the null hypothesis
\begin{equation*}
p_0(\omega) = \begin{cases}
p(\omega)/Z & \text{if $\omega \in \Omega_0$} \\
0 & \text{otherwise}
\end{cases}
\end{equation*}
where $Z = {\int_{\Omega_0} d P(\omega) }$ is a constant
normalization factor.

Now, the minimum description length for the unrestricted case will be
approximately,
\begin{align*}
MDL = - \max_{\omega} \,\, \log p(X \mid \omega) \,\, p(\omega) \\
\intertext{and for the null hypothesis, this will be}
MDL_0 = - \max_{\omega} \,\, \log p(X \mid \omega) \,\, p_0(\omega)
\end{align*}

By definition, the dimension of the set $\Omega$ is $D_\Omega =
- \lim_{\delta \rightarrow 0} \log R(\delta)/\log \delta$ where
$R(\delta)$ is the minimum cardinality of any discretization of
$\Omega$ which has maximum diameter $\delta$.  If we assume uniform
probability for each element in $\Omega$ then this cardinality of the
discretization of $\Omega$, $R(\delta)$, gives an upper bound on the
number of bits required to encode any point in $\Omega$ to within an
accuracy $\delta$.  Thus, in the limit of small $\delta$, we would
require at most $\log_2 R(\delta) = -D_\Omega \log_2 \delta$ bits to
encode any point in $\Omega$.  Similarly, for any point in the null
hypothesis, $\Omega_0$, we would require at most $\log_2 R_0(\delta) = -
D_{\Omega_0} \log_2 \delta$ bits where $R_0(\delta)$ is the
corresponding minimum cardinality of any discretization of $\Omega_0$
with diameter $\delta$.

Clearly this argument assumes that the diameter of $\Omega$ is finite,
but extension to the infinite case is also possible.  Li and Vitanyi,
for example, present a considerably more general argument than the one
given here \cite{vitanyi}.  

Continuing with our assumption of a uniform prior, we find
that the difference in the encoded size of the two models for a
uniform prior is 
\begin{equation}
\Delta MDL = \max_{\omega \in \Omega} \,\, \log p(X \mid \omega) 
- \max_{\omega \in \Omega_0} \,\, \log p(X \mid \omega)
+ (D_\Omega-D_{\Omega_0}) \log \delta
\end{equation}

There are two interesting similarities that we should note here.
First, this formula is essentally the same as Akaike's test statistic
\cite{akaike73}.  Also, the first two terms are equivalent
to the log-likelihood ratio test statistic.

We know, however, that if the assumptions of the log-likelihood ratio
test hold, then the difference of the first two terms above will be
$\chi^2$ distributed with $D_\Omega - D_{\Omega_0}$ degrees of
freedom.  The MDL criterion for selecting the unrestricted or the null
hypothesis with uniform prior is thus identical to doing a
log-likelihood ratio test with a cutoff which is a constant times the
mean of the expected distribution.  This constant is determined
primarily by the discretization of the hypothesis space used in the
MDL test.

The log-likelihood ratio test is thus a specialization of the MDL
principle.  The benefit of the log-likelihood ratio test is that we
gain some understanding of the expected distribution of the resulting
test statistic.  The benefit of the fully generalized MDL method is
that we can handle much more complex situations than will fit into the
restrictive mold of the log-likelihood ratio.

There are also situations where a log-likelihood ratio test can be
used as an approximation of a more correct MDL treatment.  In these
situations, the computational cost of the log-likelihood ratio can be
several orders of magnitude less than the cost of using MDL.  Thus,
the log-likelihood ratio statistic can be an attractive heuristic
stand-in for the full MDL methods.

\part{Applications}
\chapter{Collocation and Coocurrence\label{colloc}}

\section{Overview}

The terms collocation and coocurrence are, at times, used somewhat
interchangeably in the literature.  For the purposes of this chapter,
however, the term collocation is taken to mean the situation where two
words occur in direct juxtaposition, while coocurrence is taken to be
the case when two words both occur within some predefined textual
unit.  Commonly used textual units for coocurrence analysis include a
fixed number of words, a sentence, a paragraph or an entire documents.
Under these definitions, collocations are clearly a special case of
word coocurrence.  Some authors extend the term collocation to denote
cases where textual juxtaposition may not occur; in this thesis, such
usage is avoided.

In a data-driven or corpus-based study of language, the starting point
is generally to consider text to be a sequence of words.  The first
and most obvious step in the statistical description of text is to
determine the average frequency of the individual words in text and
the rate of introduction of novel words.  Even the simplest analysis
shows, however, that there are both microscopic and macroscopic
variations in the actual frequency of occurrence of words. It is
natural to turn to the analysis of the effect of collocation and
coocurrence on these average word frequencies as a next step in the
statistical analysis of language.

To some degree, the sequence of analytic steps which moves from words
to structure recapitulates the evolution of traditional linguistic
study.  The major difference is that the traditional approach has been
to emphasize recognition of relatively complex syntactic forms, while
the statistical approach tends to emphasize the estimation of the
frequency of relatively more simple forms.  

From the perspective of, say, a follower of Chomsky, this focus on the
frequency of apparently trivial phenomena might seem absurd.  Instead,
it would seem infinitely more productive to move immediately to the
description and cataloging of the complex structures which result from
the introspective analysis of language.  Viewed from a data-driven
perspective, however, such a step appears to introduce a dangerous new
element of variability, since the production of the linguistic
structures is yet another psychological phenomenon which must be
modeled.  This multiplication of entities can only be supported
insofar as the resulting analyses can describe language in a
scientifically defensible way.  The essence of such a scientific
defense is the quantitative comparison of different approaches, and
assuming that a syntactic description must be correct is equivalent
to donning an inferential straight-jacket because so many alternative
descriptions are eliminated {\em ab initio} by such an assumption.

It is easy to create a false opposition between syntactic and
statistical or data-driven methods for analysing language.  The
difference here is that strictly syntactic approaches start with the
assumption that the only acceptable representation for theories of
language is a particular syntactic formalism.  Commonly, the next step
is not a critical evaluation of the degree to which such an assumption
describes the language of interest in any large sense.  The
philosophical starting point for statistical language analysis is, on
the other hand, an assumption that any model of language must provide
a better fit to a wide range of empirical data than alternative
representations.  This focus on optimality and fitting to data makes
many forms of models difficult to apply, at least initially.  In the
work on statistical language analysis conducted through the 1980's,
there was a discernible lack of any syntactic content.  This lack
could and has been interpreted as a symmetrical rejection of syntactic
methods which mirrored the rejection of statistical methods by the
traditional linguistic community.  Such an interpretation could easily
encouraged by some of the apocryphal comments made during the time,
but it is probably more accurate to view the lack of syntactic
components in the statistical language analysis literature as a
natural consequence of the difficulties encountered in trying to apply
statistical methods to syntactic representations.  These difficulties
had much to do with lack of large scale training data, lack of
applicable mathematical and statistical theories, and lack of
(statistically derived) methods for lexical generalization.  These
difficulties had, in fact, relatively little to do with a categorical
rejection of any and all syntactic theories.
  
To escape from this syntactic straight-jacket and begin to derive
useful information from text corpora, collocation and coocurrence can
be used as a very simple, but quite informative, linguistic
microscope.  It can easily be shown that for very broad classes of
syntactic structure which might be present in text, coocurrence
analyses should be able to detect and quantify this structure.
Moreover, various forms of coocurrence analysis are able to compare
the contribution of various kinds of structure.  Traditionally, these
structures might be deemed to be syntactic or semantic, but these
labels are affixed by linguists and do not necessarily exist in the
data.

Not surprisingly, coocurrence analysis does show considerable
structure in text.  In the simplest case of collocation, a strong mix
of syntactic and semantic relationships is highlighted by the
statistical analysis.  As larger and larger coocurrence windows are
used, syntactic relationships become less and less apparent and
semantic relationships are all that remain.
  
As well as theoretical motivations for developing methods for
analysing the statistics of coocurrence phenomena, there are concrete
applications which can be facilitated by these methods of analysis.
For example, information retrieval systems can benefit from the
detection of sequences of words which have anomalous statistical
properties since these sequences are quite often phrasal units which
serve as terms of art in technical language.  In much the same way,
these methods can be used as part of statistically based systems for
segmentation of Asian language texts which typically do not make much
use of spaces to separate words.  These segmentation techniques can
help improve language models such as those used in speech recognition.

Just as interesting, although less applicable to the building of
software applications, is the provocative connection between text
coocurrence statistics and performance on cognitive tests.  Certain
statistical measures of anomalous coocurrence appear to correlate well
with human estimates of the relatedness of words.  It is well known
that relatedness estimates correlate very well with the degree of
priming in a variety of cognitive tasks.\footnote{Priming is the
phenomenon in which a cognitive task such as distinguishing words from
non-word character sequence is facilitated by the prior appearance of
related words.}  It would be quite reasonable to expect that the
structure of language and the priming effect are quite closely
related, and thus the known correlation between priming and
relatedness would imply that there should be a strong correlation
between concept relatedness and word coocurrence.  The exact order of
causation would not be important here; whether priming as a basic
cognitive phenomenon gave rise to the structure of language or whether
the basic structure of language induces the priming effect, the
fundamental implications of the correlation are still quite exciting.

\section{Background}

There have been a substantial number of efforts to produce a useful
test for screening for words which appear near each other anomalously
often.  Most work has used heuristic measures of association such as
the Dice or Jaccard measures (see section
\ref{methods:standard_tests}).  In general, there has been little
appreciation in the literature of the difference between measures of
association and statistical tests of significant association.  In
fact, the most desirable sort of measure for searching out
collocations is clearly a test which flags anomalous collocation.  An
additional requirement is that collocation tests must deal well with
low count situations, especially when small corpora are being
analysed.  Tests such as the likelihood ratio test fit these
requirements very well.

A number of researchers have looked at methods for detecting
collocation and related phenomena.  Francis and Kucera used a $\chi^2$
test to examine the association between genre and word frequency
\cite{kucera-francis-67}.  This association is clearly very long
range.  The $\chi^2$ test was useful for them because so many
instances of the words under analysis are available in each genre.  In
more traditional collocation studies, the typical number of instances
drops by many orders of magnitude and $\chi^2$ becomes much less
useful.

Plate examined a large number of measures of association
\cite{pathfinder} but did not directly address the question of
statistical significance.  His efforts at finding measures which would
work with words which have widely differing baseline frequencies were
clearly attempting to address the problem that statistical measures
solve very nicely.

Perhaps better known in the computational linguistic community is work
of Church, Gale and Hanks \cite{church89}.  This work focussed on the
use of single-cell mutual information as a measure of association.  In
order to address the problem of significance, two measures were taken.
First, all pairs which had low rates of occurrence were eliminated,
and secondly a test statistic (called a $Z$-score by the authors) was
developed which attempted to deal with the issue of significance.
Unfortunately, this score was dependent on an assumption of normal
distribution which is not generally valid.  When the assumption is
violated, substantial error is possible.  This potential for error
limits the applicability of Church's measure for low frequency
situations.

Somewhat later, the work in this chapter was described in the
literature \cite{dunning93}.  At about the same time Smadja developed
methods which at their core use the Dice coefficient \cite{Smadja93}.

Daille \cite{Daille:95} pulled much of this work together by actually
testing the word pairs discovered by a variety of scoring techniques
and evaluating these word pairs by having humans rate how tightly the
words were connected.  Daille reported that the likelihood ratio test
produced the closest match to human responses.

\section{Experimental Methods}

To demonstrate the efficacy of the likelihood ratio based methods, an
analysis was made of a 30,000 word sample of text obtained from the
Union Bank of Switzerland.  The intention was to find pairs of words
which occurred next to each other with a significantly higher
frequency than would be expected based on the word frequencies alone.
The text was 31,777 words of financial text largely describing market
conditions for 1986 and 1987.  This is a very small text for this sort
of analysis.  For general studies, much larger texts are available,
but it is often desirable to be able to analyse very small samples in
order to highlight the peculiarities of the samples.

The results of such a bigram analysis should highlight collocations
common in English as well as collocations peculiar to the financial
nature of the analysed text.  As will be seen, the ranking based on
likelihood ratio tests does exactly this.  Similar comparisons made
between a large corpus of general text and a domain-specific text can
be used to produce lists consisting only of words and bigrams
characteristic of the domain-specific texts.

This comparison was done by creating a contingency table which
contained the following counts of each bigram that appeared in the
text.  Table \ref{colloc:sample} illustrates such a contingency table.  
In this table, $A B$ represents the bigram in which the first word is
$A$ and the second is word $B$.  $A \lnot B$ represents
the bigram in which the first word is $A$ while the second word is
any word other than $B$.

\begin{table}[ht]
\begin{center}
\begin{tabular}[ht]{|l|l|}
\hline
$T(A B)$ & $T( \lnot A B)$ \\
\hline
$T(A \lnot B)$ & $T( \lnot A \lnot B)$ \\
\hline
\end{tabular}
\end{center}
\caption{The arrangement of bigram counts in a contingency table \label{colloc:sample}}
\end{table}

If the words $A$ and $B$ occur independently, then we would expect
$p(AB) = p(A) p(B)$ where $p(AB)$ is the probability of $A$ and $B$
occurring in sequence, $p(A)$ is the probability of $A$ appearing in
the first position and $p(B)$ is the probability of $B$ appearing in
the second position.  We can cast this into the mold of the binomial
analysis described in section \ref{methods:lr_tests} by phrasing the
null hypothesis that $A$ and $B$ are independent as $p(A \mid B) = p(A
\mid \lnot B)$.  This means that testing for the independence of $A$
and $B$ can be done by testing to see if the distribution of $A$ given
that $B$ is present (the first row of the table) is the same as the
distribution of $A$ given that $B$ is not present (the second row of
the table).  In fact, of course, we are not really doing a statistical
test to see if $A$ and $B$ are independent; we know that they are
generally not independent in text.  Instead we just want to use the
test statistic as a measure which will help highlight particular $A$'s
and $B$'s which are highly associated in text.  Section
\ref{methods:chi2_comparison} provides a justification of
this use.  The characteristics of the comparison in this section is
typical of the regime under which the log-likelihood ratio test is
used in finding collocations.

These counts were analysed using the test for binomials described in
section \ref{methods:lr_tests}, and the 50 most significant are
tabulated in Table \ref{colloc:lr_table}.  This table contains the
most significant 30 bigrams and is reverse sorted by the first column
which contains the quantity $-2 \log \lambda$.  Other columns contain
the four counts from the contingency table described above, and the
bigram itself.

\begin{table}[htb]
\scriptsize
\begin{center}
\begin{tabular}[htb]{|lllll|rl|}
\hline
$LR score$ & $T(AB)$ & $T(A \lnot B)$ & $T(\lnot A B)$ & $T(\lnot A \lnot B)$ & $A$ & $B$ \\
\hline
270.72 &  110 & 2442 &  111 & 29114 &             the & swiss\\         
263.90 &   29 &   13 &  123 & 31612 &             can & be\\            
256.84 &   31 &   23 &  139 & 31584 &        previous & year\\          
167.23 &   10 &    0 &    3 & 31764 &         mineral & water\\         
157.21 &   76 &  104 & 2476 & 29121 &              at & the\\           
157.03 &   16 &   16 &   51 & 31694 &            real & terms\\         
146.80 &    9 &    0 &    5 & 31763 &         natural & gas\\           
115.02 &   16 &    0 &  865 & 30896 &           owing & to\\            
104.53 &   10 &    9 &   41 & 31717 &          health & insurance\\     
100.96 &    8 &    2 &   27 & 31740 &           stiff & competition\\   
 98.72 &   12 &  111 &   14 & 31640 &              is & likely\\        
 95.29 &    8 &    5 &   24 & 31740 &       qualified & personnel\\     
 94.50 &   10 &   93 &    6 & 31668 &              an & estimated\\     
 91.40 &   12 &  111 &   21 & 31633 &              is & expected\\      
 81.55 &   10 &   45 &   35 & 31687 &               1 & 2\\             
 76.30 &    5 &   13 &    0 & 31759 &         balance & sheet\\         
 73.35 &   16 & 2536 &    1 & 29224 &             the & united\\        
 68.96 &    6 &    2 &   45 & 31724 &        accident & insurance\\     
 68.61 &   24 &   43 & 1316 & 30394 &           terms & of\\            
 61.61 &    3 &    0 &    0 & 31774 &           natel & c\\             
 60.77 &    6 &   92 &    2 & 31677 &            will & probably\\      
 57.44 &    4 &   11 &    1 & 31761 &           great & deal\\          
 57.44 &    4 &   11 &    1 & 31761 &      government & bonds\\         
 57.14 &   13 &    7 & 1327 & 30430 &            part & of\\            
 53.98 &    4 &    1 &   18 & 31754 &           waste & paper\\         
 53.65 &    4 &   13 &    2 & 31758 &         machine & exhibition\\    
 52.33 &    7 &   61 &   27 & 31682 &            rose & slightly\\      
 52.30 &    5 &    9 &   25 & 31738 &       passenger & service\\       
 49.79 &    4 &   61 &    0 & 31712 &             not & yet\\           
 48.94 &    9 &   12 &  429 & 31327 &        affected & by\\            
 48.85 &   13 & 1327 &   12 & 30425 &              of & september\\     
 48.80 &    9 &    4 &  872 & 30892 &        continue & to\\            
 47.84 &    4 &   41 &    1 & 31731 &               2 & nd\\            
 47.20 &    8 &   27 &  157 & 31585 &     competition & from\\          
 46.38 &   10 &  472 &   20 & 31275 &               a & positive\\      
 45.53 &    4 &   18 &    6 & 31749 &             per & 100\\           
 44.36 &    7 &    0 & 1333 & 30437 &          course & of\\            
 43.93 &    5 &   18 &   33 & 31721 &       generally & good\\          
 43.61 &   19 &   50 & 1321 & 30387 &           level & of\\            
 43.35 &   20 & 2532 &   25 & 29200 &             the & stock\\         
 43.07 &    6 &  875 &    0 & 30896 &              to & register\\      
\hline
\end{tabular}
\end{center}
\caption{Significant bigrams as detected by the log-likelihood ratio
test \label{colloc:lr_table}}  
\end{table}

Examination of the table produced using the log-likelihood ratio test
shows that there is good correlation with intuitive feelings about how
natural the bigrams in the table actually are.  This match with
intuition is important in applications where a statistical test is to
be used as a surrogate for human judgement.  This agreement with
intuition is in distinct contrast with Table \ref{colloc:chi2} which
contains the same data except that the first column is computed using
Pearson's $\chi^2$ test statistic.  The over-estimate of the
significance of items that occur only a few times is dramatic.  In
fact the entire first portion of the table is dominated by bigrams
rare enough to occur only once in the current sample of text.  The
misspelling in the bigram `sees posibilities' is in the original text.

\begin{table}[htb]
\scriptsize
\begin{center}
\begin{tabular}[htb]{|lllll|rl|}
\hline
$\chi^2$ & $T(AB)$ & $T(A \lnot B)$ & $T(\lnot A B)$ & $T(\lnot A \lnot B)$ & $A$ & $B$ \\
\hline
  31777.00 &    3 &    0 &    0 & 31774 &           natel & c               \\
  31777.00 &    1 &    0 &    0 & 31776 &           write & offs            \\
  31777.00 &    1 &    0 &    0 & 31776 &            wood & pulp            \\
  31777.00 &    1 &    0 &    0 & 31776 &          window & frames          \\
  31777.00 &    1 &    0 &    0 & 31776 &      upholstery & leathers        \\
  31777.00 &    1 &    0 &    0 & 31776 &         surveys & expert          \\
  31777.00 &    1 &    0 &    0 & 31776 &            sees & posibilities    \\
  31777.00 &    1 &    0 &    0 & 31776 &     practically & drawn           \\
  31777.00 &    1 &    0 &    0 & 31776 &         poultry & farms           \\
  31777.00 &    1 &    0 &    0 & 31776 &     physicians' & fees            \\
  31777.00 &    1 &    0 &    0 & 31776 &          paints & varnishes       \\
  31777.00 &    1 &    0 &    0 & 31776 &        maturity & hovered         \\
  31777.00 &    1 &    0 &    0 & 31776 &     listeriosis & bacteria        \\
  31777.00 &    1 &    0 &    0 & 31776 &              la & presse          \\
  31777.00 &    1 &    0 &    0 & 31776 &        instance & 280             \\
  31777.00 &    1 &    0 &    0 & 31776 &            cans & casing          \\
  31777.00 &    1 &    0 &    0 & 31776 &          bluche & crans           \\
  31777.00 &    1 &    0 &    0 & 31776 &            a313 & intercontinental \\
  24441.54 &   10 &    0 &    3 & 31764 &         mineral & water           \\
  21184.00 &    2 &    0 &    1 & 31774 &         scanner & cash            \\
  20424.86 &    9 &    0 &    5 & 31763 &         natural & gas             \\
  15888.00 &    1 &    1 &    0 & 31775 &          suva's & responsibilties \\
  15888.00 &    1 &    1 &    0 & 31775 &          suva's & questionable    \\
  15888.00 &    1 &    1 &    0 & 31775 &     responsible & clients         \\
  15888.00 &    1 &    1 &    0 & 31775 &             red & ink             \\
  15888.00 &    1 &    1 &    0 & 31775 &          joined & forces          \\
  15888.00 &    1 &    1 &    0 & 31775 &         highest & density         \\
  15888.00 &    1 &    1 &    0 & 31775 &      generating & modest          \\
  15888.00 &    1 &    1 &    0 & 31775 &         enables & conversations   \\
  15888.00 &    1 &    1 &    0 & 31775 &         dessert & cherry          \\
  15888.00 &    1 &    1 &    0 & 31775 &    consolidated & lagging         \\
  15888.00 &    1 &    1 &    0 & 31775 &       catalytic & converter       \\
  15888.00 &    1 &    1 &    0 & 31775 &           bread & grains          \\
  15888.00 &    1 &    1 &    0 & 31775 &     bottlenecks & booking         \\
  15888.00 &    1 &    1 &    0 & 31775 &        bankers' & association's   \\
  15888.00 &    1 &    1 &    0 & 31775 &       appenzell & abrupt          \\
  15888.00 &    1 &    1 &    0 & 31775 &              56 & 513             \\
  15888.00 &    1 &    1 &    0 & 31775 &              56 & 082             \\
  15888.00 &    1 &    1 &    0 & 31775 &              46 & 520             \\
  15888.00 &    1 &    1 &    0 & 31775 &              43 & classified      \\
  15888.00 &    1 &    1 &    0 & 31775 &              43 & 502             \\
\hline
\end{tabular}
\end{center}
\caption{Significant bigrams as detected by Pearson's $\chi^2$ test \label{colloc:chi2}}  
\end{table}

Of course, a simple assertion that the contents of this table are
compatible with someone's (presumably the author's) intuitions may not
be particularly persuasive.  Since the study described in this chapter
was published \cite{dunning93}, Daille has provided more rigorous
evidence that the log-likelihood ratio test is indeed more compatible
with human judgement than a variety of other measures.  Daille's work
showed that the log-likelihood ratio test was the most compatible with
human judgments of all of the tests that she could find.

\begin{table}[htb]
\scriptsize
\begin{center}
\begin{tabular}[htb]{|lllll|rl|}
\hline
$\chi^2$ & $T(AB)$ & $T(A \lnot B)$ & $T(\lnot A B)$ & $T(\lnot A \lnot B)$ & $A$ & $B$ \\
\hline
525.02 & 110 & 2442 &  111 & 29114 & the       & swiss     \\
286.52 &  76 &  104 & 2476 & 29121 & at        & the       \\
 51.12 &  26 & 2526 &   66 & 29159 & the       & volume    \\ 
  6.03 &   4 &  148 & 2548 & 29077 & be        & the 	   \\
  4.48 &   1 & 	 73 & 2551 & 29152 & months    & the 	   \\
  4.31 &   1 & 	 71 & 2551 & 29154 & increased & the 	   \\
  0.69 &   4 & 	 70 & 2548 & 29155 & 1986      & the 	   \\
  0.42 &   7 & 	 62 & 2545 & 29163 & level     & the 	   \\
  0.28 &   4 & 	 60 & 2548 & 29165 & again     & the 	   \\    
  0.12 &   5 & 2547 &   67 & 29158 & the       & increased \\
  0.03 &  18 &  198 & 2534 & 29027 & as        & the       \\
\hline
\end{tabular}
\end{center}
\caption{Bigrams where Pearson's $\chi^2$ test is applicable \label{colloc:applicable}}
\end{table}

Out of 2693 bigrams analysed, fully 2682 of them fall outside the
scope of applicability of the normal $\chi^2$ test.  The 11 bigrams
which were suitable for analysis with the $\chi^2$ test are listed in
Table \ref{colloc:applicable}.  It is notable that all of these
bigrams contain the word {\em the} which is the most common word in
English.  Thus, if the $\chi^2$ test is limited to the cases where do
not violate the assumptions involved in its derivation, then it gives
reasonable results, but the resulting scope of applicability may be
unreasonably narrow.

\section{Conclusions}

The results here demonstrate clearly that the log-likelihood ratio
test substantially outperforms the more traditional $\chi^2$ test.
This advantage has two facets.  If the $\chi^2$ test is only used
where the requisite minimum expected frequency condition is met, then
the $\chi^2$ can only be used on a small number of the cases where it
might be desired to use it (only 11 word pairs met this criterion in
this example).  Alternatively, if the minimum expected frequency
criterion is ignored, the $\chi^2$ test can produce very large values
in cases where a single coincidental juxtaposition occured.  Since the
minimum frequency requirements \cite{church89,ChurchHanks90} for
Church and Gale's use of single-cell mutual information is essentially
the same as for $\chi^2$, it is to be expected that their work would
suffer similar problems.

One interesting possibility for future work is to try to apply the
likelihood ratio test in other ways to detect interesting
collocations.  For example, in information retrieval applications,
phrases are not interesting as query terms unless using the phrase as
a phrase would have a different result from using the individual
words.  A combination approach might first filter word sequences using
a log-likelihood ratio test and then perform a statistical comparison
to see if the contexts in which statistically plausible phrases were
used were significantly different from the composite of the contexts
in which the constituents of the phrase were used.  One way to compare
these contexts would be to use a log-likelihood ratio test to compare
the frequency of words in the contexts to be compared.

A preliminary study of the efficacy of just such a test for context
shift is described in more detail in section \ref{word_pairs}.
\chapter{Document Retrieval \label{ir}}

\section{Overview \label{ir_overview}}

The likelihood ratio tests described in section
\ref{methods:lr_binomial} can be used as a basis on which to build a
document routing system.  A document routing system is a type of
document retrieval system that is given a number of sample documents
which satisfy a user's requirements.  The system then can
approximately determine whether new documents also satisfy the user's
needs.  Some routing systems can also make use of sample documents
which do {\em not} meet the user's needs, but many systems cannot make
effective use of such negative information.

This chapter describes a prototype document routing system known as
Luduan\footnote{The Luduan system is named after a mythical Chinese
creature which sat beside the emperor's throne and could distinguish
good from evil.} which is similar to many currently available research
and commercial systems in that a query is expressed as a set of
weights attached to the presence of particular words.  A document is
routed by producing a score for that document relative to a query and
comparing the score so computed to the scores for other documents or
to a fixed threshold.  The terms in the query are selected by the
system based on training data in the form of an original query and
example relevant and non-relevant documents.  The Luduan system as
described here is somewhat unusual in the way that these query terms
are selected.  Term selection in Luduan is done by using a statistical
test to determine which words are peculiar to the positive training
documents when compared to the negative training documents.  This
selection method was first mentioned in the context of computational
linguistics in \cite{dunning93, dunning93a}; subsequent tests to
analyse and validate the method are reported here.  Many previous
systems such as the one described by Buckley \cite{buckley93} have
used raw frequency to select the words to be given non-zero weight and
have been largely unable to benefit from negative examples in the
selection of query terms.  

Most current systems which use judged negative examples use the method
introduced by Rocchio \cite{frakes} and by Wu and Salton
\cite{wu/salton} in which a linear combination of query vectors
derived from positive and negative evidence is used to form a new
query vector.  Since rare words will have large weights in most of the
term weighting schemes used with Rocchio's method, the proportionally
larger variation in the small number of times these words appear in
relevant and non-relevant training documents can cause large and
unwanted deviation in the resulting query vector.  Rocchio's method
and similar vector subtraction systems provide no way to ignore
variations in term frequency which are due to mere chance.  Luduan, in
contrast, compares raw word frequencies in judged relevant and judged
non-relevant documents to derive a robust set of query terms to be
used in a traditional weighted-term retrieval system.  The robustness
of this set of query terms is due to the desirable properties of the
log-likelihood ratio test when applied to the analysis of contingency
tables with small counts.  This robustness is demonstrated in practice
by the results reported in this chapter and by manual inspection of
the queries derived by Luduan.

The unusual query term selection strategy used in Luduan results in a
considerable improvement in performance.  Luduan's query selection
depends critically on the way that the log-likelihood ratio test can
handle small counts gracefully without producing excessively large
scores when a word appears a very few times in the relevant examples
and not at all in the non-relevant examples.  A demonstration of the
performance of the log-likelihood ratio test relative to Pearson's
$\chi^2$ test is given in section \ref{methods:chi2_comparison}.  By
analysing such situations accurately, Luduan is apparently able to
avoid the introduction of terms into the resulting query due to
statistical fluctuation in small counts.

Somewhat after the publication of the original papers describing the
test used by Luduan for the comparison of word frequencies, the TREC-3
paper by Cooper and others at Berkeley \cite{berk_trec3} reported the
test of a system based on the use of a traditional $\chi^2$ test for
term selection.  Their term selection test is similar to the one
mentioned in the papers referred to earlier \cite{dunning93} and
\cite{dunning93a} and described in detail and tested here.  Their
system differed significantly in several aspects, however, from the
Luduan system.  These differences appear to have masked the
potential performance improvement in their results.  In particular,
the Berkeley system used a traditional $\chi^2$ test instead of a
log-likelihood ratio test.  This may have caused problems since many
 useful terms occur only a few times which would have caused the use of
the $\chi^2$ test to be inappropriate.  They also used a very low
threshold for selecting terms.  This low threshold increased the
number of terms used by over an order of magnitude and probably
contributed to the over-fitting problem alluded to in their paper.
Luduan is effectively immune to over-fitting since it uses term weights
which depend only on the characteristics of the entire training corpus
and not on the training examples.  This non-adaptive weighting scheme
combined with the higher threshold for term selection results in much
better overall performance.

The Luduan system is crucially dependent on the ability to determine
which words occur {\em significantly} more or less often in the sample
training documents.  Differences which might have occurred by chance
or coincidence must be ignored if a useful query is to be obtained.
Furthermore, since most content bearing words are relatively rare, it
will generally be true that the number of occurrences of any
particular content word in the training documents will be quite small;
the test used to select terms must gracefully handle this situation.
Most weighting and term selection schemes which have been used in
information retrieval applications do not have this property of
graceful degradation.  In particular, Pearson's $\chi^2$ test is often
not applicable.

Another key feature of Luduan as evaluated in this chapter is that the
negative examples examined were documents which had been nominated as
potentially relevant by some document retrieval or routing system.
This means that the documents used as negative examples had some sort
of similarity to the original query or to other relevant documents.
Using this relatively focussed set of negative examples has the
advantage of highlighting the terms which distinguish truly relevant
documents.  

Luduan is able to use positive and negative examples, can deal with
small counts, and shows significant and substantial performance
benefits over benchmark research systems.  These performance benefits
are demonstrated by the results of the two experiments described in
this chapter.

\section {Background}

\subsection {Evaluation Data}
The evaluation of the Luduan system was conducted by using a subset of
the TREC retrieval corpus produced by the U.S. National Institute of
Standards and Technology.  The National Institute of Standards and
Technology (NIST, formerly National Bureau of Standards or NBS) is
part of the United States Department of Commerce.  It has a broad
mandate for technological development which includes developing
standards for the evaluation of information processing systems.  As
part of this effort, NIST organizes annual Text Retrieval and
Evaluation Conferences (TREC).  These conferences are focussed on the
evaluation of information retrieval systems with realistic amounts and
types of text.  The participants are given a common set of documents
and search topics and their results are compared.  This comparison is
done by pooling the top 1000 documents retrieved for each query by all
of the participating systems.  These pooled documents are then
evaluated by human analysts to determine their relevance to the query
used to find them.  All judgements made by the analysts are recorded.
Due to the unprecedented scale of the TREC evaluations, the resulting
combination of queries and corpus is the most comprehensive and useful
of any publicly available document retrieval evaluation suite.

In TREC, systems are primarily evaluated on an {\em ad hoc}
retrieval task and a document routing task.  In the {\em ad hoc}
retrieval, novel queries are developed each year and participants in
TREC return the documents found by their systems for evaluation.  In
the routing task, existing queries and previously judged documents are
used by the systems to find relevant documents in a set of previously
unseen documents.

Significant improvements in {\em ad hoc} document retrieval
and routing software technology have been achieved and demonstrated by
participants in TREC.  For example, in the TREC-2 evaluations, it was
shown for the first time that purely automatic systems can produce
document routing queries which considerably out-perform manually
produced queries \cite{harman93a}.  This record continues in more
recent TREC evaluations as described in \cite{harman95}.

Since the TREC evaluation corpus is so much more extensive than other
commonly available text retrieval test corpora and since the TREC
corpus is the only one which has well defined negative examples, the
TREC corpus is the only one used to evaluate the system described
here.  The fact that the negative examples were initially found using
automated retrieval systems is significant to the operation of Luduan.
Due to their origin, these negative examples are quite similar to the
relevant documents and thus can be used to highlight those features
which distinguish relevant documents from other very similar, but
non-relevant documents.  This similarity of the non-relevant documents
to relevant documents appears to be crucial to achieving high
performance since other previous efforts in which the positive
training examples were compared to all other documents did not produce
in particularly good results.

The widely available Reuters document classification test corpus
\cite{LEWIS91c,Apte94a} was not used to evaluate the Luduan system
since that corpus only provides positive examples for training.  Since
the Luduan system requires both positive and negative examples for
training, such a corpus would not be useful for comparison.

\subsection {Document Routing}
In the automatic document routing paradigm, the system is given input
consisting of a query, a set of positive example documents which have
been judged by a human to be relevant to the query, a set of negative
example documents which have been judged by a human not to be relevant
to the query as well as a large corpus of documents which have not
been judged.  Novel documents are presented to the system, and the
system must determine if these documents are relevant to the query.

In some experimental designs, the system is allowed to evaluate a
number of novel documents and produce a ranked list, while in other
designs, the system must make a binary decision of relevant or
non-relevant for each document.  In TREC, ranked lists are used,
although a binary decision specialty track has been offered in recent
conferences.  This difference has consequences in terms of system
design since in the binary classification task, the system must be
able to commit to one classification or the other.  For many systems,
this means that a threshold must be applied.  In the ranking system,
there is no need to define a specific threshold except in relation to
all other retrieved documents.  In practical systems, the distinction
between these two modes of operation is not so sharp since users do
not usually read documents as they arrive, but rather they
periodically read the backlog of documents which the system has marked
as relevant.  This means that a ranking system can accumulate a number
of documents and set a threshold very late in the process.
Alternatively, a ranking system can adaptively set a threshold based
on feedback on each document.  There is some indication in work at
Xerox Parc \cite{hull95} that thresholds can be set adaptively.  The
InRoute system also uses an adaptive threshold setting algorithm quite
successfully \cite{callan96a} as well as being able to estimate
various corpus statistics on the fly.  This ability to work
incrementally makes ranking systems essentially as good at the binary
task as would be optimistically predicted by their performance on a
ranking task.

Currently, the most successful document routing systems all use a
combination of term weights to produce a score for each incoming
document.  These systems include the SMART system from Cornell, the
Inquery system from the University of Massachusetts and the logistic
regression based systems from the University of California at Berkeley
as well as reduced dimensional vector systems such as the LSI system
from Bellcore and Convectis from HNC Software.  All of the systems
except for the reduced dimensional vector systems can be viewed as
having a term selection phase which can be separated from a
term-weight selection phase.

The Luduan system as described in the first experiment in this chapter
uses the log-likelihood ratio test described in section
\ref{methods:lr_binomial} to select terms by comparing word
frequencies in judged relevant documents and the judged non-relevant
documents and then uses the {\tt lnc.ltc} weighting scheme from the
SMART system to weight the selected terms with no stemming.  In the
second experiment, the term weighting was done by using the default
ranking system used by the MG \cite{mg} and InQuery \cite{callan96a}
systems.  None of the Luduan systems evaluated used any sort of
compound terms (phrases) although the Luduan term selection strategy
could accommodate compound terms relatively easily.

The term selection in Luduan is done by constructing a $2\times2$
contingency table for each term in any of the relevant examples.  The
rows of the table are formed by the counts of the term in the judged
relevant ($R$) and judged non-relevant ($N$) sets while the columns are
the counts of the term and all other terms in the set.  This is
illustrated in Table \ref{table:ir1}.

\begin{table}[ht]
\begin{center}
\begin{tabular}[ht]{r r@{} l c c c l@{} l}
&&& term $t$ & & other terms & &\\
\hhline{~~-----}
relevant && \vline& $T(t, R)$ & \vline& $T( \lnot t, R)$ &&\vline\\
\hhline{~~-----}
non-relevant && \vline&  $T(t, N)$ & \vline & $T( \lnot t, N)$ &&\vline\\
\hhline{~~-----}
\end{tabular}
\end{center}
\caption[Contingency table for term selection]{
The arrangement of counts in a contingency table to find whether term
$t$ should be included in a routing query. $T(t,X)$ denotes the number
of times that term $t$ occurs in the set of terms $X$ where $X$ can be
$R$ or $N$; $T( \lnot t, X)$ denotes the number of times that a term
other than $t$ occurs.  $R$ denotes the set of documents
which have been judged to be relevant and $N$ denotes the set of
documents judged to non-relevant.
\label{table:ir1}}
\end{table}

Only terms which had a log-likelihood ratio test statistic of 20 or
more were kept in the final query.  This threshold corresponds to a
significance level of approximately $10^{-5}$.  This is a considerable
contrast to the $0.05$ significance level used by Berkeley team.
Since a large number of terms are tested during the term selection
process, a stringent test must be used to avoid a large number of
terms from being included in the routing query by accident.  Using the
lower threshold allows a large number of terms with relatively
unsurprising frequencies of occurrence in the positive and negative
training sets to be included in the final query.

The Berkeley team assumed that their logistic regression techniques
would eliminate these noise terms, but it appears instead that,
consistent with the {\em post mortem} analysis performed by the
Berkeley group, the inclusion of these terms provided the opportunity
for over-fitting.  Increasing the number of degrees of freedom of a
system increases the likelihood of over-fitting, and over-fitting is
known to severely limit the generalization properties of a system.
For this reason, the very limited success of the Berkeley system is
not entirely surprising.

Robertson and Sparck Jones \cite{robertson} also reported a variety of
term weighting measures based on term counts in relevant and
non-relevant documents.  These measures appear similar to the method
reported here, except that all non-relevant documents, both judged and
unjudged, are lumped together and only very superficial attempts are
made to deal with small counts.  This work also had no explicit term
selection phase.  The significance of these problems was not apparent
at the time that the work was done since the methods proposed by
Robertson and Sparck Jones provided a very significant performance
benefit relative to the benchmark systems of the time.  Another likely
reason for the lack of distinction between the similar non-relevant
and general non-relevant examples is the fact that the evaluation corpora of
the time were exhaustively judged for every query.  This meant that
there was no distinction made between documents which were similar to
the query but still non-relevant and documents in general.  This work
on probabilistic weighting was taken further by van Rijsbergen in his
book \cite{rijsbergen}, but even there, all non-relevant documents
were lumped together.  These efforts and the underlying intuitions
live on in the term weighting schemes used by the Okapi system (see
\cite{robertson95}), and the InQuery system (see \cite{allan95} for a
recent report) and the numerous schemes incorporated into the SMART
system (as in \cite{buckley95}).  It is clear, however, that the
differences between these weighting schemes are not particularly large
when raw performance is compared.  

In Rocchio weighting, separate queries composed from the positive and
negative examples are subtracted.  This may allow many terms to be
introduced into the final query which may only have occurred
accidentally in the training documents.

There is some evidence in the recent literature which reports efforts
to exploit non-relevant documents which are similar to relevant
documents.  For instance, the query zone work described by Amit Singhal
\cite{singhal97} uses retrieval feedback to expand the query the user
provides.  This work uses documents near the query as well as
documents not so near to the the query to provide more refined query
expansion than can be provided by normal retrieval feedback.  This
focus on the non-relevant documents which are most similar to known
relevant documents is similar to Luduan term selection in some
respects, but the use of Rocchio style query vector addition makes the
detection of {\em significantly} different rates of occurrence
impossible to detect.  In addition, the design of the system
investigated by Singhal precludes the collection of judged example
documents.  It appears that the overall effect of the query zone
method is to allow {\em ad hoc} queries to gravitate toward nearby
clusters of documents.

The Luduan term selection algorithm can be used as a term selection
technique with virtually any of the term weighting methods mentioned
above.  This algorithm performs well in the important case of small
counts and can provide a robust indication of whether or not the
apparent variation in frequency of occurrence of a word is significant.
Since Luduan rejects terms characteristic of non-relevant documents
term weighting can be done without considering the fact that the terms
were derived using positive and negative examples.  This means that
term weighting can be much simpler than in many document routing
systems.

The Luduan system described in this chapter is based on the use of the
log-likelihood ratio test to select terms based on the term
frequencies in the judged relevant and the judged non-relevant
documents.  Term frequencies in all unjudged documents in the corpus
are ignored during term selection.  The use of one statistical test to
select significant terms and a second to provide term weights is a
relatively unusual approach in a document routing system.

The distinguishing factors of using positive and negative examples,
the use of the log-likelihood ratio test and using a non-adaptive term
weighting scheme are interrelated because the comparison of only the
documents which have been judged means that small count problems are
far more prominent than when the positive examples are compared with
the corpus at large.  This use of a very limited number of negative
examples makes the use of the log-likelihood ratio test much more
important than if the positive examples are compared to the corpus at
large because the counts for the negative examples are so much smaller
than they would otherwise be.  Much of the performance of the Luduan
system can be attributed to the effective use of well-chosen negative
information, but this use is only made possible by the fact that
log-likelihood ratio test produces good results in small count
situations.  

Furthermore, the use of a smaller corpus of negative examples also
means that any system which computes term weights based on term
frequencies in the relevant and non-relevant examples will be more
prone to over fitting due to a more limited set of training examples.
By using fixed term weights, Luduan avoids this problem.  Instead,
Luduan merely {\em selects} the terms to be used.  This problem of
selection is much less prone to over fitting.  One heuristic argument
as to why this is so can be had by appealing to the MDL principle
\cite{rissanen}: pure term selection involves the computation of one bit
per potential query term while a general term weighting approach
involves the computation of a term weight per potential query term.
Inevitably, the MDL complexity of term selection is much less than the
MDL complexity of term weighting.

Another argument for why Luduan's pure term selection strategy is less
prone to over fitting can be found by estimating the effective
Vapnik-Chervonenkis (VC) dimension \cite{vapnik} of a classifier
engine which is only allowed to select which terms to use rather than
the weights for all of the terms.  Clearly, the VC-dimension of a
weight selection engine will be equal to the number of terms for which
weights are to be computed since such a machine is simply a single
layer perceptron \cite{vapnik} operating on the term frequencies.  For
a typical query in the experiments described here, the number of
potential terms will be greater than 1000.  Thus, in the terminology
used by Vapnik, a classification engine based on weight selection will
be able to shatter any set of 1000 training examples and is very
likely to overfit any smaller set of training examples.  If the term
weights are fixed by other considerations such as term frequency in
the overall corpus, and the only degrees of freedom available to the
classification engine are the inclusion or exclusion of terms in a
query, then the VC-dimension relative to a set of realistic training
examples \cite{Vapnik+Levin+Cun:1994} is roughly equal to the number of
documents required to cover all of the terms found in a putative
positive example.  Based on informal experiments, this number tends to
be on the order of 10-20 documents.  Formally speaking, the
VC-dimension should be computed relative to all possible sets of
training examples, but, in practice, it is desirable to assess the
effective VC-dimension on cases which are likely to arise in real
problems rather than the theoretical value of the VC-dimension. 

This low VC-dimension for a term selection classifier relative to a
term weighting classifier indicates that the term selection classifier
will be roughly 100 times more resistant to over fitting than the term
weighting classifier.  Furthermore, a VC-dimension as low as that
exhibited by Luduan's term selection strategy gives some hope that the
system will perform well even with a very small number of positive and
negative examples.

\section{Evaluation \label{ir:methods}}

Two experiments were done to test the effectiveness of the Luduan
system.  In the first experiment, approximately the first 20,000 AP
documents the year 1990 from the TREC corpus were used to provide
training and test corpora.  This initial experiment showed very large
improvements in performance when Luduan was compared with other
systems.  Unfortunately, several potential flaws in this first
evaluation method made these results not entirely convincing.  To
resolve the questions raised by the first evaluation, a second, much
more rigorous evaluation was also done.  In this second experiment,
all of the TREC AP documents from 1988 were used as training data and
all of the TREC AP documents from 1989 were used as test data.

\subsection{Experiment 1}

In the first experiment, a moderate sized retrieval corpus was created
by taking 80 files of data from the Associated Press portion of the
TREC corpus.  This resulted in a 50 megabyte corpus containing 17,712
documents from a period of approximately 3 months in early 1990.
Based on the approximate number of documents in this sub-corpus, it is
referred to by the name AP20K.  Relevance judgements for the documents
in the AP20K corpus were extracted from the files containing all of
the document judgements from all of the TREC conferences from 1
through 4.  The 53 TREC topics which had more than 20 judged documents
in the AP20K corpus were retained for testing. This minimum of 20
example documents was chosen arbitrarily.

A variation on a cross validation experiment was used to test the
performance of the Luduan document routing algorithm using the AP20K
corpus.  In a strict cross validation experiment, the entire corpus
would be divided into training and test portions, typically in
approximately 4:1 proportions.  The 4:1 proportion was chosen to
reserve as much training data as possible while still leaving enough
test data to get an accurate appraisal of performance as possible.
The apportionment of training examples into training and test sets is
not part of the the Luduan algorithm itself and thus will not have any
effect on the performance of the algorithm.  Strictly speaking, this
cross-validation experimental design implies that indexing must be
done repeatedly on the test portions each time a different division of
training and test data is done.  This results in a computationally
very expensive experimental design.  Since the Luduan method for query
construction does not use any documents except ones for which
judgements are available, all of the unjudged documents in the
training data (very nearly 80\% of the entire corpus) would have been
wasted in a strict cross validation design.

To avoid these two difficulties of computational cost and waste of
data the strict cross validation design was modified.  The training
corpus was composed solely of 80\% of the judged documents (positive
and negative).  None of the unjudged documents in the corpus were used
in creating the Luduan routing queries, nor were summary statistics of
the entire corpus used in query creation.  The test corpus was
composed of the remaining 20\% of the judged documents as well as all
of the unjudged documents from the 80\% portion of the test data.
Since the test corpus was artificially depleted of relevant documents
by the deletion of the training documents, a serious bias was
introduced.  This bias was compensated for by weighting unjudged
training documents at 20\% of the weight given to the test documents.
Using the unjudged documents from the training set in this way
effectively increases the size of the test set by nearly 5 and
increase the amount of smoothing in the final results.

Average precision at various levels of recall was computed by taking
the 100 documents with the highest scores for each test query.  Since
a number of unjudged documents were necessarily included in this top
100, and since it was impossible to determine automatically whether
these documents were relevant to a particular test query or not, there
was a considerable degree of uncertainty possible in the measurement
of precision.  To account for this potential error, upper and lower
performance bounds were computed by first assuming that all unjudged
documents were not relevant to the test query, and then by assuming
that all unjudged documents were relevant to the test query.  These
two assumptions gave defensible worst- and best-case recall-precision
curves.  The general convention in the information retrieval
literature is to use only the worst-case curve which corresponds to
assuming that all unjudged documents are not relevant.  Such an
assumption that unjudged is equivalent to non-relevant would lead to a
very large bias in this experiment since no Luduan-like system has
ever been part of a TREC trial.  One result of InRoute having been
entered as a TREC-5 participant is that almost all of the documents
found by InRoute have been judged.  Because Luduan is very different
from previous TREC entrants, many of the documents retrieved by Luduan
have never been judged.

Since the range between the best and worst case curves was
substantial, spot checking was done to determine how many of the
unjudged documents were actually relevant.  This spot checking
substantiated the supposition that unjudged documents which had high
scores had substantial probabilities of being relevant.  This result
is not surprising since the original TREC format involved a pooled
evaluation methodology in which the top 1000 or fewer documents from
each participating site were evaluated.  Since the full TREC
evaluation corpus involved nearly 1,000,000 documents (or
approximately 50 times as many in this trial), the 100 document
retrieval depth used in this test corresponds roughly to a retrieval
of 5000 documents or more from the TREC corpus.  It is therefore
likely that there were at least some relevant documents which were
never examined by the human judges.

It is a general consensus in the IR community that the TREC relevance
judgements are very comprehensive.  This view is generally correct,
and the number of unjudged yet still relevant documents is probably
small.  The spot checking of high scoring documents in this test
clearly shows, however, that an advanced routing system such as Luduan
can find enough of these unjudged relevant documents to add a serious
negative bias to the results.  This negative bias does not reflect on
the quality of the original judgements in the TREC paradigm, it only
reflects the fact that the judges were not able to examine all
documents for relevancy relative to all queries.

As a result of the spot checking of unjudged documents, an additional
recall-precision curve was computed by assuming that the proportion of
unjudged documents which are relevant at a particular level of recall
is equal to the worst case precision at that point.  The resulting
recall-precision curve is very near the best-case curve at low recall
levels and moves progressively closer to the worst-case curve at
higher recall levels.  These curves are shown in section
\ref{ir:results} in figure \ref{fig:ir1} and \ref{fig:ir3}.

Precision at recall levels above those achieved within the top 100
documents was assumed to be zero.  Tests with differing assumptions
about where these missing documents would have been found showed that
even very optimistic assumptions gave recall-precision curves which
were nearly indistinguishable from the curves obtained under maximally
pessimal assumptions.  The small magnitude of the change was
due to the fact that only the very high recall end of the curve was
affected by differing assumptions.  Since precision was already quite
low at this point, decreasing it had little overall effect.

Several additional recall/precision curves were used for reference.
The first was produced using a variation on {\tt tf.idf} term
weighting known as {\tt lnc.ltc} weighting as described in
\cite{buckley93}.  In this retrieval method, no training documents
are used and no query expansion is done and thus the entire AP20K
corpus could be used for evaluation.  Upper and lower bounds for
precision were computed as for the routing method, but only the
interpolated precision was retained.  Since the precision for this
method was relatively low, the interpolated precision was near the
worst case curve.

The second reference curve was produced by transcribing the reported
precision results from the SMART TREC-2 results.  Since the results
were transcribed rather than measured, no statement about the
reliability of the curve can be made.  It is to be expected that since
all of the documents returned by the SMART system were judged, the
error bounds would be relatively tight.  

The third reference curve was produced by transcribing the reported
produced results from the Berkeley TREC-3 results.  The Berkeley
results were included since it is the only known system which used a
term selection method similar to that in the Luduan system.

\subsection[Discussion]{Results and Discussion for
Experiment 1\label{ir:results}}

Table \ref{table:query} shows a sample set of query terms which were
generated by the Luduan system.  The set includes words dealing with
countries which are actively involved in proliferation or opposing
proliferation (i.e the Israelis, and the Iraqis).  It also shows how
secondary associations are detected by the system with particular
names and with things related to nuclear proliferation such as
supercomputers.

\begin{table}
\begin{center}
\begin{tabular}[ht]{| p{5in} |}
\hline
\\
\multicolumn{1}{|c|}{Topic:  Nuclear Proliferation} \\
\\
Original TREC Query: \\
\\
{\tt
Document will discuss efforts by the United Nations or those nations currently
possessing nuclear weapons to control the proliferation of nuclear weapons
capabilities to the non-nuclear weapons states.} \\
\\
\hline
\\
Terms Generated by Luduan:\\
\\
{\tt iraq iraqi israel saddam customs baghdad london british nuclear iraq's
devices middle israeli indictment smuggle iraqis chemical hussein
triggers ministry 1981 smuggling weapons components arab alleged
binary shamir reactor bomb spokesman export iran daghir boucher
acquire speckman arrested bombs company officials diego saddam's
biological arrests mubarak concern israel's countries csi trigger
executive bombed supercomputers detonators newspaper news san
capacitors seized equipment charged believed quoted spread sale aziz
criticism program agency investigation u.s.-made comment senators
bazoft israelis deny supercomputer gnehm execution triggering denied
attacked britain attempt strike accused sales computer king attempts
purposes} \\
\\
\hline
\end{tabular}
\end{center}
\caption[Machine generated query terms]{
Query terms generated by Luduan compared with the corresponding TREC
query.  The relevant training documents were statistically compared to
the non-relevant training documents to generate the Luduan terms.
Note that the Luduan system has detected the correlation of nuclear
proliferation with chemical and biological weapon proliferation and
with supercomputers.  These query terms were used in experiments 1 and
2.\label{table:query}}
\end{table}

\begin{figure}[htb]
\begin{center}
\includegraphics[]{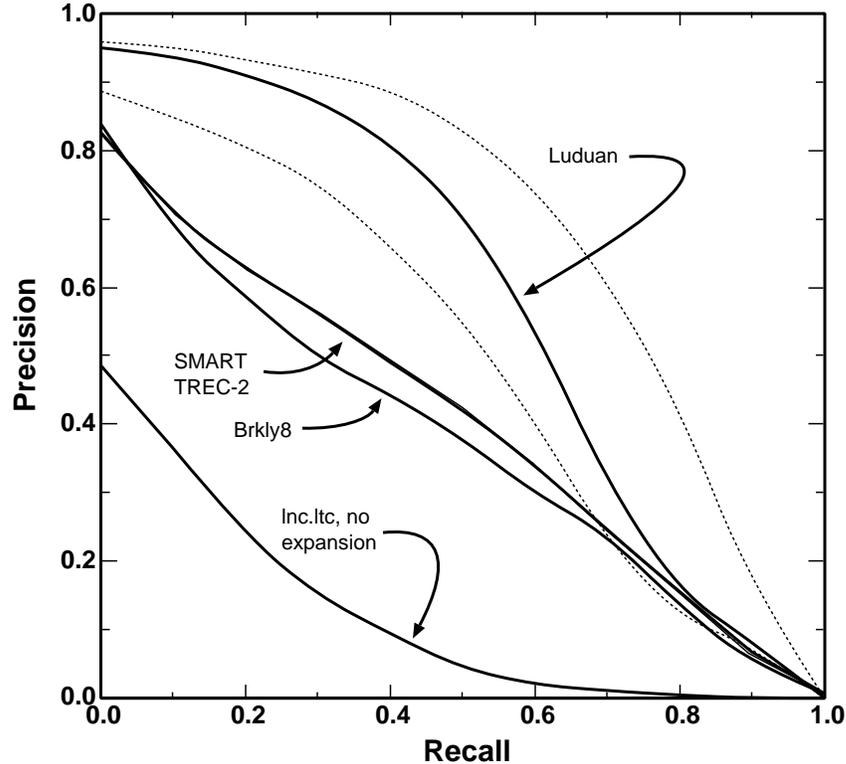}
\caption[Recall/Precision for Luduan system]{
Recall/Precision for the Luduan system in experiment 1.  These results
show the effectiveness of log-likelihood ratio tests for term
selection.  Dotted lines indicate best- and worst-case assumptions for
the Luduan results as explained in section \ref{ir:methods}.  The
solid Luduan line indicates the interpolated estimate of performance
as described in section \ref{ir:results}.\label{fig:ir1}}
\end{center}
\end{figure}

Figure \ref{fig:ir1} shows the
interpolated precision for the Luduan system compared to the SMART
and Berkeley routing systems and a
non-routing implementation of the {\tt lnc.ltc} term weighting method
which is common to the SMART and Luduan systems.  The dashed lines
represent extreme upper and lower bounds on
the performance of Luduan system, while the solid line between the
dashed lines is an interpolated estimate of the actual performance of
the Luduan system.

\begin{figure}[htb]
\begin{center}
\includegraphics[]{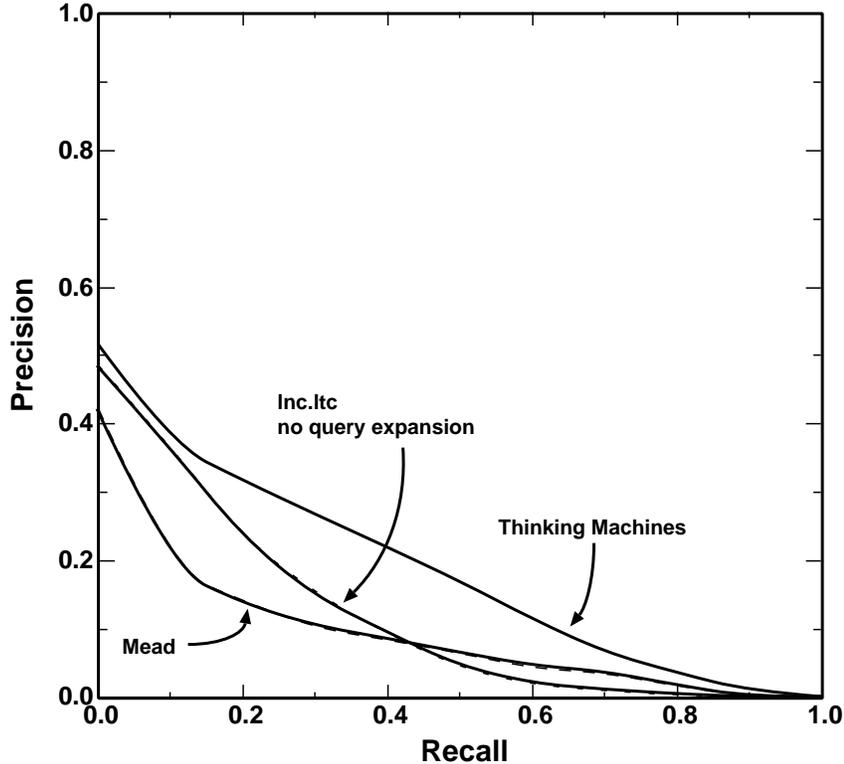}
\caption[Recall/Precision comparison]{Recall/Precision of the  {\tt lnc.ltc} scheme
compared to Mead Data Central and Thinking Machines systems
\label{fig:ir2}}
\end{center}
\end{figure}

As can be seen, the Luduan precision is considerably higher than the
SMART and Berkeley results in this test.  To insure that the SMART and
Berkeley results can be directly compared to the Luduan results, the
{\tt lnc.ltc} results were compared to two similar systems from TREC-2
as shown in Figure \ref{fig:ir2}.  In this figure, the {\tt lnc.ltc}
results are compared with the results achieved by the Mead Data
Central and Thinking Machines systems which are quite similar in
design to the {\tt lnc.ltc} reference system.  These systems were
chosen for comparison since they were two of the few full scale TREC
systems which did not use automatic query expansion and would thus
provide the closest basis for comparison.  As can be seen, the
performance of these two systems is similar to the {\tt lnc.ltc}
system.

The results of the evaluation on the Luduan system show that the
log-likeli\-hood ratio term selection method can make very substantial
improvements in the performance of a document routing system.  In the
experiment described here, 11 point average precision for Luduan was
41\% higher than that achieved by the SMART system (57\% for Luduan versus
41\% for SMART).  Since the TREC-2 SMART routing used essentially the
same term weighting scheme, most of this improvement can be attributed
directly as a quality of the term selection system.  This term
selection system critically depends on the ability of the
log-likelihood ratio test to deal with small numbers of observations.

There are several factors, however, which might make the extraordinary
results obtained less than convincing.  The design of this experiment
was novel and relatively unconventional.  This novelty could
conceivably hide a subtle bias.  The reweighting scheme applied in
order to allow the use of the unjudged documents from the training set
might be considered suspect.  The TREC reference systems were all
judged on a much wider range of documents than the AP20K corpus used
with Luduan.  Also, the set of relevance judgements which were
available to systems in the early TREC evaluations was more limited
than the currently available set.  It is difficult even to determine
precisely which relevance judgements were used by these systems.
Finally, the set of queries used here was the set of queries for which
there is adequate training data in the AP20K corpus; the systems in
the TREC evaluations had to work with all of the queries.

In Luduan's favor, the set of documents which were judged in the TREC
evaluations was based largely on the documents retrieved by the
systems evaluated.  Since the set of TREC judgements is static, if an
unjudged relevant document appears high in the Luduan's results, there
is no way for this document to be added to the set of TREC judgements.
Any document which appeared high in the results for the SMART or
Berkeley TREC entries would have been judged by a human analyst.  The
result is a negative bias of uncertain size in the Luduan results reported
here.

Just as worrisome, there are known to be a number of duplicate or
nearly duplicate documents in the TREC AP corpora.  The selection of
test and training documents at random as was done here raises the
potential that there were documents which were in both training and
test sets.

Given the enormous improvements apparently achieved in this first
experiment by Luduan relative to some of the best available research
systems, it is reasonable to re-examine the experimental design in
order to better assess the extent of the real advantages inherent in
the Luduan approach.  The second Luduan experiment does just that.

\subsection{Experiment 2}

To address the uncertainties raised by the potential flaws in the
first Luduan experiment, a second experiment was done which much more
closely reflects an operational scenario for a document routing system
and which directly compares the performance of two Luduan
configurations with 
two ``gold standards''.  These reference systems were the InRoute
system from the University Massachusetts and the Convectis system from
Aptex Software (originally developed at HNC Software, the parent
company of Aptex).  The version of InRoute that was used was the
TREC-5 routing entry \cite{Callan96}.  Convectis is a document routing
system which is in commercial use at a number of customer sites where
it is used primarily to categorize newswire, email and web pages
\cite{caid}.

In this experiment, all of the AP TREC documents from 1988 were used
as training data while all of the AP TREC documents from 1989 were
used as evaluation data.  Sixteen test queries were selected which had
at least 20 judged positive and negative documents in both 1988 and
1989.  Luduan queries were generated as described in the first
experiment.  These queries were then passed to the MG retrieval system
\cite{mg} and the Inquery retrieval system \cite{Callan96} for
evaluation.  This differs slightly from the retrieval step in the
first experiment in that stemming was not used by the Luduan system in
the first experiment but was used by both of the two systems in the
second experiment.  With a strong query expansion system such as
Luduan, it may be that stemming is actually counter-productive since
the query expansion provided by stemming is likely to already have
been done by the Luduan query expansion.  As tested here, the queries
generated by the Luduan system consisted of all morphological variants
which were found significantly more often in the relevant documents.
Stemming is known to produce a degree of over-expansion of the query
due to cases where morphologic distinctions are semantically
significant.  A Luduan query, on the other hand, can only contain
morphological variants which have been demonstrated to be useful in
determining relevance.

The fact that the test documents were separated in time from the
training documents should be expected to cause some decrease in
routing performance relative to the first experiment.  The degree of
this decrease is not well explored in the literature in spite of the
fact that temporally separated training and test sets are more
realistic scenarios for many applications.  For example, with topic
150 (regarding campaign financing), there are a number of articles
describing the actions of Robert Mossbacher in his position as
Secretary of Commerce under President Bush.  During the first part of
the year, these articles mentioned the fact that Mossbacher had
previously acted as campaign fund-raiser for Bush, but mostly these
articles had nothing to do with campaign fund-raising (at least they
had nothing to do directly with fund-raising).  In the training
documents, Mossbacher is never mentioned as the Secretary of Commerce
since he didn't have that job then.  This change in role on the part
of Mossbacher is likely to confuse virtually any document routing
system.

The Convectis system was used as a reference system.  Convectis was
developed by HNC Software as part of the DARPA-funded Tipster project
\cite{convectis}.  Convectis is currently being marketed by HNC's
subsidiary Aptex Software.  Convectis uses a uniform reduced
dimensional representation for both words and documents.  This
representation is developed by using a self-organizing training
algorithm which initially assigns pseudo-random vectors to each stem
which appears in a training set.  These random initial conditions are
then modified using an adaptive algorithm based on coocurrence
statistics to produced trained word vectors.  These trained word
vectors have the characteristic that words which appear in similar
contexts have similar trained vectors.  This training algorithm is
similar to, but significantly different from, the one used in the LSI
system \cite{Dumais88,deer1990} where term occurrence statistics are
analysed using singular value decomposition.

The word vectors in Convectis can be combined to produce document
vectors.  These document vectors are then used in Convectis to learn
hyper-conical decision surfaces using the LVQ algorithm
\cite{kohonen}.  Convectis can learn multiple vectors for a single
category simultaneously, but normally this is not necessary.  To
categorize documents, each document's vector is computed and the dot
product of these document vectors with each category vector is
computed.  If any dot product exceeds a preset threshold, then that
document is assigned to the corresponding category.  Normally,
Convectis only uses 80\% of the training data for learning category
vectors, holding the other 20\% out for the purpose of
self-evaluation.  In the trials described here, all of the training
data were committed to learning, and no self-evaluation was performed.
The dot product threshold was fixed at 0.3 which is standard procedure
for Convectis.  For each category, learning was run until performance
on the training documents reached 100\% or a fixed number of
iterations was reached.  Only two queries failed to reach 100\%
performance on the training data.  In order to produce traditional
recall/precision figures, the Convectis threshold was decreased after
learning, and the actual Convectis score was used as a relevance
indicator.

\subsection[Discussion]{Results and Discussion for Experiment 2}

The second reference system was the InRoute system.  This system is
consistently among the top performing systems in the TREC evaluations.
InRoute uses a variety of methods to produce queries which include
phrases and words.  These queries are then applied to the test corpus
using term weights derived from incremental statistics gathered as
documents pass through the system.  The InRoute results presented here
were produced by the same system as was used in TREC-5
\cite{Callan96}.  The InRoute and Luduan/Inquery retrievals were
performed by the Inquery team at the University of Massachusetts
\cite{callan97}. 
\begin{figure}[htb]
\begin{center}
\includegraphics[]{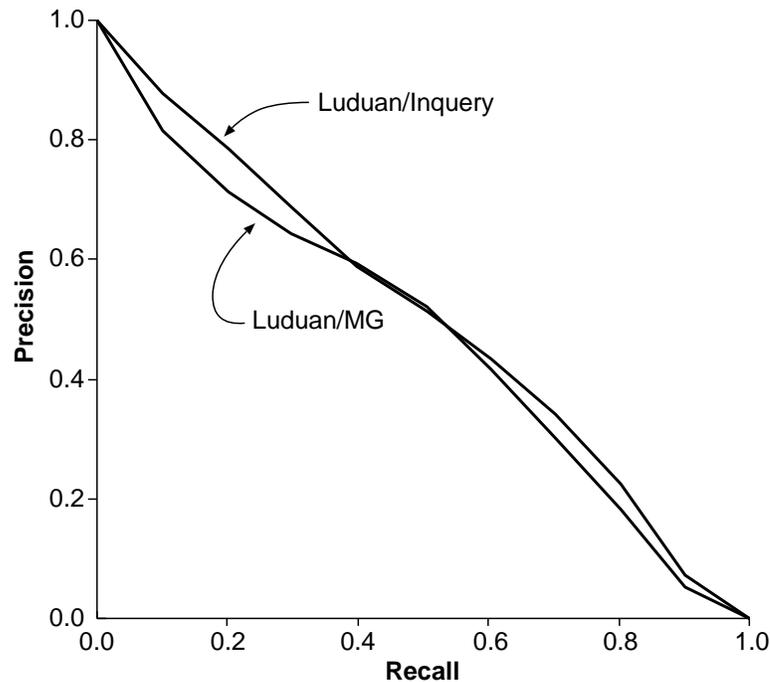}
\caption[Recall/precision of two versions of Luduan]{
Recall versus Precision of the Luduan/MG and Luduan/Inquery systems in
experiment two.  The high precision achieved by Luduan appears to be
relatively insensitive to the underlying term weighting scheme.
\label{fig:ir3a}}
\end{center}
\end{figure}
\begin{figure}[htb]
\begin{center}
\includegraphics[]{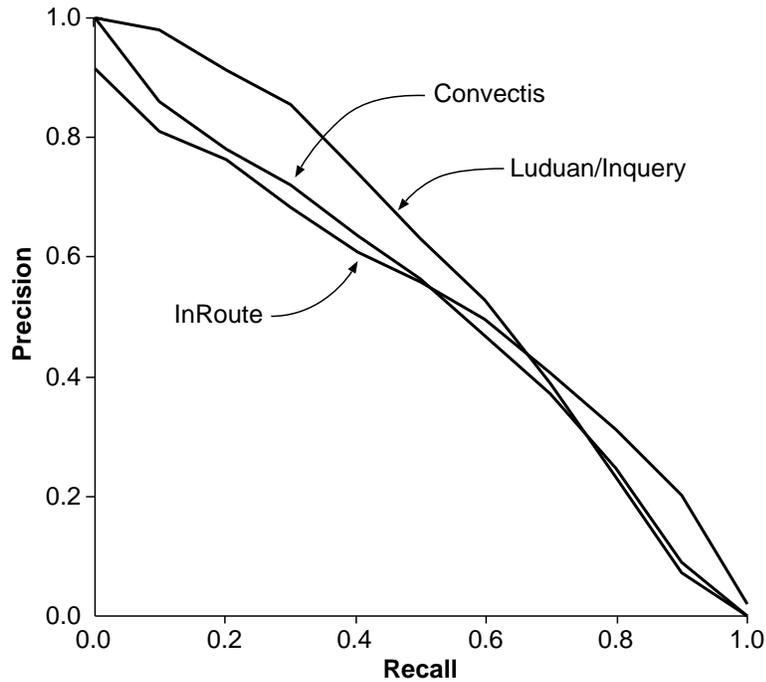}
\caption[Luduan performance compared with reference systems]{
Recall/Precision of the Luduan/Inquery system compared to Convectis
and InRoute systems in second experiment.  Additional judgements were
made to determine how the presence of documents which were unjudged in
the original TREC evaluations would affect the results.  Convectis
benefited less from these additional judgements, possibly due to
previous participation in TREC by Convectis.  Inroute did not benefit
from additional judgements probably because it was the TREC-5 entry
from UMASS.  The InRoute reference curve shows the performance of the
InRoute query generation and document routing sub-systems.
\label{fig:ir3}}
\end{center}
\end{figure}
Figure \ref{fig:ir3a} shows how the two Luduan variants compare.  As
can be seen, using the Luduan query generation with the more advanced
Inquery retrieval engine produces somewhat better results than when
Luduan is used with MG (average precision decreases from 0.4969 to
0.4683, a 6\% difference).  The difference is not substantial,
however.  This indicates that the major performance gains attained by
Luduan are not due to a peculiar interaction with the underlying term
weighting scheme.  This is a highly significant and somewhat
surprising property of the Luduan term selection method.

The results for Inroute, Convectis and the Luduan/Inquery systems are
shown in figure \ref{fig:ir3}.  Lower bounds on performance were
computed as in the first experiment by considering all unjudged
documents as not relevant.  To determine more precisely how accurate
these results were, an additional 665 documents were judged.  These
documents were taken from the top scoring unjudged documents for
Luduan/Inquery and Convectis.  This additional relevance information
permitted an accurate assessment of precision to be made up to at
least 10\% recall.  Since the Inroute reference system was a TREC-5
entry, no unjudged documents were found in the top 10\% recall.

As expected from the results in the first experiment in this chapter,
the recall/precision results for the Luduan-based system improved
substantially at the high precision end of the curve as a result of
the additional judgements.  Interestingly, the results for Convectis
did not improve as much as the Luduan/Inquery results improved.  This
may be due to the fact that an early version of Convectis was
evaluated in the TREC trials.  Since that early version of Convectis
had similar categorization performance to the current version, the
TREC judgements already include most of the documents the current
version of Convectis is likely to find.  With Luduan-based systems, on
the other hand, no similar system has been used in TREC.  The result
is that the overlap of the documents retrieved by a Luduan-based
system and the TREC judgements is somewhat lower.  This is especially
likely to be true at the low-recall end of the curve.

\begin{figure}[htb]
\begin{center}
\includegraphics[]{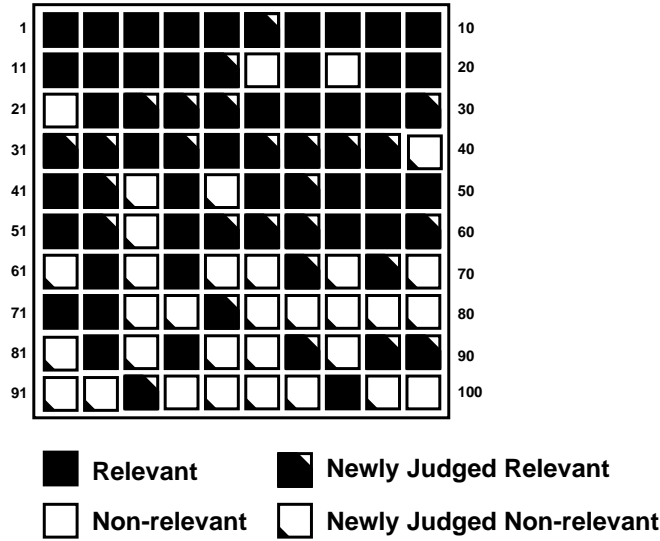}
\caption[Relevancy of top documents for a typical query]{
Relevancy of top documents for a typical query in second experiment.
In this diagram, each square represents a document.  The top ranked
document is in the upper left corner and succeeding documents are ordered
left-to-right, top-to-bottom.  Black squares indicate documents which
were judged relevant by TREC analysts, white squares those documents
which were judged not relevant by TREC analysts.  Documents judged
relevant as part of this experiment are black with one white
corner, those judged non-relevant as part of this experiment are
white with one black corner.  The performance of Luduan on this
query is relatively typical of the queries in this test.
\label{fig:ir4}}
\end{center}
\end{figure}

As can be seen by the results of this second experiment, Luduan
performs very well.  Precision at zero recall is perfect and this high
precision persists for substantial levels of recall.  The highest
scoring document was relevant for both Luduan variants for every query
tested.  More detailed examination of the results shows that all of
the queries produced good results.  Figure \ref{fig:ir4} shows
detailed results for a typical topic (number 114, having to do with
satellite launches).  In this diagram, the top 100 documents are each
represented by a box.  This character is filled in different ways to
indicate whether the corresponding document is known to be relevant or
not based on either the original TREC evaluations, or on the
additional judgements.  As can be seen, the top documents are mostly
relevant and the first document known to be not relevant is at the
16th position in the list.  This is typical for these results; indeed,
Luduan achieved relatively uniform performance on all queries tested
with no query proving much harder or easier than any other.  This
uniformity of performance is highly unusual.

\section{Conclusions}

Based on the results presented in this chapter, Luduan appears to be a
highly effective algorithm for automatically generating queries for
document routing applications.  The performance achieved by Luduan is
at least competitive with, if not superior to, the best research
systems available.

Another virtue of the Luduan system described here is that
the positive and negative exemplar documents can be converted into
queries in a matter of a few seconds.  This time compares quite
favorably with the query preparation time of hours or days which is
required for some of the more advanced adaptive methods such as those
reported by the OKAPI team at TREC-4 \cite{robertson95} and is well
suited for inclusion into user interfaces which help users craft
document routing queries.

Overall, the Luduan technique provides surprisingly good performance
with a very simple system design.  Further, Luduan can easily be
retrofitted into conventional retrieval systems.  As such, Luduan
merits substantial additional attention in both research and
commercial arenas.  Some more detailed suggestions for further work
are found in section \ref{luduan_future}.  These suggestions include
alternatives which would try to improve either the generalization
behavior or the specificity of the queries generated by Luduan.

\chapter{Language Identification\label{lingdet}}

This effort involved the development of a robust language
identification program which used Markov models.  The results achieved
by this identification program were very good; a few thousand words of
training data was enough to provide $>99\%$ accuracy for very short
test strings.  Other efforts had required nearly 2 orders of magnitude
more training data and even then did not achieve comparable accuracy.
The program described in this section incorporates no linguistic
presuppositions other than the assumption that text can be encoded as
a string of bytes.  Such a program can be used to determine the
language of small bits of text.  It also shows a potential for what
might be called `statistical philology' in that it may be applied
directly to phonetic transcriptions to help elucidate family trees
among language dialects.  The question of measuring the degree of
diachronic language evolution might also be addressed with a tool such
as the one described here.

A variant of this program has been shown to be useful as a quality
control in biochemistry.  In this application, genetic sequences are
assumed to be expressions in a language peculiar to the organism from
which the sequence is taken.  Thus language identification becomes
species identification.  The results of the species identification
effort are described in section \ref{bio_id}.

\section{Introduction}

Given the following strings, each 20 characters long,
\begin{verbatim}
       e pruebas bioquimica
       man immunodeficiency
       faits se sont produi
\end{verbatim}
it is hardly surprising that a person can identify the languages as
Spanish, English and French, respectively.  It is not even surprising
that a person who speaks very little French or Spanish can do this
correctly.  Clearly, language understanding is not required for
language identification.  Furthermore, only the French string contains
any closed class words, and none of the strings contain any accented
characters which are unique to that language (erroneously so in the
Spanish example).

Given this simple fact, it is a natural step to wonder just how simple
a computer program could be which is capable of performing a
comparable job of language identification.  It is also interesting to
ask whether such a program might be derived from relatively general
mathematical principles, or whether it could be written in such a way
that it could learn the characteristics of the languages to be
distinguished.  Another important question is how many characters are
needed to identify the language of a string with high reliability.
Finally, it is important to know just how broad the applicability of
such a program actually is.

The basis for a statistical test for comparing two strings to
determine whether they can be characterized by a single Markov model
was developed in section \ref{methods:source}.  This chapter describes
what is needed to apply that statistical test to the problem of
language identification.  The program used to implement the test can
be quite short, a few hundred lines of C suffice, of which the
statistical classifier takes up considerably less than 100 lines of
code.  The program is not based on hand-coded linguistic knowledge,
rather it learns the characteristics of the languages in question from
the training data.  An early version of this program was effective
enough that it was adopted by Sun Microsystems to be part of the
internationalization support for Java.

This program works moderately well at classifying strings as short as
10 characters, and works very well when given strings of 50 or more
characters.  The amount of training data needed is quite modest; as
little as a few thousand words of sample text is sufficient for good
performance.

Further, the program exhibits interesting generalization behaviors.
For instance, when it is trained on English, French and Spanish, the
program tends strongly to classify German as English.  This is
provocative given the historical basis of English as a Germanic
language.  The program can be used on a wide variety of materials with
the extreme case so far being the analysis of genetic sequences.

Adapting the program described in this chapter to new languages should
be relatively simple.  The only language dependent assumption
in the program is that whitespace characters can be collapsed into a
single character.  This assumption holds even in languages such as
Japanese and Chinese which do not normally use as much whitespace as
do languages such as English or Spanish.

\section{Previous Work}

The problem of the identification of the language of a text sample has
been addressed a number of times as described below.  In many
cases, these efforts are not reported in the literature, and in any
case there has been no systematic comparison of approaches.
Furthermore, most approaches taken so far have used some degree of
prior linguistic knowledge.  This use of linguistic intuitions makes
direct comparison with methods which are trained on a limited amount
of data difficult since the amount of data on which the intuitions is
based cannot be easily quantified.  Furthermore, many approaches have
been based on assumptions that may not hold for all languages of
interest, or for all applications.  For example, the amount of text
available to be classified may be quite limited, or one or more of the
languages in question may not be easy to tokenize into words.  In the
extreme case where the texts being analysed are not actually human
language at all, (e.g. genetic sequences) such assumptions may break
down entirely, thus severely limiting the utility of a particular
approach in some potential area of application.

While language identification is an extremely simple task when
compared to other canonical computational linguistic tasks such as
machine translation or information retrieval, it provides a perhaps
surprising amount of depth in terms of performance tradeoffs and the
variety of approaches which have been tried.  The availability of a
test set such as the one described in this section will make it
possible for new approaches to be tested with minimal effort on the
part of the implementor and will also allow the results of such tests
to be compared directly to the results obtained by other methods.

\subsection{Unique letter combinations}

There has been some commentary in electronic fora regarding the
possibility of enumerating a number of short sequences which are
unique to a particular language.

In particular, the following list of characteristic sequences was
proposed by a variety of people \cite{churcher94}

\begin{table}[htb]
\begin{center}
\begin{tabular}[htb]{|ll|}
\hline
\bf{Language}&\bf{String} \\
\hline
Dutch & \verb*|vnd| \\
English & \verb*|ery | \\
French & \verb*|eux | \\
Gaelic & \verb*|mh| \\
German & \verb*| der | \\
Italian & \verb*|cchi | \\
Portuguese & \verb*| seu | \\
Serbo-croat & \verb*|lj| \\
Spanish & \verb*| ir | \\
\hline
\end{tabular}
\end{center}
\caption{List of character sequences supposedly characteristic of a language.  The space characters in each string are significant.}
\end{table}

Clearly these particular strings do not serve as reliable signposts.
No one could claim that any text that discussed zucchini or Pinocchio
would have to be Italian, only a killjoy would claim that Serbo-Croat
had the only `lj' strings or that Amherst, Elmhurst, farmhands or
farmhouses could only be discussed in Gaelic.  And borrowing of
geographical terms such as Montreaux or the simple adoption of words
such as milieux into English would also confound these tests.

While better strings could no doubt be devised, the problems with the
unique string method are clear and do not go away when more effort is
made to be clever.  The problems do not lie in lack of cleverness, they
lie in the fact that such a unique strings approach depends on the
presence of only a very few strings, and that it fails to weight the
evidence that it does find.  Adding all observed strings leads to the
$n$-gram methods and picking an optimal weighting scheme gives the
Bayesian decision rule system described later.  Thus, the unique
strings approach incorporates a valid intuitive germ, but this germ
alone is not a strong enough foundation on which to build a usable
decision rule.

\subsection{Common words}

Another approach which has been proposed a number of times is to
devise lists of the words which commonly appear in a number of
different languages.  Since such common words make up substantial
amounts of running text, this approach works relatively well, if
enough text is available to be classified.  This approach has been
tested by the group at Leeds \cite{johnson} and is also rumored to
have been tried by researchers with the Linguistic Data Consortium.

As the 20-character sequences given in the introduction show, however,
as the strings to be classified become very short, there is a
substantial chance that the strings will not contain any common words.
This problem is exacerbated by the bursty nature of text when viewed
on a small scale; the closed class words form a structure around
pockets of less common words.  Furthermore, the short strings which
are interesting to identify are often exactly the strings which do not
contain the relatively common closed class words.  Rather, the strings
of interest tend to be the comparatively longer rare words and proper
names.  For example, a machine translation program might be very
interested in being able to identify English phrases in a primarily
Spanish text, or to determine that an unknown company name was likely
to be Japanese in origin.  In neither case could the common word
approach be expected work well.

Additionally, since common words tend to be short, systems such as the
$n$-gram which statistically weight the evidence provided by the short
strings found in text will naturally incorporate what information is
available from short words.  They will also be able to incorporate the
information provided by common suffixes and prefixes.  Furthermore,
the common word methods are somewhat limited in that they depend on
the ability to define and recognize common words.  If tokenization
into words is difficult (as in Chinese), or if defining a set of
common words is difficult (as in the case of genetic sequences) then
the common word approach may be difficult or impossible to apply.

\subsection{$N$-gram Counting Using Ad Hoc Weighting}

The group at Leeds has experimented with low order $n$-gram models as
described in \cite{hayes} and \cite{churcher}.  They used an {\em ad
hoc} statistical model to build their classifier which appears likely
to introduce some brittleness into their system.  Additionally, due to
the design of their test corpus, it is difficult to disentangle the
different factors which affect performance.  Their scoring method was
based on directly using estimated probabilities which are combined
linearly.  Basic statistical considerations show that if linear
combination is to be used, then the logarithms of the estimated
probabilities should be used, and some sort of smoothing should be
done when these probabilities are estimated.  Adding the logarithms of
estimated probabilities is essentially equivalent to using Bayes'
decision rule which has well known and provably optimal
characteristics \cite{dudahart} if the cost of errors can be characterised
by a constant loss function.  The non-optimal combining method used by
the Leeds group could well be responsible for some of the anomalous
results reported by the Leeds group.

\subsection{$N$-gram Counting Using Rank Order Statistics}

Cavnar and Trenkle \cite{cavnar} have reported the results of using
an {\em ad hoc} rank order statistic to compare the prevalence of short
letter sequences.  They first tokenize the test and training texts in
order to avoid sequences which straddle two words.

A rank ordered list of the most common short character sequences was
derived from their training data after tokenization.  Language
identification was done by compiling a similar list of common short
strings found in the string to be classified and comparing this list
to the profiles for the different training corpora.  This comparison
was done using an {\em ad hoc} rank ordering statistic which was designed to
be relatively insensitive to omissions.

They were able to extract a test database from a variety of network
news groups.  This provided samples of 8 different languages (English,
Dutch, French, German, Italian, Polish, Portuguese and Spanish).
These samples ranged in size from 22Kbytes for Dutch to 150Kbytes for
Portugese.

This system was able to achieve very good performance when classifying
long strings (4K bytes or about 700 words of English or Spanish).  They
report that their system was relatively insensitive to the length of
the string to be classified, but the shortest strings that they
reported classifying were 300 bytes.  This would be about 50 words in
English.

Some weaknesses in this work are that it requires the input text to be
tokenized and that the statistical characteristics of their rank order
comparison are not known and appear to be difficult to derive
rigorously.  It may well be that their tokenization step is not
necessary, which would allow their method to be applied in cases where
tokenization is difficult.  This issue was not explored in their
reported work.  It is possible that removing character sequences which
straddle word boundaries from consideration would eliminate any word
sequence information form consideration which might be harmful to
performance.  Other than the tokenization step and the particular
statistic used, their work is similar to the work described in this
paper.

A highly significant aspect of Cavnar and Trenkle's work is that they
have agreed to make their test and training data available to other
researchers.  This means that it may be possible to compare the
accuracy of different approaches under comparable conditions.

\subsection{Classification for Speech Synthesis}

Kenneth Church received a patent \cite{church_patent} which included a
description of a character $n$-gram based language identification
system which was intended to be used as part of a speech synthesis
system.  The goal was to prevent a speech synthesizer from saying
non-English words (especially names) using phonographic rules
appropriate for English.  The patent describes the use of a
statistical classifier which is similar in principle to the one
described here, but no details are given for the estimation of
parameters.  Doing this estimation well is crucial for obtaining good
performance.  Also, the system described in the patent works on a word
by word basis which is appropriate for the task described (word by
word speech synthesis), but is entirely inappropriate for the task of
identifying the language of a random bit of text.

\section{Classification Methods}

In section \ref{methods:source}, a method based on log-likelihood
ratios was derived which gives a measure of how compatible a test
string is with a longer training string.  This is done by constructing
a probability model which incorporates information from Markov models
of various orders.  This model is modified slightly by the observed
test data to construct a composite model which is compared to the
counts for various tuples in the observed string.  The resulting
compatibility score is largest when the composite model and the
observed data are most compatible.  This score can thus be used to
select which training string is most compatible with a test string.

It was found in practice that applying less smoothing to the training
model than is indicated in the derivation in section
\ref{methods:source} provides better performance.  In practice,
giving the test data a weight in the range $0.1-0.3$ rather than $1$
was observed to produce superior performance.  This result is not
particularly surprising since the derivation in section
\ref{methods:source} effectively make use of a uniform prior density.
It is well known \cite{Zipf49} that the distribution of word
frequencies is far from uniform and this same non-uniformity applies
to letter $n$-grams.  Thus, the value of heuristically decreasing the
value of $\alpha$ is not particularly surprising.  The compatibility
score between test string $S_1$ and training string $S_2$ used in this
chapter is given by
\begin{equation}
-2 \log \lambda \approx 2
\sum_{\sigma_0^k \in S_1}
T(\sigma_0^k, S_1) \log
{\frac
  {T(\sigma_0^k, S_1) \left( \alpha T(*^k, S_1)+T(*^k, S_2) \right)}
  {T(*^k, S_1) \left( \alpha T(\sigma_0^k, S_1)+T(\sigma_0^k, S_2) \right) }
}
\end{equation}
where $\alpha$ is the heuristic weighting factor mentioned above. 

This is in contrast to the score that would be derived by computing
the probability that the test string would have been generated by a
Markov generator whose parameters were estimated using Bayesian
methods with a uniform Dirichlet prior might be suggested.  This
alternative score is given by
\begin{equation}
\log \hat p(S_1) =
Z \sum_{\sigma_0^k \in S_1}
T(\sigma_0^k, S_1) \log
{\frac
  {\alpha + T(\sigma_0^k, S_2)}
  {\alpha \abs{\Sigma} + T(\sigma_0^{k-1}, S_2) }
}
\end{equation}
where $Z$ is a normalization factor dependent only on $S_1$.

The problem with this form is that when $\alpha$ is large, too much
smoothing is done resulting in very poor discrimination.  On the other
hand, for small values of $\alpha$, the penalty for strings in $S_1$
which are not in $S_2$ is too large and the system becomes excessively
brittle.  Very small values of $\alpha$ are required to avoid
over-smoothing since the smoothed probabilities apply to all possible
symbols.  The results for the Bayesian model on the problem described
in this chapter is that for small training sets, performance is nearly
pessimized.  On the other hand, it can be seen that in the score which
is based on the log-likelihood ratio test, the smoothing applies only
to symbols observed in $S_1$.  This smoothing effect is considerably
more concentrated than in the case of the Bayesian score and the
result, as will be seen, is good performance even in the case of a
small training set.

\section{Practical Results}

The performance of our $n$-gram language classifier was evaluated using
a specially constructed test corpus.  Versions of the classifier using
different size $n$-grams and various values of the weighting factor
$\alpha$ were compared.  Detailed comparisons with other scoring
methods such as a Bayesian estimator with uniform prior were not
conducted because in limited initial tests the performance of the
alternatives was clearly inferior to the model presented here.

\subsection{Construction of the Bilingual Test Corpus}

The following conditions were considered relevant to the performance
of a language classifier:
\begin{enumerate}
\item{how the test strings are picked,}
\item{the amount of training material available, }
\item{the size of the strings to be identified,}
\item{the number of languages to be identified, and}
\item{whether there is a correlation between domain and language.}
\end{enumerate}

It should be noted that language identification programs sometimes
detect the domain of a text.  For instance, Cavnar and Trenkle's
$n$-gram classifier was used with some success as the basic engine in
an information retrieval application \cite{cavnar}.  Other work along
these lines includes the work by Damashek and others as reported in
\cite{damashek}.  This means that it may be
difficult to separate the tasks of language identification and domain
(or subject) identification.  This is especially true if a language
classifier is trained and tested on material in which the domain of
discourse is strongly correlated with the language used.

We have avoided this problem of domain dependence and addressed all
the pertinent variables except the number of languages by constructing
a test corpus from a parallel translated corpus in English and
Spanish.  This test corpus provides 10 training texts with lengths of
1000, 2000, 5000, 10,000, 20,000 and 50,000 bytes respectively, and provides
100 test texts with lengths of 10, 20, 50, 100, 200 and 500 bytes.  In
total, 50 training texts and 600 test texts were selected initially in
each of the 2 languages.  The training and test texts were selected by
concatenating all of the documents in a particular language
and then selecting a contiguous sequence of bytes with a uniformly
distributed starting point.

All of the test strings were examined manually, and strings which were
not clearly Spanish or English were excluded from the test set and
replaced with newly generated strings.  Unfortunately a few test
strings which were defective escaped this filtering pass and were only
detected after initial results had been reported in technical reports,
and the test suite had already been used commercially.  These
defective strings were retained in the test suite in order to maintain
compatibility with earlier studies.  This problem of identifiability
was particularly a problem with the very short test strings since a
substantial number of the short test strings consisted almost entirely
of white space, numbers or punctuation.  There were also a few
situations where a string from the Spanish side of the parallel corpus
was actually in English or visa versa.  These cases were also excluded
from the final corpus.  All excluded texts were preserved.  This
procedure allows estimates to be made of the performance of a language
classifier under a number of different operating assumptions.

None of the training texts were excluded from the test corpus,
although one of the shorter training texts consisted primarily of
upper case text from a table of contents and another consisted
primarily of numbers taken from a table.  The effect of these
problematic training texts was limited by using the median instead of
the mean when plotting performance figures.  Running the tests with a
large number of training and test sets is a form of bootstrapping
analysis \cite{efron82, efron91}.  This form of analysis allows us to
obtain more than just a single performance figure for each combination
of training and test string size.  Best and worst performance figures
across all training and evaluation texts were plotted to indicate the
range of performance which can be expected with the method described
in this chapter.

\subsection{Test Methods}

The results presented in this section were obtained by training the
log-likelihood ratio classifier described previously on each training
set from the Eng\-lish/Span\-ish test corpus, classifying all test strings
and then recording the result.  In order to highlight when the
classifier had insufficient data to make a valid decision, the
classifier was run in such a way that the default answer in the
absence of data was incorrect.  In some extreme cases, it was
therefore possible to get error rates well in excess of 50\%.  In
normal operation, error rates this high are extremely unlikely, but
for comparison purposes, rigging the test against the program in this
way helps highlight failures.

No reference algorithm was used for this study since no alternative
methods described in the literature are likely to achieve any results
at all under the more extreme conditions of this test.  Of the methods
described earlier in this chapter, only the method described by the
Leeds group could even come close.  In the published accounts of this
method \cite{hayes, churcher}, however, it was stated that over 1
megabyte of training data was required to obtain accurate results even
for test strings of several hundred bytes.

Figure \ref{fig:lingdet1} shows the variation in error rate depending
on the order of the underlying Markov model and the size of the test
string for 50Kbytes of training data.  Figure \ref{fig:lingdet2} is the
same figure, except that the size of the training data is only 5Kbytes.
As can be seen, over this range there is little change in the
relationship between error rate, test string length and model order.
There is a very weak optimum in performance using the tetragram model
($k=3$) when the test string is very short.  This pattern holds for
most combinations of test string and training data sizes, although
with longer strings (200 characters or more) and large amounts of
training data, the error rate is essentially zero for any $k>0$.

\begin{figure}[htb]
\begin{center}
\includegraphics[]{lingdet1.eps}
\caption[Error rate for 50Kbytes training]{Variation in error 
rate with various parameters.  Note that models of order 3 and larger
give very good performance under a wide range of conditions.
\label{fig:lingdet1}}
\end{center}
\end{figure}
\begin{figure}[htb]
\begin{center}
\includegraphics[]{lingdet2.eps}
\caption[Error rate for 2Kbytes training]{Variation in error 
rate with various parameters.  Note that models of order 3 and larger
give very good performance under a wide range of conditions.
\label{fig:lingdet2}}
\end{center}
\end{figure}

\begin{figure}[htb]
\begin{center}
\includegraphics[]{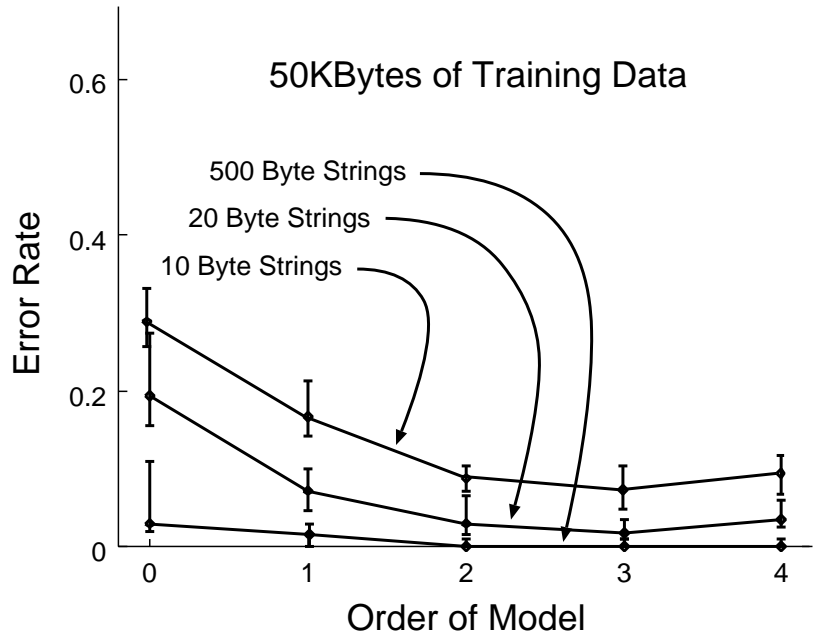}
\caption[Variability in error rate at 50Kbytes training]{Variability in error rate for large training set (50Kbytes) \label{fig:lingdet3}}
\end{center}
\end{figure}

\begin{figure}[htb]
\begin{center}
\includegraphics[]{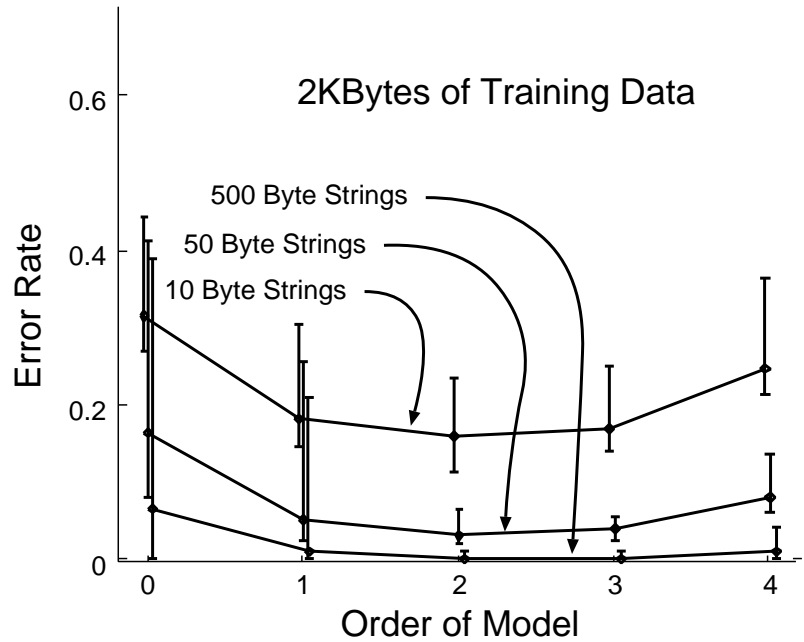}
\caption[Variability in error rate at 2Kbytes training]{
Variability in error rate for very small training set (2Kbytes).  Note
that even with such a training set equivalent to about a page of text,
accurate and reliable results can be obtained for long test
strings. \label{fig:lingdet4}}
\end{center}
\end{figure}

Figures \ref{fig:lingdet3} and \ref{fig:lingdet4} provide a closer
view of the variation in error rate with large (50Kbytes) and very small
(2Kbytes) training sets.  They also illustrate the large variability
in error rate when small training sets are used with higher order
models.  This effect is probably due to over-training.

\section{Conclusions}

Accuracy of about 92\% can be achieved with only 20 bytes of test text
and 50Kbytes of training, this improves to about 99.9\% accuracy when 500
bytes are examined.  With only 5Kbytes of training text, examining 500
bytes gives an accuracy of roughly 97\%.  All of these estimates of
classification accuracy are subject to statistical variation, but for
longer test strings ($>$100 bytes) and larger training sets (50Kbytes
or more) we can have roughly 90\% confidence that the accuracy will be
$>$99\%.  

Classification accuracy is good and improves with longer test strings
and with more test data.  With shorter test strings or with less
training data, bigram models appear to perform best.  With large
amounts of training data, and longer test strings, only the unigram
model has high enough error rates to measure easily with our
experimental design.

Idiosyncratic training data can be a problem, especially if only a
limited amount is available.  These effects can be minimized by using
bigram models.

Since there are literally billions of 4 and 5-grams, it is clear that
50 Kbytes of training data is nowhere near large enough to estimate
all model parameters accurately.  On the other hand, good accuracy is
obtained with only this much training data.  Apparently, the model
parameters which are responsible for discrimination performance
converge to sufficient accuracy long before all parameters can be well
estimated.

Because this experiment was designed to avoid accidentally correct
classification, and to hide confounding sources of information from
the classifier, it seems likely that performance in practice will be
even better than the results obtained in these experiments.  Providing
the classifier with data which are more uniform in nature should allow
it to achieve higher accuracy.  One way to do this would be to
normalize the case of all characters presented to the system.  In the
case of many character sets, the concept of data normalization can be
extended considerably beyond case normalization.  For example, the JIS
character set contains multiple copies of the ASCII characters as well
as copies of the Cyrillic and Greek character sets, many of which have
identical glyphs as ASCII characters.  Documents encoded using JIS
commonly have characters which are encoded by appearance rather than
function and often, double byte versions of the ASCII characters are
used in preference to the single byte equivalents.  Clearly, data
normalization could be of considerable value which character sets are
mixed.  Using a universal character set such as Unicode would
inherently solve some of these problems, but not all.

Simple classifiers such as those described in this thesis have a
number of applications.  For instance, a machine translation system
might be able to use such a system to determine whether short strings
which are difficult to translate should be carried verbatim, or which
language a source file is in.  Web browsers may find it convenient to
be able to guess accurately in what language pages are written in
order to fine-tune the presentation of such pages.

It may be that such systems could also be used in conjunction with a
speaker independent phoneme recognizer to identify spoken as well as
written languages.  This would have clear utility in certain
intelligence applications.  The input data would be much noisier than
the textual data used to produce the results described in this paper,
but the success of this system when used to classify short bits of
genetic sequences (as reported in chapter \ref{bio_id}) indicates that
noise may not be a problem with this sort of classifier.  Clearly,
techniques which depend on tokenization or which are sensitive to
noise would not be suitable candidates for this task.

It is also possible that this method could be applied to hand
generated phonetic transcriptions of speech to provide an objective
measure of similarity between different spoken dialects.  This process
would be subject to bias introduced by the hand coding process.  In
some sense, this measure of relatedness would be similar to the
language clustering work described in a report by Batagelj et alia
\cite{batagelj}, but while they classified languages according to the
editing distance between corresponding renditions of 15 common words,
applying an $n$-gram metric would allow the comparison of language
similarities based on realistic texts rather than a few specially
chosen words.  While the work of Batagelj et alia clearly shows that
orthographic normalization is not strictly necessary, using a phonetic
transcription would avoid some of the tendency to do comparisons of
orthographic families rather than linguistic or dialectic families.
Unlike much corpus based work, only a relatively modest amount of
transcribed material would be needed (a few thousand words).

The experiment in chapter \ref{bio_id} demonstrates how the algorithms
described in this chapter can be used to identify the species of
origin for a short bit of genetic information.  This ability to
distinguish species may be useful as a quality control in
biochemistry.  If a simple classifier such as this could be integrated
into sequence authoring or database submission tools, it might be
possible that major procedural errors could be detected much earlier
in the sequencing process, with the effect of improving the cost
effectiveness of sequencing projects.

Since the techniques necessary to produce good results 
are so straightforward, language identification can also serve as a
useful pedagogical tool.  The availability of standard data sets makes
it possible to have students implement their own algorithms and
compare them directly to standardized results.  Problems in
computational linguistics at this level of complexity are relatively
rare and standardized test cases are essentially unavailable for
problems other than language identification.

The program and test suite described in this section are available from
the author.  This test suite includes a test frame which allows other
language identifiers to be compared with the current program.  It
appears that a version of this software will be included in the
internationalization support for Java \cite{resnick}, but details on
this inclusion are not yet available.


\chapter{Species Identification\label{bio_id}}

The language identification algorithm described in section
\ref{lingdet} can be used for more than just the identification of
language.  A powerful and pragmatic application of this algorithm is
found in biology.  With this same algorithm it is possible to analyse
small amounts of genetic information in order to determine the
biological species from which the sequence was taken.  Given a
training set of reference sequence from each species of interest, it
is possible to subsequently analyse test sequences smaller than a
single gene and accurately identify the species of origin.

Of particular interest is the fact that the species identification
does not depend on finding a near match between the test sequence and
reference data.  Instead, very short-range statistical properties of
the training data are used to make the identification.  This makes the
method considerably more useful, as well as making it be of
considerably higher import in terms of biological theory.  This import
is largely because species or phylum specific patterns of this sort
have not previously been demonstrated.

\section{Overview}

In 1991, a French laboratory was embarrassed by the discovery that
over 1000 sequences of ``human'' DNA which they had released to the
public sequence databases were in fact from yeast DNA.  This error had
occurred because the laboratory had used a yeast cell line which had
been severely corrupted.  Instead of replicating bits of carefully
tagged human DNA, this cell library was busily replicating
accidentally tagged bits of yeast DNA.

This error was first demonstrated by an implementation of the
likelihood ratio test described in section \ref{methods:lr_binomial}.
The first public description of these results was at a U.S. Department
of Energy contractors workshop in Santa Fe.  Shortly after that, the editors of {\em Science} discussed this development and
the impact on large scale sequencing efforts in their ``News and
Comment'' editorial section (\cite{science_yeast}).  That summer a
more formal description of these initial studies appeared in
\cite{white}.

Since that time, the algorithm has been substantially improved.  The
source identification methods described in section
\ref{methods:source} have proven to be considerably better because
these methods are much faster and inherently give estimates of their
reliability.

One particularly exciting aspect of this work was the discovery in
1994 that exactly the same methods which could identify the species of
a small DNA sequence could also identify the language that a small
snippet of human language was in.  Section \ref{lingdet} describes the
result of applying these same methods to language identification.

\section{Experimental Methods}

In the work described in \cite{white}, discrimination between three
species was done ({\em E. coli}, {\em S. cerevisiae}, and {\em
H. sapiens}).  In order to explore more fully the trade-offs between
number of species and error rate, this study includes three separate
experiments in which two, three and eight groups were used.  For the
two source case, {\em E. coli} and {\em H. sapiens} were used.  The
three source case is a replication of the situation examined in
\cite{white}.  In the eight source case, samples of DNA from 
{\em C. elegans}, {\em D. melanogaster}, {\em H. sapiens}, {\em
G. domesticus}, {\em S. cerevisiae}, {\em E. coli} were used.  In
addition multi-species samples from monocotyledonous and
dicotyledonous plants were used.  This multi-species grouping was done
because insufficient data were available for any single plant species,
and it was considered important to have representatives from the plant
kingdom in the study.  Comparing multi-species groups to single
species groups does mean that the classification categories used in
this study do not reflect consistent phylogenetic groupings.

The performance of the classifiers is reported both as raw error
percentages and also in terms of the cross entropy between the
estimated and true probability of particular classifications.  Error
rates were reported because they provide a better visual
discrimination between systems.

Only exons were used for most of the organisms since the original
purpose of the experiment was to distinguish human expressed sequence
tags (EST's) from yeast or bacterial contamination introduced as part
of the yeast artificial chromosome (YAC) or bacterial artificial
chromosome (BAC) procedure.  Because of the nature of the procedure
for introducing human sequence into the YAC or BAC, the crucial
comparison to make is the distinction between expressed sequence from
the human DNA (i.e. exons) and any sequence from the substrate
organism.  This means that the predominant alternatives in the
experiments which originally motivated this work are that a sequence
will either be a human expressed sequence or a random bit of sequence
from the substrate organism.  In the case of {\em E. coli}, nearly all
of the genome is expressed, so this distinction is less important.

\subsection{Data Preparation}

The data used in this study consisted of all exons found in GenBank
from {\em Caenorhabditis elegans} (a nematode or roundworm), {\em
Drosophila melanogaster} (fruit fly), {\em Ho\-mo sapiens} (human),
{\em Gallus domesticus} (domestic chicken) as well as for a group of
dicotyledonous plant species [largely {\em Nicotiana tabacum}
(tobacco), {\em Solan\-um lycopersicum} (tomato), {\em Alfalfa
meliloti} (alalfa), and {\em Arabidopsis thaliana} (the flowering
plant with the smallest genome)], and a group of several
monocotyledonous plant species (mostly {\em Zea mays} (corn) and {\em
Oryza sativa} (rice)).  In addition, all of chromosome 3 from {\em
Saccharomyces cerevisiae} (a species of yeast) and a substantial
contiguous portion of the genome of {\em Escherichia coli} (a commonly
occurring enterobacterium) were also used.  Together, these datasets
represent a highly diverse phylogenetic range including prokaryotes
(bacteria) and a broad sampling of eukaryote species from three of the
four eukaryotic kingdoms (only protists are not represented here).

For data other than from {\em S. cerevisiae} and {\em E. coli}, only
exons bordered by complete introns were used and only the exon was
retained.  Any exons which did not meet a small set of sanity checks
were eliminated.  These checks included scanning for zero length
sequences as well as sequences whose entries in the database had been
corrupted or could not be parsed.

To avoid problems with duplicate sequences appearing in different
database accessions for a single species, sequences which were
approximately the same length and which were more than 95\% identical
to a shorter sequence from the same category were eliminated.  This
reduction was accomplished by sorting all sequences by length and then
using a dynamic programming alignment algorithm on all sequences which
met a heuristic test for equality.  The alignment used a variation on
the well known algorithms for finding minimum edit distance between
two strings.  The heuristic comparison was done by recording the
distinct $n$-mers in each string in a bit table indexed by a hash of
the $n$-mer.  This allowed comparison between strings to be done using
bitwise {\em and}.  Since the bit tables were fixed in length, this
comparison took a constant and relatively small amount of time, no
matter the length of the strings in question.  The number of $n$-mers
found in both strings (and from this the number which did {\em not}
appear in both) could be conservatively bounded by the number of bits
which remained after the {\em and} operation.  The number of $n$-mers
not found in both strings in turn can provide a lower bound on the
edit distance between the strings.  The resulting program could
exclude duplicate sequences from files containing thousands of
sequences in roughly a minute using a Sun workstation (SparcStation 2
with approximately 64MB RAM).  This performance is a substantial
improvement over the naive approach of directly comparing all
sequences which took hours to days to complete.  The worst case time
complexity for this algorithm is still quadratic, but for this
application, the time required was quite moderate.

Eliminating all near duplicates in this manner runs the risk of
eliminating sequences which actually are duplicated in multiple
locations in the genomes.  Since the lack of precise location
information in many entries makes it impossible to determine
accurately which entries in the database reflect the same sequence
locations, such a conservative strategy of eliminating all possible
near duplicates was the only safe option.  The effect of the duplicate
removal process should be to decrease the accuracy of a species
identification algorithm by artificially increasing the diversity of
the training and test data.  This makes the overall analysis presented
here very conservative.

The resulting database of exons was divided into two portions.
Approximately 90\% of the data was used for training, while the
remaining 10\% was used as test data.  The {\em S. cerevisiae} and
{\em E. coli} training data were divided arbitrarily into 100 parts to
allow the bootstrap technique to be applied \cite{efron82, efron91}.
The test data for {\em S. cerevisiae} and {\em E. coli} were
artificially segmented into parts with the same distribution of
lengths as that observed for the human exons to prevent length from
being a cue as to origin.  The remaining sequences from other species
were exonic and had natural boundaries, so further segmentation was
not necessary.

Table \ref{table:bio1} shows the sizes of the training and test data
in bytes (bases).  Since all available sequences were used, the sizes
of the different files vary somewhat.
\begin{table}[htb]
\begin{center}
\begin{tabular}[t]{|rrr|}
\hline
\bf{Category} & \bf{Training} & \bf{Test}\\
\hline
chicken & 44547 &  5440 \\
dicots & 181940 &  19816 \\
bacteria & 82277 &    9735 \\
fruit fly & 185050 &    18421 \\
human & 251783 &  27391 \\
monocots & 92642 & 10156 \\
nematode & 328282 &   29201 \\
yeast & 287121 &  30308 \\
\hline
\end{tabular}
\end{center}
\caption[Number of nucleotides in training and test data files]{Number of nucleotides in training and test data files.
Categories do not represent parallel phylogenetic distinctions since
some include only a single species while others contain many.\label{table:bio1}}
\end{table}
Table \ref{table:bio1a} shows the number of training and test
sequences used.  The number of test sequences was sufficient to
provide good accuracy in the final results, so the exact value of the
training/test ratio is not likely to have a significant impact on the
results.  The use of the bootstrap technique in the experimental
design allows confirmation of this insensitivity of the overall
results to this ratio.
\begin{table}[htb]
\begin{center}
\begin{tabular}[t]{|rrr|}
\hline
\bf{Category} & \bf{Training} & \bf{Test}\\
\hline
chicken &    332  &    37 \\
dicot   &    965  &   107 \\
ecoli   &    553  &    62 \\
fly     &    415  &    47 \\
human   &   1651  &   184 \\
monocot &    413  &    46 \\
worm    &   1097  &   122 \\
yeast   &   1898  &   211 \\
\hline
\end{tabular}
\end{center}
\caption{Number of training and test sequences\label{table:bio1a}}
\end{table}
For comparison purposes, Table \ref{table:bio2} shows the total genome
sizes for selected organisms.  It is clear that only a tiny fraction
of the total genome of any of these species is represented in the data
used in this study.

\begin{table}[htb]
\begin{center}
\begin{tabular}[t]{|rrl|}
\hline
\bf{Organism} & \bf{Genome Size} & \bf{Comment}\\
\hline
{\em C. elegans} & $ 8 \times 10^7$ & nematode\\
{\em A. thaliana} & $ 7 \times 10^7$ & smallest plant genome \\
{\em Z. mays} & $ 5 \times 10^9 $ & corn (monocot)\\
{\em A. cepa} & $ 1.5 \times 10^{10}$ & onion (monocot)\\
{\em D. melanogaster} & $ 1.65 \times 10^8$ & fruit fly\\
{\em H. sapiens} & $ 2.9 \times 10^9$ & human\\
{\em G. domesticus} & $ 1.2 \times 10^9$ & chicken\\
{\em S. cerevisiae} & $ 1.35 \times 10^7$ & yeast\\
\hline
\end{tabular}
\end{center}
\caption[Total size in bases of genomes for selected organisms]{Total size in bases of genomes for selected organisms.  Some
classification categories are single species and some contain multiple
species. \label{table:bio2}}
\end{table}

The classification of individual test sequences was done using the
source identification method described in section \ref{methods:source}.

The classification method used depends on producing a Markov model
from the training data for each category.  For a given test string,
these Markov models are then used to compute the probability that the
test string might have been produced by each of the Markov models.
The key design parameter is the order of the Markov models used.  The
order of the models is one less than the window size.

In order to explore the space of design parameters for the classifier,
the window size was varied, and the error rate was measured for each
window size.  This allows some extrapolation to be done regarding the
performance of the classifier under different conditions.  In
addition, by examining conditions under which the classifier performed
well or poorly, it is possible to draw some tentative conclusions
about the biological basis for the performance of the classifier.  For
instance, a persistent criticism of the earlier study \cite{white} was
that the classification performance could have resulted from the well
known difference in average base composition between organisms.  Since
that study used only a single window size ($n=6$), this criticism
could not be refuted.

On the other hand, if a classifier with window size 6 were shown to
work well, but a classifier with window size 1 were shown to work
poorly, this criticism could be dismissed.  This is because with a
window size of 1, the only information available to the classifier is
average base composition of the training data.  If this is not
sufficient to classify the test data accurately, then the superior
performance of the classifier with a larger window size must be due to
the fact that the larger window size allows the use of more structural
information than average base composition.  To explore the design
space further, the difficulty of the classification task was varied by
running three experiments: one with two species, one with three
species and the other with all eight categories.  This variation
allows comparison of the current results with previous work, as well
as extending the range of knowledge about this class of classifier.

The results of the classification algorithm are either a strict
classification or an estimate of the likelihood that a particular
sequence came from each of the categories.  The strict classification
was scored on the basis of observed error rates.  The soft classifier
was scored using the cross entropy ($d$) which is defined as
\begin{equation}
d(p,q) = - \sum_{\sigma \in \Omega} p(\sigma) \log q(\sigma)
\end{equation}
where $p$ is the true probability distribution of the events in
$\Omega$ and $q$ is our estimate.  $\Omega$ is the set of all possible
events.  Of course, it is impossible determine the exact value of this
divergence since the value of $p$ is, in general, not known.  This
requires the use of a population estimate based on the test data where
we know what the true classification is.  The value of this estimated
cross entropy is
\begin{equation}
{\hat d}(p,q) = 
- \sum_i \frac {\sum_{s \in {\mathcal S}_i} \log q(s=i)} {\abs{{\mathcal S}_i}}
\end{equation}
where ${\mathcal S}_i$ is the $i$-th set of test sequences.  For the
three species test, $i$ would range over the set $\{ H. sapiens,
S. cerevisiae, E. coli \}$ and ${\mathcal S}_i$ would be the test
strings for each of these species.  The similarity of this score to
the log-likelihood ratio score described in \ref{methods:lr_tests} is
not accidental.  It derives from the fact that $\hat d$ is minimized
exactly when $p = q$ which makes $\hat d$ a natural measure of the
similarity of $p$ and $q$.  This measure is closely related to the
Kuhlback-Liebler divergence which is defined as
\begin{equation}
D(p,q) = \sum_x p(x) \log p(x)/q(x)
\end{equation}
The Kuhlback-Liebler divergence has a virtue in that it has a minimum
value of 0 when $p = q$, but it is considerably more difficult to
estimate than cross entropy.  This difficulty arises since $p$ (or at
least $\sum_x p(x) \log p(x)$) must be estimated explicitly.  Since
$q$ is generally already our best estimate of $p$, a straightforward
approach to estimating the Kuhlback-Liebler divergence leads to an
estimate of zero.  By using cross entropy instead, this problem of
estimating is avoided at the cost of no longer having a known minimum
value.

The Kuhlback-Liebler divergence can be estimated by a procedure which
involves dividing the sampled set into two parts and estimating $p$ on
one half and $q$ on the other (and visa versa).  Unfortunately, this
estimation technique buys a known minimum at the cost of greater
expected error in the estimate.  Also, the estimate of divergence can
actually be negative which is an upsetting result when estimating a
value which is known to have a minimum value of zero.

\section{Experimental Results}
Figure \ref{fig:bio_id1} below shows the estimated cross entropy of
the classifiers for the three classification tasks (with two, three
and eight categories) and for a variety of window sizes (corresponding
to different Markov model orders).  For all three tasks, both average
error rate and cross entropy show a broad minimum with best
performance occurring at a window size of 6 or 7 nucleotides.
\begin{figure}[htb]
\begin{center}
\includegraphics[]{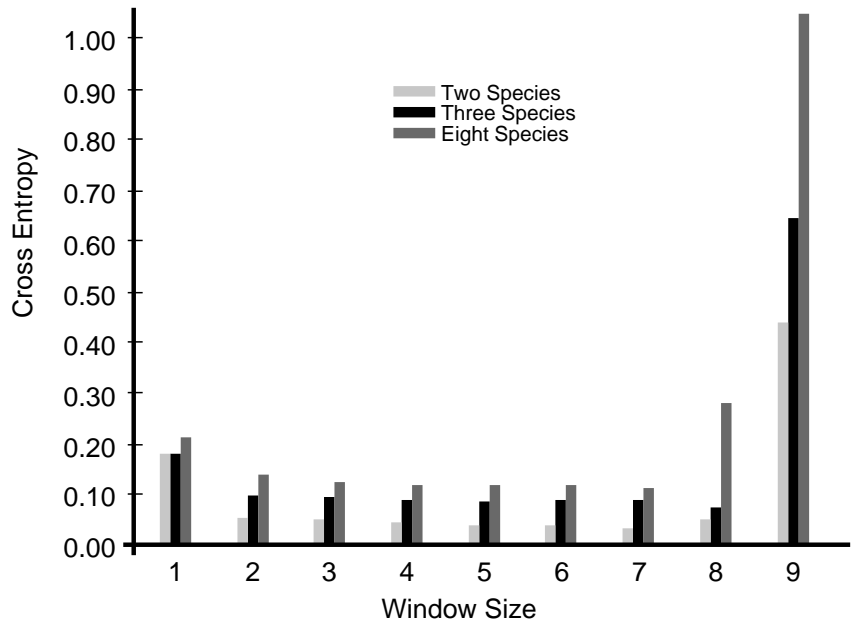}
\caption{Cross Entropy of Classifiers \label{fig:bio_id1}}
\end{center}
\end{figure}

\begin{figure}[htb]
\begin{center}
\includegraphics[]{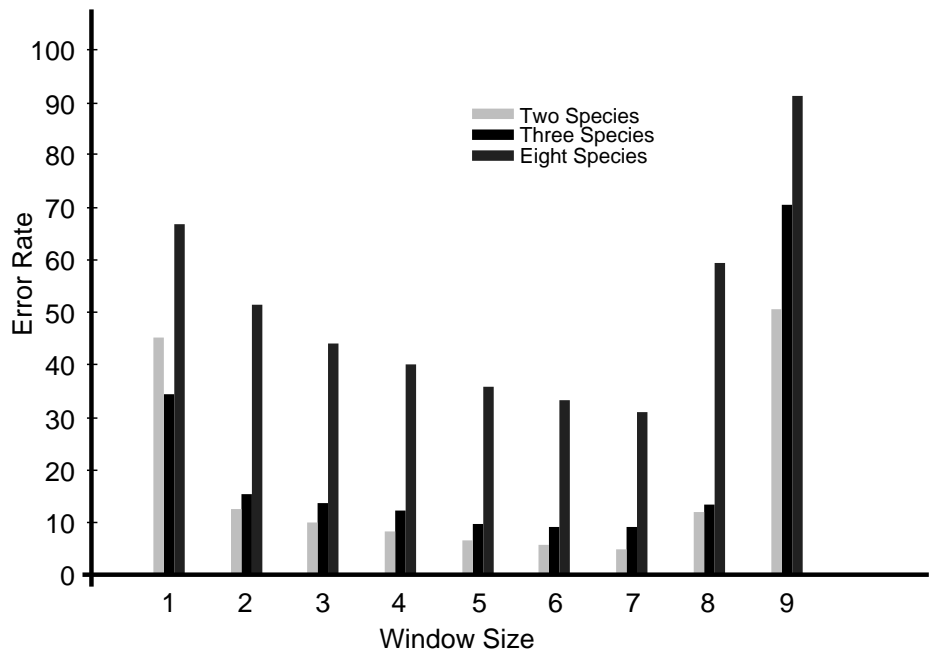}
\caption[Error rate of Sequence Classifiers]{Error rate of Sequence Classifiers.  Error rate is shown
instead of rate of correct identification in order to highlight
differences in performance.\label{fig:bio_id2}}
\end{center}
\end{figure}

Figure \ref{fig:bio_id2} shows the average error rate for the same
range of conditions as shown in figure \ref{fig:bio_id1}.  The rate of
correct identification (which is of greater interest practically) can
be easily determined by subtracting the error rate from 100.  Error
rate was shown since it provides better visual discrimination between
cases with nearly identical performance.

\section{Conclusions}
From a biological perspective, the simplest question about
interspecies nucleo\-tide composition is whether or not the average base
composition varies significantly between species.  This question is of
interest because GC complementary pairs bind more tightly than AT
pairs, and this difference in binding strength is thought to have
biological function because there are long stretches of DNA which are
high in AT which may result in a micro environment of less tightly
bonded or ``breathing'' DNA.  Similarly, local regions where
methylation plays a strong role are expected to have more GC content.
The method of analysis presented here addresses overall base
composition when the window size is one.  The relatively poor
performance on all classification tasks at a window size of 1
indicates that the differences in overall base composition (particular
in the average frequency of GC relative to AT) between different
organisms cannot explain the performance of the classifier.

The optimum error rate of the classifier described here on the three
species task is at least 30\% lower than the error rate of the
classifier described in \cite{white}.  The previous effort resulted in
error rates of approximately 15\%, while the current system has error
rates under 10\%.  The speed of the current system is also
considerably better (it is faster by 3 orders of magnitude or more)
than the performance of the earlier system.

The sudden degradation in performance at window sizes of 8 and 9 is
probably due to over-training.  Over-training occurs when the classifier
is unable to generalize from idiosyncrasies in the training data and
thus performs very poorly on novel test data.  Increasing the amount
of training data substantially might make it possible to develop a
classifier which performed well at these larger window sizes.  In
contrast to the case of language identification, these larger window
sizes are not a substantial fraction of the size of the typical test
string (in this study this is 300 base pairs or more), so end effects
are probably not responsible for the degradation of performance with
large window sizes.  Extrapolating from the performance at lower
window sizes, however, it is unlikely that this classifier could
obtain substantially better error rates even if it were practical to
train a classifier using longer windows.

The degradation of performance at larger window sizes has no
implications biologically.  It can be shown that performance with
larger window sizes cannot be worse than the performance with short
windows, if we assume that accurate probability estimates can be made.
The current classifier automatically uses information from all smaller
window sizes, but the details of the implementation and parameter
estimation method cause this property to fail at large window sizes.
Since the method reported in this chapter takes no pains to make such
accurate estimates, the decrease in accuracy at larger window sizes is
predictable without invoking biological explanations.

Moreover due to the relatively even distribution of the frequencies of
different $n$-mers, however, it is likely that using better parameter
estimation techniques with the current training data would only allow
classifiers with large window sizes to match the performance of the
classifiers with smaller windows but not to exceed their performance
significantly.  This limitation is more a feature of the classifier
and estimation methods than of the biological system being studied.

Conversely, if a classifier were to be designed which showed
significantly higher performance with window sizes larger than eight
or nine, there would be substantial biological implications.  These
import of such results would extend the range over which species
specific motifs are known to exist from the short motifs found in this
work to a much larger scale.  The design of such a classifier is an
excellent focus for further research.

These results collectively show that distinctions between biological
species occur at several very short-range levels of genetic structure.
Importantly, the method shown here differs in principle from other
methods for database search and analysis.  It is common to compare a
specific DNA sequence that is of biological interest to an existing
database of known sequences from a variety of species.  A search for
direct sequence similarity is then used as the basis of the test
fragment.  This method of search has obvious limitations in that the
method depends on the accuracy of pre-existing identification of
related sequences.  In such near match search programs, there is no
basis for correctly assigning the species of origin of a fragment if
it is for a completely novel gene, or if it belongs to a species
without an extensive presence in the sequence databases.

In contrast, the method described here relies on detecting inherent
structural properties of the genome of a species that distinguish it
from other species.  In much the same manner as the language
identification task described in section \ref{lingdet}, no lexicon and
no exact matching of gene fragments or words are required.  In
addition to the pragmatic task of quality control of genetic
databases, these results suggest the existence of functionally
significant species specific genomic characteristics based at a level
of resolution quite different from that of the gene.

%



\chapter{The Structure of Introns\label{intron}}

\section{Overview}

Genetic information in all known living organisms is encoded using DNA
(deoxy-ribonucleic acid) or RNA (ribonucleic acid).  These molecules
are chemical compounds known as polymers because they are constructed
from a very small number of constituents repeated many times.  Both
types of nucleic acid polymers consist of linear structures of great
length.  DNA often occurs as two complementary linear chains weakly
bonded to each other to form a double helix.  These molecules reach truly
stupendous size (they are actually millimeters or centimeters long in
some cases).  Even more stupendous is the fact that all known living
organisms use nucleic acid to accomplish essentially the same
information storage function with only minor variations in the coding
system used to convert from DNA into protein.  The universality of
genetic code is often adduced as strong evidence for biological
evolution.

Although there exists universality of genetic code, there are
differences in the arrangement and packaging of genetic information in
different living systems.  The simplest cells are those of bacteria,
known as prokaryotic cells.  These cells consist primarily of one
(relatively) large compartment enclosed by an inner and outer cell
membrane and a more rigid cell wall.  The genetic material is stored
as a single large loop of DNA, protected only by the cell itself.
Other living cells are called eukaryotic cells, and they contain
sub-cellular compartments (organelles).  One of these, the nucleus,
contains the genome (the genetic information) of the organism in the
form of chromosomes.  Each chromosome contains a DNA molecule
elaborately packaged with protein into a very regular, very condensed
structure.  There are other sub-cellular compartments that contain
DNA, but surprisingly, this other DNA more closely resembles the
single naked DNA molecule which forms the prokaryotic genome rather
than the eukaryotic chromosomes.  These sub-cellular compartments
include the mitochondria and, in photosynthetic cells, the
chloroplasts.  The DNA of these compartments embedded in eukaryotic
cells replicates independently of the nuclear genome.

Mitochondria and chloroplasts also have their own protein building
structures (ribosomes) to translate genetic information into proteins.
The ribosomes of mitochondria and chloroplasts also bear a closer
resemblance to those of prokaryotes than to the ribosomes of the
eukaryotic cells in which the mitochondria and chloroplasts are
embedded.  These similarities as well as other characteristics of
mitochondria and chloroplasts have led many scientists to speculate
that the ancestors of these organelles were prokaryotes which
parasiticized other cells.  Eventually an endosymbiotic relationship
developed between parasite and host, giving rise to the modern
eukaryotic cell with sub-cellular organelles.

The data contained in genomes include sequences whose function are
reasonably well understood, and sequences whose function is still more
or less obscure (with the emphasis on more).  In general, most of the
information-bearing genome portions that have been studied are the
ones used by cells to encode the structure of proteins or to construct
functional RNA molecules of several classes, such as ribosomal RNA or
transfer RNA.

The process of using or expressing information from these sequences of
known function initially involves the synthesis of an RNA copy of the
sequence of nucleotides in a stretch of DNA; the order of bases in the
newly formed RNA corresponds directly to those in the DNA being copied
and thus are said to be complementary.  This process of making an RNA
copy of information encoded in DNA is called transcription.

In some cases, the transcription product is an RNA molecule that
carries out its terminal function without the need for conversion into
another molecular form, although some transcripts may require cleavage
and some nucleotides are chemically modified.  Ribosomal RNA and
transfer RNA fit this description; both play a role in protein
synthesis.  In the case of DNA sequence which encodes protein, the
transcript produced is an RNA molecule used to convey information to
the cellular apparatus that synthesizes the protein.  This type of RNA
is called messenger RNA (mRNA), and it serves as an intermediate in
the transfer of information from genetic storage (as DNA) to protein
structure.  Messenger RNA contains a sequence of nucleotide triplets
complementary to the DNA sequence.  These triplets directly encode the
sequence of amino acids which form the linear structure of the
corresponding protein.  This process is called translation.

The genetic sequences which are transcribed are called genes (there is
a tendency by some writers to restrict the use of the word gene to
only those sequences which code for protein even though many genes
code for functional RNA molecules such as ribosomal RNA instead of
protein).  The original RNA copy formed during the transcription
process is called a primary transcript.  The primary transcript may be
used as is by a ribosome to direct the synthesis of a protein, or it
may require some modification prior to its translation into protein
structure.  In that case, the primary transcript is known as an
immature mRNA.  One step in the maturation of an mRNA is the removal
of chunks of sequence which do not correspond to protein structure.
These non-coding fragments, known as introns (originally known as
intervening sequences), are embedded in the RNA transcript.  These
fragments correspond to non-coding but transcribed sequences in the
DNA.  Introns are most common in eukaryotic genes, although they are
found in some mitochondrial and chloroplast genes.

The regions of the transcript which remain after maturation of the
messenger RNA are called exons.  Joined together after the introns are
removed, these exons generally encode the primary structure of a
protein but also may be functional RNA such as ribosomal RNA.  Since
there is a one-to-one correspondence between the original gene and the
primary transcript, it is convenient to refer to the original DNA sequence as
containing exons and introns.

The removal of introns from transcripts is called splicing.  The
details of the splicing process are not entirely understood,
particularly in higher eukaryotic species.  Simpler species use what
are called group I and group II introns.  In group I introns, the
excision of introns is generally done by the intron itself, which
acts as a catalyst with the assistance of a guanine-containing
nucleotide.  In group II introns, no external nucleotide is needed for
splicing, but the autocatalysis is still generally present.  In both
group I and group II splicing, the introns being removed by the
splicing operation apparently act as catalysts during their own
removal.  Group I introns have been observed in lower eukaryotic cells
and some mitochondria and chloroplasts mainly in ribosomal RNA genes,
while group II introns have been observed in yeast mitochondria.
Other categories of introns are also known to exist in which
splicing is mediated by much more elaborate complexes of protein and
RNA-based enzymes called spliceosomes.  Many of the details of the
more complex splicing operations (such as occur in human cells) are
not known.  Much of what is known was described by Cech in
\cite{cech86}.

\begin{figure}[htb]
\begin{center}
\includegraphics[]{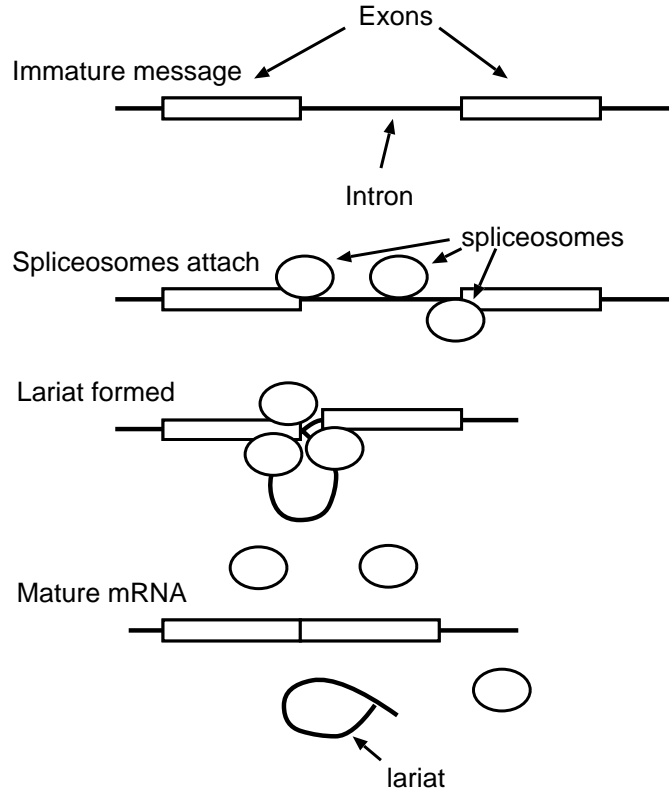}
\caption{Schematic of the splicing operation \label{fig:splicing}}
\end{center}
\end{figure}

A schematic diagram of one version of the splicing process is shown in
figure \ref{fig:splicing}.  Here, exons are illustrated by the boxes,
and introns by the thinner lines connecting the exons.  Initially, the
transcript consists of alternating sequences of exons and introns.
The spliceosomes attach themselves to the splice points and possibly
to each other and cause the removal of the intron, which forms a
looping structure called a lariat in the process.

It is clear that there must be some way for spliceosomes to recognize
the ends of an intron because errors in splicing are apparently
relatively rare.  In fact, examination of exon/intron boundaries shows
that there is a very high degree of sequence conservation at those
positions.  These conserved sequences apparently assist the
spliceosomes in recognizing the splice point.  At the beginning of
introns in the sense strand of the DNA, the sequence GT predominates,
and at the downstream ends of introns, the sequence AG occurs
predominantly, with the very occasional GG or CG.  In fact, an
inventory of the patterns at each end of the intron shows that several
other bases in the neighborhood of the intron/exon or exon/intron
boundary are not evenly distributed.  These uneven distributions are
reflections of relatively common motifs called consensus sequences
near splice sites and are well known.

One way that these consensus sequences can be illustrated is by
calculating the degree to which position relative to the exon/intron
boundary can predict which nucleotide is found there.  One useful
measure of the predictiveness is information content.  This approach
is similar to that used by \cite{white92}.

\begin{figure}[htb]
\begin{center}
\includegraphics[]{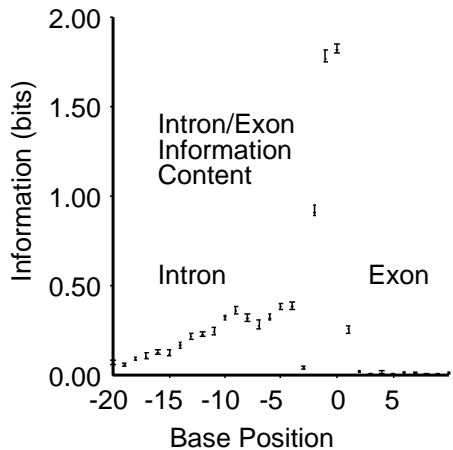}
\caption{Information near Intron/Exon Splice Site \label{fig:info1}}
\end{center}
\end{figure}

For instance, in human DNA, the distribution of nucleotides within 20
positions of the intron/exon boundary is distinctly uneven.  The
observed probabilities are approximately $p(A) = 0.23$, $p(C) = 0.24$,
$p(G) = 0.31$, $p(T) = 0.22$.  The amount of information conveyed by
the knowledge that we are dealing with a nucleotide with such a
distribution is, however, only about $0.015$ bits.  Adding information
about the position relative to the intron/exon boundary allows some
nucleotides to be predicted to some degree even outside of the almost
perfectly conserved GT consensus.  Figure \ref{fig:info1} shows the
amount of information given by position measured in terms of bits of
information for various positions relative to the boundary.  The peaks
at positions -1 and 0 represent the GT consensus, but there is a clear
deviation from the baseline at positions to the left of the boundary
(in the intronic sequence), and little systematic deviation from the
baseline to the right of the boundary (in the exonic sequence).

\begin{figure}[htb]
\begin{center}
\includegraphics[]{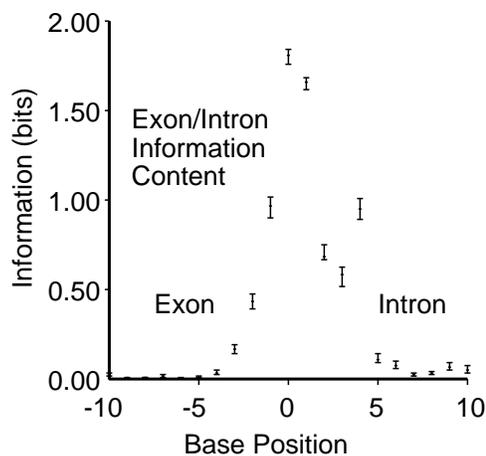}
\caption{Information near Exon/Intron Splice Site \label{fig:info2}}
\end{center}
\end{figure}

At the exon/intron boundary, as shown in figure \ref{fig:info2}, we
see some very small indication of conserved structure to the left of
the boundary (in the exon) in positions -4 through -1 as well as the
AG consensus at 0 and 1 and the weaker consensus in positions 2
through 4.

The data in these figures were calculated using maximum likelihood
estimators for the frequencies since the counts were relatively large.
Of particular interest are the error bounds.  Previous work on the
information content at these boundaries did not include the
calculation of accurate error bounds.  In
\cite{white92} a Monte Carlo method was used which gave error bounds
based on the assumption of a multinomial distribution, but the
calculation was very costly; several days of computer time on an
advanced workstation were required to compute the error bounds.

The figures above have error bounds which were computed using the
bootstrap method as described in \cite{efron82}.  The resulting error
bars do not depend on any distributional assumptions and take only a
second or so to compute for each data point even using a relatively
slow laptop computer.  Avoiding distributional assumptions has the
benefit that the error limits reflect the observed data.  If the
multinomial distribution assumption which was used in
\cite{white92} is substantially incorrect, then the error bounds
computed using that method will be substantially smaller than they
should be leading to over-confidence in the derived numbers.

Unfortunately, the patterns found at the ends of introns by these
information based methods are not distinctive enough to tell the whole
story of splicing in higher organisms.  Not only do they represent a
good deal of variation, but these consensus sequences appear elsewhere
in both introns and exons.

Even though the mechanism used during splicing introns is not well
understood in detail, it is reasonable to suggest that there is still
some correspondence or dependency between the ends of the intron since
it is eventually formed into a looping structure during splicing.  If
there were a relatively small number of consensus sequences recognized
during splicing, or if most introns were evolutionarily related to a
small number of ancestral introns, then there would tend to be a
correlation between nucleotides in different positions in the intron.

In addition to expecting some correlation between the ends of introns,
it is reasonable to expect that there would be little or no
correlation between the ends of the exons separated by introns or
between the proximate ends of successive introns.  For example,
although the work described in \cite{peng}, \cite{voss},
\cite{dreismann}, \cite{wli92a} and \cite{wli92b} found long range
correlations in exonic regions, there were also some indications in
this work that these correlations might exist only in exonic sequence
which lacked introns.  This observation is interesting because it has
been hypothesized that introns provide a mechanism for the eukaryotes
to reuse helpful functional domains.  This would be done by the
reassembly of functional domains using introns as ``glue''.  If the
exons on each side of an intron are essentially interchangeable parts,
then little correlation between their ends would be expected.
Furthermore, the very flat floor on the information plots in the
exonic region would tend to indicate that as much as possible of the
available information bandwidth is being used to encode protein structure.

There has been considerable speculation about the origin of introns,
with some claiming that they must be holdovers from an earlier world
in which RNA was dominant \cite{orgel,rnaworld} and they disappeared in
prokaryotes due to selective pressure.  Others claim that introns
arose relatively late, after the split between prokaryotes and
eukaryotes.  Indications of structure in and around introns such as
can be found using the techniques described in this section may have
implications regarding this controversy.

An effective and sensitive way to test for any hypothesized dependency
is to extract pairs of nucleotides from a number of introns or exons
and test them for correlation using a log-likelihood ratio test.  The
elements of the pairs would be sampled from various positions relative
to the intron/exon junction.  The result of this sampling is a four by
four contingency table which can be analysed by the methods described
in section \ref{methods:lr_tests}.  For example, table
\ref{table:single_position} shows the counts obtained at locations 3
bases to the left of the 5' splice site and 4 to the right of the 3'
splice site for human intron sequences.  By varying the positions of
the samples, certain aspects of the structure of introns can be
determined.  Gross correlation between regions can be demonstrated by
computing the correlations between every pair of locations in each
region and comparing the cumulative distribution of the resulting test
statistics to the theoretical $\chi^2(9)$ distribution or by a control
constructed by resampling the original data.  The resulting deviation
from the expected distribution is a measure of the structure which is
found in the sequences.  This chapter explores the use of this measure
of regional correlation in a number of different species.  The
analysis provided here highlights structure where the intron/exon
junction is the primary landmark, but other studies with other
landmarks are also possible.

\section{Experimental Methods}

\subsection{Data Preparation}

In order to examine the structure associated with introns in various
organisms, all introns bordered on both sides by intact exons were
extracted from Genbank for {\em Caenorhabditis elegans} (a nematode or
roundworm), {\em Dro\-soph\-ila melanogaster} (fruit fly), {\em Homo
sapiens} (human), {\em Gallus domesticus} (domestic chicken) as well
as for a group of dicotyledonous plant species (largely {\em Nicotiana
tabacum} (tobacco), {\em Lycopersicum esculentum} (tomato), {\em
Alfalfa meliloti} (alalfa), and {\em Arabidopsis thaliana} (the
flowering plant with the smallest genome)), and a group of several
monocotyledonous plant species (mostly {\em Zea mays} (corn) and {\em
Oryza sativa} (rice)).

\begin{table}[htb]
\begin{center}
\begin{tabular}[h]{|ll|}
\hline
\bf{Organism} & \bf{Size of Genome} \\
\hline
{\em C. elegans} & $8 \times 10^7$ \\
{\em A. thaliana} & $7 \times 10^7$ \\
{\em Z. mays} (corn) & $5 \times 10^9$ \\
{\em A. cepa} (onion) & $1.5 \times 10^{10}$ \\
{\em D. melanogaster} & $1.65 \times 10^8$ \\
{\em H. sapiens} & $2.9 \times 10^9$ \\
{\em G. domesticus} & $1.2 \times 10^9$ \\
{\em S. cerevisiae} & $1.35 \times 10^7$ \\
\hline
\end{tabular}
\end{center}
\caption{Sizes of Complete Genomes (in base pairs in haploid genome)}
\end{table}

All exons bordered by complete introns were also extracted for each
organism under consideration.  The complete intron (or exon) was
retained as well as 30 bases on each side of the splice sites.  Any
entries which did not meet a small set of sanity checks were
eliminated.  These checks included deletion of introns of zero length.

Duplicate sequences were eliminated from the dataset using the same
method described in chapter \ref{bio_id}.  Eliminating all near
duplicates in this manner runs the risk of eliminating sequences which
actually are duplicated in multiple locations in the genomes.  Since
it is impossible to determine accurately which entries in the database
reflect the same sequence locations, such a conservative strategy of
eliminating all possible near duplicates was the only safe option.
The effect of the duplicate removal process should be to decrease any
observed structure making the overall analysis very conservative.

The resulting sequences after eliminating duplicates were collected
into separate files for the intron and exon centered sequences for
each of the eight organisms.  The number of sequences and their sizes for
the various organisms are shown in Table \ref{table:sequence_counts}.
\begin{table}[htb]
\begin{center}
\begin{tabular}[]{|llll|}
\hline
\bf{Organism}&\bf{Raw Sequences}&\bf{No Duplicates}&\bf{Total Size (bp)} \\
\hline
{\em C. Elegans} & 1267(E), 1589(I) & 1225(E), 1527(I)& 431K(I) 460K(E)\\
{\em D. melanogaster} & 655(E), 1027(I) & 498(E), 758(I)& 314K(I) 272K(E) \\
{\em G. domesticus} & 435(E), 527(I) & 379(E), 471(I) &254K(I) 80K(E)\\
{\em H. Sapiens} & 2415(E), 3417(I) & 1835(E), 2797(I)&2279K(I) 422K(E) \\
Monocots & 550(E), 752(I) & 475(E), 679(I)& 221K(I) 143K(E) \\
Dicots & 1149(E), 1683(I) & 1081(E), 1587(I)&536K(I) 289K(E) \\
\hline
\end{tabular}
\end{center}
\caption{Amount of data used in analysis\label{table:sequence_counts}}
\end{table}

\subsection{Data Analysis}

Data analysis was performed by creating a contingency table for each
pair of positions from the left and right ends of each sequence.  Only
bases within 30 nucleotides of the splice site were considered.  As an
example, the table for bases 3 to the left of the 5' splice site and 4
to the right of the 3' splice site for the human intron-centered
sequences is shown in table \ref{table:single_position}.

\begin{table}[h]
\begin{center}
\begin{tabular}[h]{|l|llll|}
\hline 
&\bf{A}&\bf{C}&\bf{G}&\bf{T} \\
\hline
\bf{A} & 212 & 224 & 298 & 188 \\
\bf{C} & 238 & 267 & 421 & 139 \\
\bf{G} & 134 &  85 & 180 & 104 \\
\bf{T} &  82&   64&  122&   39 \\
\hline
\end{tabular}
\end{center}
\caption[Base pair counts at opposite ends of an intron]{Base pair counts show distinct, but weak dependency between
bases 3 to the left of the 5' splice site (row labels) and 4 to the
right of the 3' splice site (column labels) for human introns.  Thus, a
G at the 3' and an A at the 5' end was found 134
times. \label{table:single_position}}
\end{table}

It can be seen from this table that if the leftmost base of the pair
is C or T, then the frequency of T in the rightmost base of the pair
is substantially decreased.  The log-likelihood ratio test gives a
score of 48.03 for this table.  If the two positions are independent,
then such a high value would be highly unlikely.  In fact, this value
is significant at a level of $p < 0.000005$.  Note that many of these
tables have expected cell frequencies too low to allow safe
application of traditional $\chi^2$ tests.  This makes the use of the
log-likelihood ratio crucial in this analysis.  The log-likelihood
ratio can be used as a measure of dependency for the reasons described
in section \ref{methods:chi2_comparison}.

The average mutual information for this table is about 0.01 bits which
indicates that while the degree of observed correlation is highly
unlikely to have occurred by chance, the value of knowing one base
from a fixed position is quite low in the context of trying to predict
the second base (if the mutual information were 2, then knowing one
base would allow the other to be determined without error).  Single
cell mutual information cannot be used in this case since there is no
distinguished cell.  This is in strong contrast with word coocurrence
problems where the resulting $2 \times 2$ contingency table at least
has one distinguished cell (the case where both words are present).
The table shown here is fairly typical of the tables obtained from the
human sequences in this study.

There are also other possible explanations for the observed small
degree of mutual information.  For instance it is possible that a very
strong relationship between bases (i.e. high mutual information) would
have been found if slightly different positional reference points had
been chosen.  The observed mutual information could then be a remnant
of a strong correlation which is almost entirely hidden by the
smearing caused by shifts relative to the positional references.
Another possibility is that there are many factors which jointly
determine the base composition, and we are examining them separately
and thus see only a shadow of the real correlations.  Regardless of
the cause, the correlation shown later is very strong.

\subsection{Control Cases}

\begin{figure}[htb]
\begin{center}
\includegraphics[]{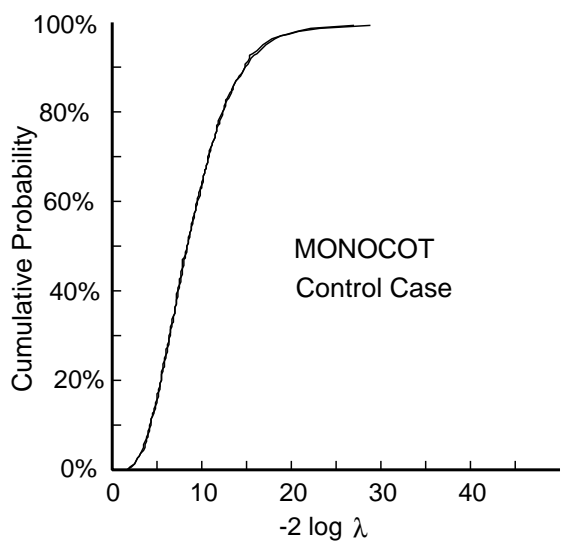}
\caption[Control and Theoretical Distributions]{Control and Theoretical Distributions.  Note that the two
plots are virtually indistinguishable. \label{fig:intron_control}}
\end{center}
\end{figure}

The contingency table approach allows the exploration of the
dependency of single bases on each other.  Plotting the cumulative
distribution of these scores for specific regions allows the
dependency of one region on another to be examined, a task which
cannot be done directly using contingency tables.  If all of the bases
in the two regions were independent, then these scores should be
$\chi^2$ distributed with 9 degrees of freedom.  A test of this
theoretical distribution can be performed by analysing a synthetic
dataset constructed by randomly associated left and right halves of
real sequences.  A cumulative distribution obtained in this way from
the exonic tails of the monocot intron-centered sequences is shown in
figure \ref{fig:intron_control}.  This plot is typical of the control
cases for all organisms examined.  A Kolmogorov-Smirnov test of this
distribution shows that there is no significant difference from the
theoretical $\chi^2(9)$ distribution.  Based on this result, the
theoretical $\chi^2(9)$ distribution was used as the reference in the
analyses presented in this section instead of the synthetic control
data set.

\section{Experimental Results}

The first important result found in this study was that the results
obtained by Kinkead showing structure in human DNA and reported in
\cite{kinkead} were replicated even after a rigorous screening for
potential duplicate sequences was done.  Since the earlier study did
not screen for duplicates, there was a serious potential that the
earlier results were simply the result of duplicated sequences.

More importantly, structure similar to that found in humans was found
in a variety of other organisms including chicken, fruit flies, dicots
and monocots, but not as strongly in {\em C. elegans}.  This finding
substantially extends the previous work in this area in which only
human sequences were examined.  Detailed examination of the
contingency tables involved shows that this dependency does not appear
to be simple complementary base pairing, but rather more subtle
relationships are involved.

In spite of the fact that structure was detected strongly in 5 out of
the 6 organisms examined, there were substantial differences in the
degree of correlation observed.  Human DNA showed, by far, the largest
amount of structure.  Flowering plant DNA (monocots and dicots) also
had large amounts of structure.  DNA from {\em C. elegans} had dramatically
less observed structure than the other organisms, although even in
this case, the observed structure was statistically significant at
better than a $10^{-4}$ level in the weakest case.

\begin{figure}[htb]
\begin{center}
\includegraphics[]{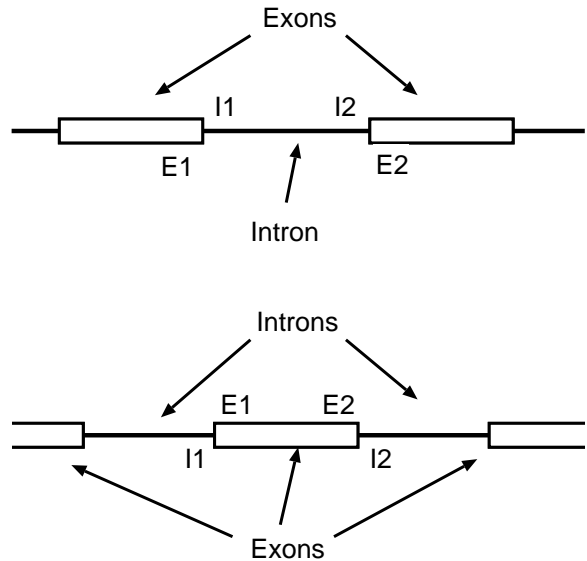}
\caption{Key to comparisons shown in figures \label{fig:splice_names}}
\end{center}
\end{figure}

\begin{figure}[htb]
\begin{center}
\includegraphics[]{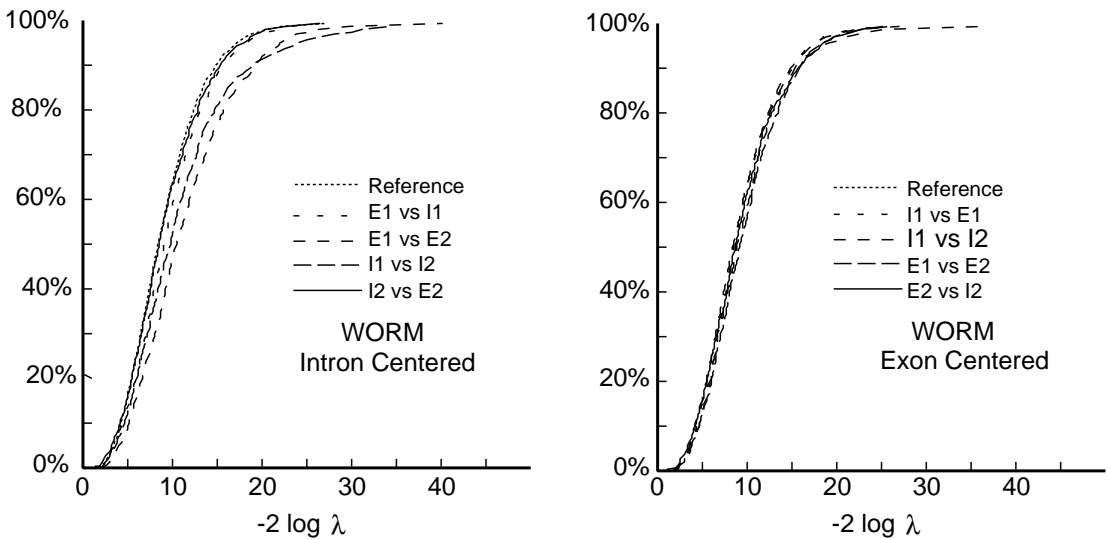}
\caption[Medium range structure in worm genome]{Worm genome.  See diagram in figure \ref{fig:splice_names}
for clarification of the comparisons. \label{fig:worm}}
\end{center}
\end{figure}

Figure \ref{fig:splice_names} shows the key for the comparisons which
were plotted in figures \ref{fig:human_cum} through \ref{fig:worm}.

A substantial degree of structure for different organisms can be seen in figures
\ref{fig:human_cum} through \ref{fig:monocot} in which the cumulative
distributions for the various categories of sequence are shown.

\begin{figure}[htb]
\begin{center}
\includegraphics[]{human.eps}
\caption[Medium range structure in human genome]{Human genome \label{fig:human_cum}.  See diagram in figure \ref{fig:splice_names}
for clarification of the comparisons.}
\end{center}
\end{figure}
\begin{figure}[htb]
\begin{center}
\includegraphics[]{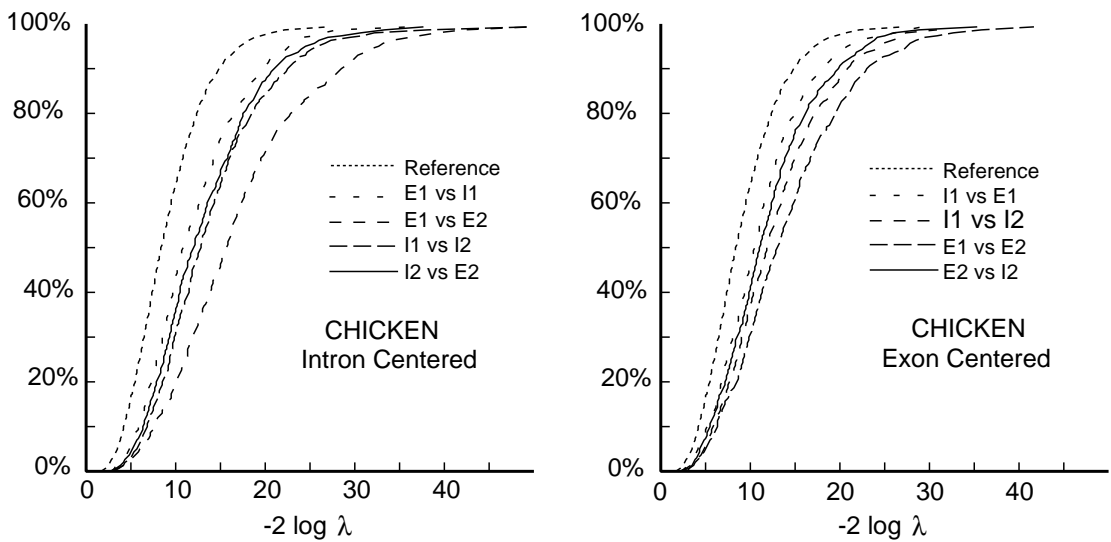}
\caption[Medium range structure in chicken genome]{Medium range structure in chicken genome.  See diagram in figure \ref{fig:splice_names}
for clarification of the comparisons. }
\end{center}
\end{figure}
\begin{figure}[htb]
\begin{center}
\includegraphics[]{fly.eps}
\caption[Medium range structure in fly genome]{Fly genome.  See diagram in figure \ref{fig:splice_names}
for clarification of the comparisons.}
\end{center}
\end{figure}
\begin{figure}[htb]
\begin{center}
\includegraphics[]{dicot.eps}
\caption[Medium range structure in dicot genomes]{Dicot genome.    See diagram in figure \ref{fig:splice_names}
for clarification of the comparisons.}
\end{center}
\end{figure}
\begin{figure}[htb]
\begin{center}
\includegraphics[]{monocot.eps}
\caption[Medium range structure in monocot genomes]{Monocot genome.    See diagram in figure \ref{fig:splice_names}
for clarification of the comparisons. \label{fig:monocot}}
\end{center}
\end{figure}

On the other hand, in {\em C. elegans}, there was almost no structure
visible on the cumulative distribution diagram.  This can be seen in
figure \ref{fig:worm}.  

\begin{figure}[htb]
\begin{center}
\includegraphics[]{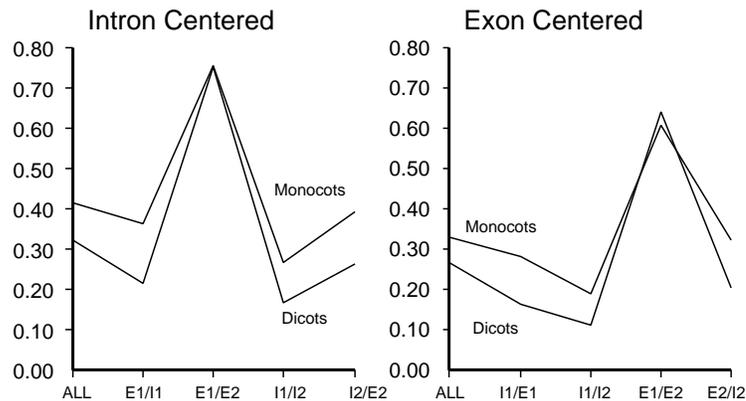}
\caption{Measures of structure in Angiosperms (flowering plants) \label{fig:cots}}
\end{center}
\end{figure}

\begin{figure}[htb]
\begin{center}
\includegraphics[]{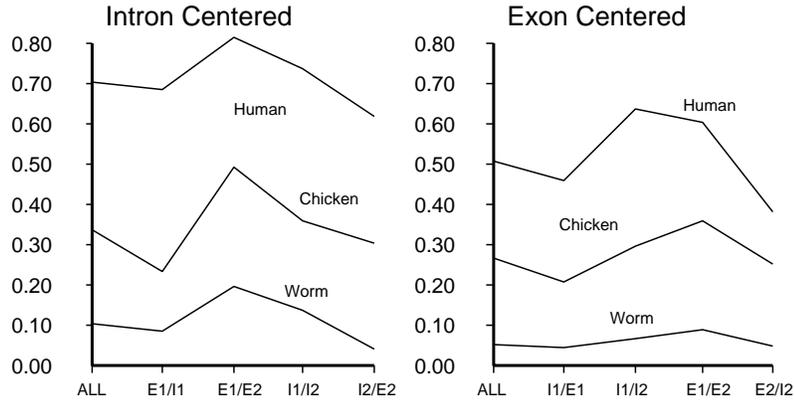}
\caption{Measures of structure in Human, Chicken and {\em C. Elegans} \label{fig:misc_struct}}
\end{center}
\end{figure}

\begin{figure}[htb]
\begin{center}
\includegraphics[]{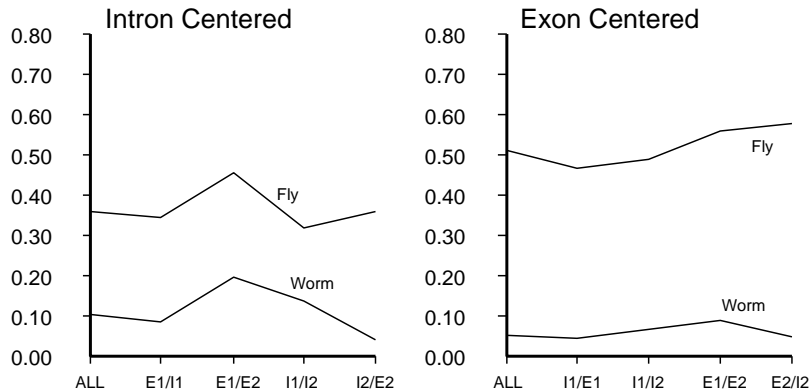}
\caption{Measures of structure in {\em D. melanogaster} and {\em C. Elegans}
\label{fig:worm_fly}}
\end{center}
\end{figure}

In figures \ref{fig:cots} through \ref{fig:worm_fly}, the
Kolmogorov-Smirnov statistic is used to summarize the degree of
structure detected.  This statistic is equal to the largest vertical
deviation between the reference $\chi^2(9)$ distribution and the
cumulative distribution from the previous diagrams.  The
Kolmogorov-Smirnov statistic is normally used as part of a
distribution free test for whether an observed distribution is the
same as a theoretical distribution.  Here, it is used instead as a
measure of how different an observed distribution is from a
reference.

The Kolmogorov-Smirnov statistic is plotted for the combinations used
in figures \ref{fig:human_cum} through \ref{fig:worm} and all left
side positions versus all right side positions (ALL).  The points are
connected only to make the relation between the various values
apparent.  These plots highlight the relative magnitude of the
structure observed between elements.

It is clear from figure \ref{fig:cots} that the pattern and magnitudes
of detected structure within angiosperm sequences (monocots and
dicots) are quite similar despite the great diversity between monocots
and dicots.  The marked dependency of E1--E2 in the intron centered
case for these plant species is surpising, as is the observation that
with the exon-centered comparisons, the dependency E1--E2 is also the largest.

In figure \ref{fig:misc_struct} it can be seen with {\em
G. domesticus} (chicken), {\em H. sapiens} (human) and {\em
C. elegans} (worm) that there is a similar pattern in the relative
magnitudes of the dependency detected by the tests described here.
For example, the ordering of the measure of structure in the
intron-centered case is generally (from largest to smallest) E1--E2,
I1--I2, E1--I1 and finally I2--E2 for all three of these diverse
species.  This pattern indicates that coherence in exonic structure
even across the intronic gap is stronger than the coherence between
the ends of the intron and much stronger than the coherence between
intronic and exonic sequence.  Regardless of this repeated motif,
however, there is a large discrepancy in the magnitude of the
structural dependency between these organisms.

In figure \ref{fig:worm_fly} it can be seen that the structural
patterns in {\em D. melanogaster} are similar to the very
much weaker structure patterns found in {\em C. elegans}.  This
commonality is striking given the taxonomic distance between these two
organisms.

\section{Conclusions}

Enormously statistically significant structure was observed in exon
and intron comparisons for a variety of species.  The biological
implications are not entirely clear.  It is quite possible that these
data have implications regarding whether introns were an early feature
of DNA based life or whether they were a late innovation on the part
of eukaryotes.  In particular, substantial correlation between
successive exons was found.  This finding can be viewed as contrary to
the idea that exons are separable and independent functional units
which can be reassembled relatively freely.  In addition, strong
correlation was noted between intron and exon sequences across the
splice site.  If introns appeared late, then this correlation could be
due to there being preferred insertion points for introns.  Most
importantly, though, very substantial correlation was seen between
successive introns.  This would be {\em very} surprising if successive
introns are completely independent.  The strong controls
used in this experiment preclude these correlations being due to
multiple copies of the same intron.

Further analysis and experimentation is likely to be necessary to
clarify the correlations which have been observed.  It is of
particular interest that there is a rough correlation between the
overall degree of structure observed and the complexity of the
organism involved.  Examination of similar structure in very simple
eukaryotes such as the {\em Tetrahymena} as well as of more varieties
of complex organisms would be very interesting.  As more genomic
sequences become available from plants, it would also be interesting
to compare the degree of structure in individual species rather than
grouping all monocots and all dicots together.

\part{Discussion}
\chapter{Potential for Further Application \label{future}}

This chapter describes some of the potential for additional
applications where log-likelihood ratio tests appear to be useful.
These cases are presented as a brief survey, unlike the detailed
evaluation presented in the previous chapters.

\section{Finding Interesting Word Pairs \label{word_pairs}}

Chapter \ref{colloc} demonstrated that log-likelihood ratio tests
could efficiently detect words which appear near each other with
anomalously high frequency.  Finding words which appear near each
other more than is expected is quite a popular intuitive approach for
finding phrases.  The log-likelihood ratio framework provides other
alternatives, however, for performing this task.

For example, in recent work by this author done with the assistance of
Paul Miniero, an alternative algorithm for finding phrases based on a
new log-likelihood ratio test was partially evaluated.  This test
looked for interesting word pairs by comparing the coocurrence profile
of the pair with the product of the profiles of the individual words
in the pair.  When the profile of the pair was significantly different
from the prediction, then the pair is marked as interesting.  A null
hypothesis for finding pairs this way can easily be constructed using
the log-likelihood ratio framework.

In a preliminary step to test this new method for finding interesting
word pairs, a set of approximately 4,000 word pairs was used.  These
word pairs were selected from all of the several million word pairs
which appeared in the TREC documents.  The instructions to the people
doing the selection were that they were to keep only word pairs which
would act substantially like a single unit and which had a meaning
different as a pair than might be induced by the composition of the
individual words.  Subsequent evaluation of this list has shown that
the precision of the list was relatively good, but that the recall was
relatively low.  In particular, since the word pairs were originally
examined in decreasing frequency order, there was a substantial bias
towards the pairs with higher frequencies.

Next, a subset of approximately, 18,000 documents from the 1990 AP
documents found in the TREC corpus was examined automatically, and a
variety of scores was computed for each of the million or so word
pairs found in these documents.  Cumulative distributions were plotted
for the raw frequency, the log-likelihood ratio test described in
chapter \ref{colloc} and the log-likelihood ratio test described here.
As might be expected based on how the reference set of word pairs was
constructed, raw frequency proved to be a reasonable way to find
members of the reference.  The original log-likelihood ratio test was
also able to find interesting word pairs.  The new product profile
test described here was much more effective than either of these other
tests.

%

These results indicate that this test may provide a much more
effective method than earlier methods which simply found cases of
anomalously large coocurrence frequencies.  Further evaluation of this
method should include testing in an information retrieval system as
well as manual spot testing of word pairs in order to estimate the
relative precision and recall of this new method.

\section{Improvements to Luduan System \label{luduan_future}}

There are a number of changes which might make the Luduan system
described in chapter \ref{ir} more effective as a document routing
system.  For example, the use of stemming may have either a positive
or negative net impact on performance depending on whether the
generalization behavior that stemming provides is useful in a
particular query.  Most retrieval systems are designed in such a way
that they must commit to either using stemming or not using it.
Furthermore, most retrieval systems benefit, on average, by using
stemming.  Luduan is unusual, however, in that it can select which
words to stem and which to keep in their original form.  This can be
done by using both stem and word counts during the statistical query
generation process.

This innovation can be extended further.  One option would be to use a
lexical clustering scheme such as was described in \cite{brown92a} and
elsewhere.  In this approach, the set of all words is hierarchically
clustered so that each word can be described by a variable length
bit-string.  Furthermore, various sets of words which appear in
similar contexts can also be described using the prefix common to each
member of the set.  To be used with Luduan, each word would be
considered an occurrence both of the surface form as well as the
bit-strings which are prefixes of that word.  The statistical query
generation process can then select among the various word forms and
word-set names to find those which are particularly distinctive of
relevant documents.  

It is reasonable to expect that if Luduan chooses to include word-sets
in the final query, then these word-sets will help final retrieval
quality.  Additional passes of human feedback might be required to
prevent over-generation using this method, but ultimately the process
should converge on those features which are characteristic of relevant
documents, while not representative of non-relevant documents.  The
experimental question then, is whether or not such meta-words would
actually be of any utility in distinguishing relevant from
non-relevant documents.  If they are, then the system should be able
to generalize better than if it is restricted to words along.

Going the other way towards more specificity instead of towards more
generalization, Luduan can select short phrases as well as individual
words.  The normal reason for including phrases is to increase
precision.  At lower levels of recall, however, Luduan based systems
already produce nearly perfect precision.  This high level of
performance makes improvements due to using phrases essentially
impossible.  The effect of selecting phrasal features is therefore
difficult to predict, although it is hard to believe that they would
hurt performance.

It is tempting to use Luduan to compute weights for terms as well as
for selecting terms rather than using an existing system to compute
weights based on global corpus characteristics.  The Berkeley TREC
results mentioned in section \ref{ir_overview} stand as a cautionary
flag on this path.  The problem of over-fitting is a serious obstacle
for this alternative since allowing the routing system to pick weights
on all of the possible query terms increases the number of degrees of
freedom available to the system very substantially.  Once the number
of degrees of freedom is large enough, the system becomes
under-determined, and over-fitting is an almost inevitable consequence.
Such over-fitting will almost inevitably lead to a decrease in the
generalization performance of the system.

\section{Fisher's Exact Statistic as a Log-Likeli\-hood Ratio}

Fisher's exact method \cite{fisher,agresti} for evaluating the
significance of contingency table test statistics uses a hypergeometic
probability to compute the likelihood of generating a particular
contingency table which specified row and column marginal sums.
Fisher's exact statistic is a measure of the significance of the value
of any other test statistic such as Pearson's $\chi^2$ test.
This significance is defined as the probability of other tables which
have the same marginal totals, but which have more extreme values of
the test statistic.

Interestingly, if Stirling's formula is used to approximate the
factorials in Fisher's probability, then the logarithm of the quantity
obtained is very nearly identical to the log-likelihood ratio
statistic for the table.  This identity indicates that the
log-likelihood ratio may be a useful shortcut in the computations
required by Fisher's exact method.  Since the computations involved in
Fisher's methods rapidly become intractable for large tables, such a
short cut may provide an attractive alternative to the stochastic
approximation methods which are normally used for large tables.  Since
the cost of computing the log-likelihood ratio statistic is relatively
small, substantial savings in computational complexity may be
available.

This identity may also provide insight into the situations where the
log-likelihood ratio is likely to work well and situations where it is
not likely to be useful.  As was seen in section
\ref{methods:chi2_comparison}, the log-likelihood ratio consistently
(although moderately) underestimates the significance of many
contingency tables with small counts.  This underestimate might be
understandable if, in fact, the cumulative probability of Fisher's
method is a more accurate estimate and if the logarithms of the
individual probabilities in Fisher's are well approximated by the
log-likelihood ratio.

\section{Iteratively Constructed Markov Models}

Generalized likelihood ratio tests can be used to construct Markov
language models such as are used in speech recognition.  Two methods
for doing this are described here.  The first method is an extension
of the method of backoff models such as were first described by Katz
\cite{katz}.  The second method produces models which are similar to
the interpolated $n$-gram models used by Brown and Mercer
\cite{brown92}.  Each of these methods can be evaluated using the
standard measures such as perplexity as is customarily done in the
speech recognition community.

\subsection{Iteratively Devised Mixed Order Markov Models \label{methods:iterative_models}}

When the conditional probability parameterization for Markov models is
used, it is easy to see that if $m < n$, any order $m$ model can be
written as an order $n$ model by taking
\begin{equation*}
p(\sigma_{n+1} \mid \sigma_1 \ldots \sigma_n) =
p(\sigma_{n+1} \mid \sigma_{n-m+1} \ldots \sigma_n)
\end{equation*}

These relations can be considered to be a set of very simple linear
constraints.  Relaxing some of these constraints while retaining
others is essentially what is done in backoff models.
There are, however, other criteria which can be applied so that as
many constraints can be eliminated as possible without encountering
over-fitting.  One algorithm for building a model in this way is
described below.  This approach to text modeling is, as far as I
know, novel.

In fact, the log-likelihood ratio test for multinomials can tell us
when a higher order or mixed order Markov model is justified by the
data.  This test is described more fully in section
\ref{methods:lr_tests}.  Using this test, the null hypothesis
that a higher order Markov model will not help model a training
sequence can be tested.  

This log-likelihood ratio test can be incorporated into a greedy
algorithm which produces a mixed order Markov model.  Starting with
the zero order Markov model, all of the possible extensions to the
model which consist of adding one symbol to any of the known contexts
are tested.  Whichever extension has the best log-likelihood ratio
measure relative to the unextended model is chosen.  The
log-likelihood ratios are the same as the changes in the log of the
likelihood of the observed data which means that this algorithm is a
steepest descent algorithm which attempts to find the maximum
likelihood estimator.  By only taking steps which represent
substantial improvements in log-likelihood, this algorithm can avoid
deriving an over-fitted model without the cost of the optimization step
in the held out smoothing.

Assuming that the alphabet of symbols is $\Sigma$, the current set of
contexts is $\mathcal C$, and the training string is $S$, this algorithm can be
described as
\paragraph{S0:}
set ${\mathcal C}$ to the empty set, $\emptyset$.  This corresponds to
the order zero Markov model (which is a multinomial model) which is
described by the probabilities $p(\sigma), \sigma \in \Sigma$.
\paragraph{S1:} 
for each candidate context $\sigma c$ in the set $\Sigma \times {\mathcal
C}$, set
\begin{equation}
\delta (\sigma c) =
-2 \sum_{w \in \Sigma} T(\sigma c w, S)
\left( \log \frac {T(\sigma c w, S)} {T(\sigma c, S)} -
       \log \frac {T(c w, S)} {T(c, S)} \right)
\end{equation}

This is the same as testing the following $|\Sigma| \times 2$
contingency table using the log-likelihood ratio tests described in
section \ref{methods:lr_tests}.

\begin{center}
\begin{tabular}{r r@{} l c c c c c l@{} l}
\hhline{~~-------}
&& \vline & $T(cw_1, S)$        & \vline & $T(cw_2, S)$        & \vline & $\ldots$ & \vline \\
\hhline{~~-------}
&& \vline & $T(\sigma cw_1, S)$ & \vline & $T(\sigma cw_2, S)$ & \vline & $\ldots$ & \vline  \\
\hhline{~~-------}
\end{tabular}
\end{center}
\paragraph{S2:} 
if $\max_c \delta (c) > \epsilon$ then set ${\mathcal C}$ to be ${\mathcal C}
\cup \{ \argmax_c \delta(c) \}$ and go to S1.  (i.e. add the context
which causes the best improvement)
\paragraph{S3:} 
no more significant improvements are possible.  The final model
consists of the conditional probabilities $p(\sigma \mid c)$ where
$\sigma \in \Sigma$ and $c \in {\mathcal C}$
\paragraph{}
In this algorithm, $\epsilon$ determines how much over-fitting is
allowed.  If $\epsilon = 0$, then this algorithm will construct a
model which can produce no other string except $S$ because eventually
$\mathcal C$ will contain all possible partial prefixes of $S$.  A
$\epsilon$ is increased, successively less over-fitting is allowed, and
the models produced become more and more generalized.  For some
critical value, the algorithm will terminate, leaving the original
multinomial model unchanged.  Between these extremes, mixed order
multinomial models should be produced which fit the training text
well, but which also generalize well.  The resulting mixed order
Markov model will be structurally similar to the backoff models
conventionally used in speech recognition work, but the decision to
back off the order of the model will be made in a more principled
manner.  For example, if a higher order model does not provide any
predictive power, it is quite possible that the construction algorithm
described here would opt for a simpler model, while conventional
techniques might opt for a more complex model simply because the
number of observed $n$-grams was high enough.  This would incur a space
penalty on the part of the back-off model as well as potentially
decrease performance.  On the other hand, the algorithm presented here
could easily opt for a more complex model based on a small amount of
evidence if the available evidence clearly contradicted the
predictions of a simpler model.

\subsection[Quasi-Bayesian Mixed Order Models]{Quasi-Bayesian Estimates for Mixed Order Multinomial Models}

In many cases, it is possible to use Bayesian estimates for the
conditional probabilities to develop a model which has many of the
properties of the mixed order multinomial models produced by held out
smoothing or the iterative process described in the previous section.

In this method, multinomial models of several different orders are
derived from the training text using estimators similar to Bayesian
estimators.  The log probability of a test string is estimated by
averaging the estimates from each of these models.  The result is that
where the higher order models have insufficient information, they will
tend toward a neutral value and thus not contribute much to the
overall estimate.  On the other hand, where the higher order models
are able to produce a good estimate, they will tend to produce a much
sharper estimate than the lower order models and thus will allow the
composite model to distinguish similar strings that the lower order
models would not be able to distinguish.

This method of interpolation by Bayesian estimation can be taken
further by recursively taking the prior estimator for the parameters
of the order $k$ model to be the estimator of the order $k-1$ model.
This procedure is indefensible from the standpoint of strict
statistical argument, but it has considerable intuitive appeal.  In
this approach, the parameters of the order $k$ model would be defined
as

\begin{equation}
p(w_i \mid w_{i-k}^{i-1}) =
\frac
{T(w_{i-k}^{i}) + m p(w_i \mid w_{i-k+1}^{i-1})}
{T(w_{i-k}^{i-1}) + m}
\end{equation}
where $m$ is the size of the vocabulary.

In general, this form is much too conservative because it requires
that we see $m$ or more examples in a particular context before giving
it much credence.  Since we {\em know}, in many cases such as human
language, that the distribution of symbols will be distinctly uneven,
we also know that substantial information about the correct values of
the parameters dealing with common words can be had long before we
have enough information to deal with rare ones.  Adding an additional
parameter $0 \le \alpha \le 1$ allows us to adjust the conservativism
of this algorithm.  The parameters thus become,

\begin{equation}
p(w_i \mid w_{i-k}^{i-1}) =
\frac
  {T(w_{i-k}^i) + \alpha m p(w_i \mid w_{i-k+1}^{i-1})}
  {T(w_{i-k}^{i-1}) + \alpha m}
\end{equation}

The additional parameter $\alpha$ can be adjusted automatically by
held out smoothing, or on the basis of experience with a particular
type of data.  It is unlikely that this form of model will actually
outperform a model based on the full form of held out smoothing, but
the computational requirements are substantially less than for
held out smoothing.

A natural value for $\alpha m$ is the average perplexity of the data
stream relative to the best model constructed so far.  This value is
an estimate of the effective vocabulary at any given point.  Even
better is to estimate the average perplexity using the order $k-1$
model.  This gives the following as the value for $\alpha m$:

\begin{equation}
\alpha m = e^{H(w_{i-k+1}^{i-1})}
\end{equation}
where
\begin{equation}
H(w_{i-k+1}^{i-1}) =
  \sum_\sigma p(\sigma \mid w_{i-k+1}^{i-1})
        \log p(\sigma \mid w_{i-k+1}^{i-1})
\end{equation}
This perplexity can be estimated using maximum likelihood methods, or
it can be evaluated on held-out data using the methods described in
chapters \ref{lingdet} and \ref{bio_id}.

\section{Constructing Lexica Using MDL Methods}

Carl de\ Marcken showed that the minimum description length principle
(MDL) could be useful in the construction of text segmenters for
English as well as Asian languages such as Chinese \cite{demarcken}.
In this work, a lexicon is constructed which is evaluated based on how
well the lexicon allows text to be compressed.

The close relationship between MDL and generalized log-likelihood
ratio tests was described in section \ref{methods:mdl}.  This close
relationship can be used to speed up the construction of lexicons
using methods similar to deMarcken's.

The benefits of such work might be a truly language independent
information retrieval system which induces suffix stripping rules or
text segmentation rules directly from the corpus under analysis.  The
suitability of the segmentation of the text based on human judgement is
not of particular concern since the use of the segmentation would be
entirely internal to the system.  

A particular virtue of this system would be its ability to
adapt to specialized jargon automatically.  Tokenization of text
containing specialized terms such as those found in electronics or
finance is notoriously difficult.  For example, it is generally not
useful to retrieve documents based on their containing the number 458,
but documents containing the number 486 are very likely to relate to
microprocessors.  Similarly, it might be very bad to consider the
string ETC-104 as the token ``ETC'' (very similar to an abbreviation
generally ignored by retrieval systems) followed by the number 104
(also generally ignored by retrieval systems).  The automatic
detection of such cases would be of great potential value to those
delivering information retrieval systems into a wide variety of highly
specialized user communities.

%

\chapter{Summary and Conclusions}

The huge and growing quantity of text and genetic sequence data
has created an urgent need for efficient and practical methods for
analyzing the structure in symbolic sequences.  Not only is there a need
for recognition and sorting of specific structures, but there is also a need
for methods that help us to understand and quantify the inherent
structure in symbolic information.  

The log-likelihood ratio tests developed and demonstrated in this
thesis have wide applicability in the development of real world
software.  The application examples presented here use these methods
to achieve results which are either competitive with or superior to
the best known alternative methods.  More importantly, though, is the
fact that the tests derived here are merely exemplars of a general
class of methods which can be used in an even wider range of
situations.  The examples shown here form a far from exhaustive
catalog of what can be done with this class of statistical method.
New tests can be developed using the techniques described here with
relative ease, and it is to be expected that many novel applications
can be found to use these new tests.  The potential range of use is
hinted at by the fact that exactly the same algorithms that work so
well for the analysis of human language also appear to work very well
for the analysis of genetic sequences.

The practical utility of these tests and the general methods for
deriving them are of great value.  At least as important, though, is
the fact that there is an extensive theoretical underpinning to these
tests which provides more than just a factory for more and more
analytic techniques.  This theoretical foundation also has deep and
fundamental parallels in a wide range of other theoretical areas.
These parallels provide, on the one hand, a philosophical framework
for log-likelihood ratio methods in terms of minimum description
length methods.  This mathematical connection was described in section
\ref{methods:mdl}.  This connection implies that log-likelihood ratio
methods are fundamentally in accord with the philosophy that a good
theory is a parsimonious and accurate description of the universe.  On
the other hand, these parallel areas are rich resources in terms of
potential applications for log-likelihood ratio methods.  Since the
log-likelihood ratio tends to be extremely inexpensive to compute,
many problems which are either expensive or even not feasible to solve
with other methods can be solved approximately using log-likelihood
ratio methods.

Apart from any practical considerations, it is also natural to wonder
what it really means that systems based on such different symbolic
sequences succumb so similarly to identical techniques.  It would be
presumptuous to claim that these commonalities arise from anything
deeper than independent evolution, but even that claim is provocative.
Darwinian evolution is by definition at work in the development of the
genetic code, and it has long been asserted that there are at least
analogies to natural selection and mutation which are at work in the
development of human language.  These analogous processes may lead to
the existence of shared properties of evolving informational systems.

Setting aside the biological basis for the cognitive prerequisites to
linguistic behavior and the resulting obvious connection with
biological evolution, language still moves.  Certainly it is true that
there is an advantage to linguistic forms which convey information
concisely, and certainly it is true that new expressions appear.
Variants form from these new expressions and these expressions, new,
old and variant, ``propagate'' to other speakers.  Whether these
analogies between the evolution of genetic sequences and natural
language are strong enough to result in similar evolutionary process
is not, however, a question that can be answered given the current
state of knowledge about how evolution works.

To study evolution and language, we do not, however, need to follow
the entire chain of reasoning from theory to practice.  The results
presented here provide evidence from the practical end of the chain.
Instead of starting with the cause, we can claim that there is
substantial similarity in the effect.  Whatever the mechanisms,
whatever the similarities of process, we can now definitively say that
there are interesting and deep parallels between the structure of
genetic sequences and human languages.

\chapter{Glossary}

The terms in this glossary were selected automatically using a
log-likelihood ratio test.  Terms selected were those which appeared
significantly more often than in newswire text.  As such, this
glossary is itself a demonstration of the utility of the
log-likelihood ratio approach to text analysis.  This chapter is yet
another demonstration of the applicability of the log-likelihood ratio
test to practical problems.  Here, the log-likelihood ratio was used
to extract key terms from a large body of text.

This glossary was constructed by manually defining terms which
appeared in this dissertation with {\em significantly} higher
frequency than in a random selection of AP newswire from the year
1988.  Testing for a difference in frequency was done using a
log-likelihood ratio test in a manner similar to the test used in
chapter \ref{colloc}.  

\paragraph{$\alpha$} 
Used in this dissertation to indicate a heuristic constant whose value
cannot be entirely justified on theoretical grounds.

\paragraph{$\mu$}
Mostly used here to represent an estimated probability which is derived
by combining two samples; also used to represent an interpolation
constant or the arithmetic mean of a number of quantities.

\paragraph{$\Sigma$}
Generally used here to indicate the alphabet over which symbols may
range. 

\paragraph{$\sigma$}
Generally used to indicate a dummy variable of summation which is
taken from some alphabet $\Sigma$.

\paragraph{$\theta$}
Used in this dissertation to represent the parameters of a statistical
model.

\paragraph{$\chi^2$ distribution}
A well-studied distribution of random variables.  The sum of the
square of a random variable which has unit normal distribution and
zero mean is $\chi^2$ distributed with one degree of freedom.  The sum
of the squares of $n$ independent random variables each of which have
unit normal distribution and zero mean is $\chi^2$ distributed with $n$
degrees of freedom.

\paragraph{$\chi^2$ test}
Also known as Pearson's $\chi^2$ test.  This test is a method for
analysing a contingency table.  The result of a $\chi^2$ test is
$\chi^2$ distributed if a few basic assumptions are met, hence the
name.  This test is called Pearson's $\chi^2$ test in this work to
avoid confusion with other statistics which have the same asymptotic
distribution.

\paragraph{$\omega$, $\Omega$ and $\Omega_0$}
The symbol $\omega$ is used as a formal variable which is a member of
either $\Omega$ or $\Omega_0$.  $\Omega$ is the set of all legal
parameter values for a model, and $\Omega_0$ is the subset of $\Omega$
which is consistent with the null hypothesis.

\paragraph{algorithm}
A detailed and more or less formalized description of how to perform a
computational task.  An algorithm is generally somewhat mathematical
in nature.

\paragraph{alfalfa}
See {\em Alfalfa meliloti}.

\paragraph{{\em Alfalfa meliloti}}
A dicotyledonous flowering agricultural plant related to clo\-ver
and commonly called alfalfa.  {\em Alfalfa meliloti} is a
nitrogen-fixing plant that enriches soil.

\paragraph{alphabet} 
A set of symbols from which the individual elements of a symbolic
sequence are taken.  In this work, the alphabet is generally finite,
but generalization to countably infinite alphabets is not difficult.

\paragraph{{\em Arabidopsis thaliana}}
A dicotyledonous flowering plant often used in genomic research
because it is relatively easy to work with experimentally and because
it has a relatively small genome for a plant.

\paragraph{argmax}
A mathematical notation for the value of some formal variable at which
the maximal value of a function is found.  Specifically,
\begin{displaymath}
\argmax_x f(x)
\end{displaymath}
is a handy way to express the value of $x$ for which $f(x)$ is
maximized.  This is in contrast to $\max_x f(x)$ which would merely be
the maximum value of $f(x)$, not the value of $x$ where this maximum
was attained.

\paragraph{association}
The degree that two events tend to occur in concert more than would be
expected by the frequency at which they appear in isolation.

\paragraph{assumption}
Usually, a statistical assumption where it is postulated that the true
distribution of some phenomenon matches some distribution, or that
events are independent in some way.  The assumption that the axioms of
probability theory apply to the real world is also usually made
without comment.  Making certain assumptions about independence and
distribution allows mathematical inference to proceed, but it must be
recognized that the original assumptions are almost never strictly
true.  This means that statistical inferences are almost always an
approximation of the truth, albeit often a very good one.

\paragraph{asymptotic}
In the limit.  Generally, in statistical parlance, asymptotic refers
to a value in the limit of a large number of observations where the
central limit theorem applies.

\paragraph{BAC}
Abbreviation for Bacterial Artificial Chromosome.  An experimental
tool by which the amount of some DNA sequence of interest is
amplified by introducing the DNA into a bacterial cell which is then
allowed to multiply.

\paragraph{backoff}
A class of mixed-order Markov model in which higher-order conditional
probabilities are used as long as they are justified by held out
training data.  Described in \cite{katz}.

\paragraph{Bayesian}
A school of thought (named for Thomas Bayes, an English cleric)
regarding the interpretation of statistical evidence as probabilities.

\paragraph{Bayes Theorem}
The theorem which relates joint and conditional probability,
$p(A,B) = p(A \mid B) \,\, p(B) = p(B \mid A) \,\, p(A)$.  By
definition, $A$ and $B$ are independent if $p(A) = p(A \mid B)$.

\paragraph{Bayesian Estimator}
An estimate of the parameters of a distribution which minimizes the
expected value of some loss function.  A Bayesian
estimator combines observed data with prior expectation about the
possible distribution of model parameters (the prior) to come up with
a refined distribution (the posterior) for the model parameters.  The
value which minimizes a loss function given this posterior
distribution is the Bayesian estimator.  See also MDL.

\paragraph{bigram}
A 2-gram.  See also $n$-mer and $n$-gram.

\paragraph{binomial}
A family of probability distributions for random variables which can
take on only two discrete values.

\paragraph{bit}
The natural unit of information or uncertainty.  If two events each
have probability 0.5, then the uncertainty involved is 1 bit.

\paragraph{bootstrap}
A method for estimating the potential variation in values estimated
from a sample of a random variable whose distribution is either not
known or which is not convenient.  The bootstrap involves sampling
with replacement from observed data to build synthetic datasets.

\paragraph{byte}
Generally, the smallest addressable unit in a computer.  Almost always
contains 8 bits in modern architectures.  A byte is roughly equivalent
to a single character for most European languages and roughly half a
character for most East Asian languages.

\paragraph{{\em Caenorhabditis elegans}}
A nematode (or roundworm) often used in genetic and developmental
research.  The name is generally abbreviated as {\em C. elegans},
although there is a plant species with the same abbreviated name.

\paragraph{chromosome}
In eukaryotic cells, a nuclear structure composed of DNA packaged with
proteins.  The genome of a eukaryote consists of multiple chromosomes
in the nucleus.  Eukaryotic sub-cellular organelles, such as
mitochondria and chloroplasts, also contain DNA that serves as genetic
material.  The major prokaryotic genetic material, generally a primary
loop of DNA, is also sometimes called a chromosome.  Prokaryotes may
also contain independently replicating extra-chromosomal DNA organized
as additional loops.  These additional loops are called plasmids.

\paragraph{classifier}
An algorithm which is used to determine some characteristic of a set
of observations.

\paragraph{collocation}
The use of two words directly adjacent to each other.  Contrast with
collocation.  Sometimes used as a synonym of collocation, although
that usage is avoided in this dissertation.

\paragraph{complexity}
Formally, the ability of a symbolic sequence to resist description in
a concise manner.

\paragraph{conditional}
A conditional probability is the probability that one event will occur
given that some other event happens.  It should be emphasized that a
conditional probability does not imply causality or temporal
sequence.  A large conditional probability does not even imply any
sort of correlation.  In contrast with the conditional probability,
the joint probability is the probability that two events will both
happen.  Bayes theorem relates conditional and joint probabilities.  

\paragraph{contingency}
A contingency table is a method for organizing observations into a
rectangular table so that the independent influence of different
factors can be analysed.

\paragraph{coocurrence}
The use of two words in proximity to each other.  Contrast with
coocurrence.  Sometimes used as a synonym for coocurrence, although
that usage is avoided here.

\paragraph{corpus}
A body of text.

\paragraph{correlation}
A lack of independence.  There is a correlation between two events if
they are not independent.  Independence of $A$ and $B$ is defined as
the situation where $p(A) \, p(B) = p(AB)$

\paragraph{count}
The number of times something occurs.  The number of times $x$ occurs
in $S$ is written here as $T(x, S)$.  Note that the term {\em
frequency} is used here only to denote a count which has been
normalized by the number of times $x$ could have occurred in $S$,
i.e. $T(x, S)/T(*, S)$.  The $T(*,S)$ notation is used to indicate the
number of times anything occurred in $S$.  This quantity is not
generally quite the same as the length of $S$.

\paragraph{data}
In this dissertation data are usually in the form of counts of the
number of times some pattern was noted.

\paragraph{Dice coefficient}
For two sets $D$ and $Q$, the Dice coefficient is defined as the size
of the intersection between the two sets divided by the sum of the
sizes of the two sets or ${2 | D \cap Q |}/{( |D| + |Q| )}$.

\paragraph{dicot or dicotyledonous}
A class of flowering plants in which the seed contains two seed
leaves.  See also monocotyledonous.

\paragraph{dimer}
A 2-mer.  See $n$-mer.

\paragraph{discrete}
A discrete random variable can only have values taken from a countable
set.

\paragraph{distribution}
A probability density function is a mathematical function which
formalizes the intuitive concept of probability.  A probability
density function is subject to several mathematical constraints
including $p(x) \ge 0$ and $\int p(x) \,\, dx = 1$.  Conceptually,
$\int_a^b p(x) \,\, dx = Pr(x \ge a \wedge x \le b)$.  A cumulative
density function can be defined when $x$ is taken from ${\mathbb R}$.
It is defined as $P(a) = Pr(x \le a) = \int_{-\infty}^a p(x) \,\,dx$.

\paragraph{document}
A convenient unit of text.  For the purposes of text retrieval, a
document may or may not actually correspond to what might normally be
called a document.  Generally, a document is the unit that is actually
retrieved.

\paragraph{document retrieval}
A specialized form of information retrieval concerned only with the
retrieval of textual documents.

\paragraph{document routing}
A form of document retrieval in which queries are kept constant and
documents are sequentially matched to these queries.  Generally,
substantial numbers of example documents which are known to be good
and bad exemplars for a query are given as training data.

\paragraph{{\em Drosophila melanogaster}}
A fruit fly.  Fruit flies are a traditional research tool in
genetics.

\paragraph{DNA}
Abbreviation for {\underline d}eoxyribo{\underline n}ucleic {\underline a}cid.  See also RNA.

\paragraph{encoding}
The representation of one symbolic sequence by another.  The trivial
encoding is one in which a sequence is represented as itself.  The
most compact encoding in terms of binary digits allows the definition
of the concept of complexity in terms of bits.

\paragraph{entropy}
A measure of disorder or lack of certainty.  Surprisingly, given a few
very general desiderata, there is only one possible mathematical form
for entropy.  Entropy in an information theoretic sense has close
parallels to the entropy used in statistical mechanics and
thermodynamics.


\paragraph{{\em Escherichia coli}}
An enterobacterium used ubiquitously in biological research which is
often referred to by the abbreviation {\em E. coli}.

\paragraph{estimate}
A numerical value based on observations which is the putative value
for a parameter of a statistical model.

\paragraph{estimator}
A function or algorithm which can be used to convert observations into
an estimate for some model parameter.

\paragraph{eukaryote}
An organism whose cell or cells have a nucleus.

\paragraph{exons}
The portion of genetic sequence which is retained after splicing of
mRNA.  For genes which code for proteins, exons are the portion of the
gene which are ultimately represented in the protein sequence.

\paragraph{exponential distribution}
An exponential distribution is the probability density function
defined by $p(x) = \lambda e^{-\lambda x}$.  The mean of the
exponential distribution is $1/\lambda$.

\paragraph{exponential model}
An exponential model defines the probability of some observation based
on feature functions.  The mathematical form of an exponential model
is 
\begin{displaymath}
p(x) = \frac 1 Z \exp \left(\sum_i \lambda_i f_i(x) \right)
\end{displaymath}
Here $Z$ is a normalization constant which makes this function into a
valid distribution, the $f_i$ are the feature functions, the
$\lambda_i$ are weights which set the relative importance of the
various feature functions as well as the sign of their contribution.
When the feature functions are binary functions, exponential models
can be fitted to observations without enormous computational effort.
Otherwise, building exponential models can be extremely expensive.

\paragraph{Fisher}
A prominent statistician from the first half of the 20th century.
Fisher is considered by biologists to be a prominent geneticist.

\paragraph{roundworm}
In this work, {\em Caenorhabditis elegans}.

\paragraph{flowering plant}
An angiosperm.  The only flowering plants considered in this work are
a small number monocotyledonous and dicotyledonous plants.

\paragraph{frequency}
A count which is normalized by the number of samples.

\paragraph{$G^2$-statistic}
The log-likelihood ratio statistic which results from using a null
hypothesis that asserts that two or more multinomial distributions
are identical.

\paragraph{{\em Gallus domesticus}}
A domestic chicken.

\paragraph{generalized likelihood ratio}
A ratio of maximum likelihoods.
See section \ref{methods:lr_def} for detailed definition.

\paragraph{genetic}
Traditionally taken as ``Of or having to do with genes''.  Now used
more generally as pertaining to the behavior of heritable information
stored as RNA and DNA in a cell.

\paragraph{genome}
The totality of the DNA in a cell.

\paragraph{$n$-gram} 
A short sequence of symbols.  See also $n$-mer and tuple.

\paragraph{heuristic}
Essentially, a rule of thumb.  An heuristic formulation might be one
which is inspired by rigorous derivations, but which cannot itself be
derived rigorously.  Heuristic algorithms are algorithms which are not
guaranteed to succeed but which make use of an internal assumption
about how to proceed and which may give good performance in many
practical situations.

\paragraph{hidden}
Not observable.  A hidden Markov model is a Markov model where we
cannot observe the state sequence of the model but must instead infer
both the structure of the model as well as the probable sequence of
states through which the model is likely to have gone.

\paragraph{hypothesis}
Some restriction on the values of the parameters of a model.  One
hypothesis might be that two events are independent.  If the model
under consideration is defined in terms of conditional probabilities,
then this hypothesis is the same as constraining the conditional
probabilities so that $p(A) = p(A \mid B) = p(A \mid \neg B)$.

\paragraph{information}
The amount of decrease in entropy.

\paragraph{information retrieval}
This is the field concerned with the retrieval of information stored
in a repository.  It is generally assumed that this information is not
in a form suitable for retrieval by traditional databases.  Examples
of information not suitable for the application of traditional
database techniques which have been the focus of research in the
information retrieval community include the retrieval free-form text
or images.  Often, the general term information retrieval is used to
denote the specific case of document retrieval.

\paragraph{internationalization}
The process of making software usable in multiple human languages.
This is in contrast to localization which is the process of making
software usable in a single new language different from the language
for which the software was originally intended.

\paragraph{intron}
Originally short for intervening sequence.  The genetic sequence which
is represented in immature message RNA but which is removed during
splicing.  Introns are found only very rarely in prokaryotes, but are
common in many eukaryotes.  The origin and function of introns is not
known.

\paragraph{IR}
An abbreviation for Information Retrieval.

\paragraph{Jaccard coefficient}
For two sets $D$ and $Q$, the Jaccard coefficient is defined as the
quantity ${| D \cap Q |}/{| D \cup Q |} $.  See also Dice coefficient.

\paragraph{Lagrangian multiplier}
The use of Lagrangian multipliers is a general technique for
converting a constrained optimization problem into an unconstrained
optimization by the introduction of new independent variables.
Specifically, to maximize $f(x)$ subject to the constraint that $G(x)
= 0$, we can instead maximize $R(x, \lambda) = f(x) + \lambda G(x)$ in
terms of both $x$ and the new Lagrangian multiplier $\lambda$.  The
maximum value of $R(x, \lambda)$ must occur where $G(x) = 0$ and thus
the value of $x$ which maximizes $R(x, \lambda)$ will also maximize
$f(x)$ without violating the original constraint.

\paragraph{language}
A natural language is a language spoken or written by humans.  A
formal language is a definition of a set of symbol sequences.  This
definition is in terms of an alphabet and a set of rules for combining
symbols into sequences.  A probabilistic language is a formal language
which also has a method for assigning probabilities for all possible
symbol sequences.  Information and entropy are defined for
probabilistic languages.  Metaphorically, we can refer to the
``language of the genome'' based on the way that DNA encodes
information about the real world.

\paragraph{likelihood}
A particular value of the probability density function for a
statistical model.  The likelihood of an event is distinguished from
the probability of that event in that the likelihood is an {\em
estimate} of the probability.

\paragraph{likelihood ratio}
The ratio between likelihoods for the same event.  Comparing
likelihoods for observed events for various values of model parameters
allows a variety of conclusions to be drawn.  This characteristic is
what makes likelihood ratios into powerful tools for analysing
observations. 

\paragraph{linguistic}
Of or having to do with languages.  Can also be used to mean of or
having to do with linguists.

\paragraph{log-likelihood}
The logarithm of a likelihood function.

\paragraph{LSI}
Latent Semantic Indexing.  An information retrieval technique
developed at Bellcore which uses singular value decomposition.

\paragraph{Luduan}
A particular document routing system which is described in chapter
\ref{ir}. 

\paragraph{Markov}
A Russian mathematician from the early 20th century.  Markov's
contributions to the theories of probability and computation were
substantial.

\paragraph{Markov model}
A statistical model for describing symbolic sequences.
See section \ref{methods:mm}.

\paragraph{maximum}
The largest value.  The maximum of a function over some domain is the
largest value taken on by that function for any argument in the
domain. 

\paragraph{MDL}
An abbreviation for Minimum Description Length.  Using MDL techniques,
parameters are estimated by taking the values which maximize posterior
likelihood.  This is in contrast with maximum likelihood methods and
with most Bayesian estimators.  MDL methods share some of the
desirable invariance properties of maximum likelihood methods, but
they have much of the additional robustness of Bayesian methods.

\paragraph{$n$-mers} 
A short sequence of chemical units, especially nucleotides or amino
acids.  See also $n$-gram and polymer.

\paragraph{mitochondria}
Sub-cellular organelles which may have once been symbiotes but which
are not now independently viable.  Mitochondria provide the key
machinery for aerobic metabolism.  Mitochondria are found only in
eukaryotic cells.  Plants also have other similar sub-cellular
organelles such as chloroplasts, which perform photosynthesis.

\paragraph{model}
A function of data and parameter.  A model is a conditional
probability $p(x | \theta)$ of data $x$ given parameters $\theta$.

\paragraph{monocot or monocotyledonous}
A class of flowering plants in which the seed contains one cotyledon
(seed leaf).  See also dicotyledonous.

\paragraph{morphological analysis}
The analysis of words into their components such as the stem, prefixes
and suffixes.  Morphological analysis may also involve the reduction
of the stem to a canonical form called the lemma.  For information
retrieval purposes, morphological analysis typically is only concerned
with the removal of suffixes.

\paragraph{maximum likelihood}
The maximum of the likelihood function of some observed data where the
parameters of a probabilistic model are free to vary over some set
defined by any hypothesis we are examining.  We can contrast this with
maximum posterior likelihood which accounts not only for the variation
in likelihood of the observed data but also allows different values
of parameters to have different prior probabilities.  Minimum
Description Length methods are equivalent to maximum posterior
likelihood methods.

\paragraph{multinomial}
A distribution over a discrete and usually countable alphabet of
symbols in which each symbol in the alphabet has a fixed probability.
By definition, subsequent values of a multinomially distributed
random variable are independent.

\paragraph{mutual information}
A measure of the correlation between two sets of events.  Originally
framed in terms of a communications channel where it was interesting
to measure the amount of information that could be transferred through
the channel by measuring the probabilities that symbols input to the
channel would be output in some other form.  Mutual information is
desirable here since it allows the input symbols to be different from
the output symbols, which allows for the analysis of arbitrary coding
schemes.  Mutual information is defined as $MI(X,Y) = H(X,Y) - H(X) -
H(Y)$ where $H(X,Y)$, $H(X)$ and $H(Y)$ are the joint and
individual entropies of the random variables $X$ and $Y$.  Mutual
information is closely related to the log-likelihood ratio test.

Mutual information is often used in the computational linguistics
literature to refer to a slightly different quantity which is referred
to in this dissertation as single cell mutual information $SCMI(X,Y) =
\log {\frac {p(x,y)} { p(x) p(y)}}$.

\paragraph{{\em Nicotiana tabacum}}
The tobacco plant.

\paragraph{normal}
A normal distribution (sometimes called a Gaussian distribution) is
defined by a mean $\mu$ and variance $\sigma^2$.  The probability
density function for a normal distribution is given by 
\begin{displaymath}
p(x) = \frac 1 {\sigma \sqrt{2 \pi}} \, e^{-(x-\mu)^2/2\sigma^2}
\end{displaymath}

\paragraph{nucleic}
Of or having to do with nucleotides.

\paragraph{nucleotides}
The small repeating molecular components (called monomers) from which nucleic
acid polymers (RNA and DNA) are assembled.  In most naturally occurring
DNA, the nitrogenous base components of nucleotides in opposing
strands of DNA are paired.  For the most part, there are four kinds of
nucleotide bases occur in RNA or DNA.  The bases in the nucleotides in
DNA are adenine, cytosine, guanine and thymine.  In RNA, thymine is
replaced by uracil.

\paragraph{null hypothesis}
An hypothesis which is proposed largely as a mathematical straw man.
When looking for a correlation, the null hypothesis used is that there
is no correlation.  If the data contradict the null hypothesis, it can
be said that the data either supports or at least does not contradict
the opposite conclusion.  This inverted logic is used because most
statistical tests are only able to reject hypotheses definitively, not
substantiate them.

\paragraph{observations}
Observations are the result of doing some sort of experiment.  In this
thesis, the experiments described take on the form of counting the
number times some phenomenon takes place and thus the observations are
the collected counts.  The term ``observations'' is nearly synonymous
with ``sample'', but with ``observations'', there is a greater
connotation of real-world application as opposed to theoretical
rumination. 

\paragraph{OKAPI}
A research software system which does text retrieval.

\paragraph{order}
The order of a Markov model is the number of previous states on which
the probability of the next state depends.

\paragraph{organelles}
Membrane enclosed compartments within cells which provide isolation
for various purposes.  Mitochondria and chloroplasts are examples of
organelles.

\paragraph{organism}
An individual living system.  

\paragraph{{\em Oryza sativa}}
The species of rice most commonly found in genetic databases.  

\paragraph{over-fitting}
A situation which arises when a model has enough degrees of freedom so
that it can fit all of the peculiarities of the observed data but as a
consequence be unable to deal well with novel data.  It is desirable
to use a model with enough complexity so that it can model the true
structure found in the data, but not so much complexity that it begins
to model all of the adventitious statistical fluctuations in the
data.  The detection of over-fitting is generally done by holding some
data apart so that they can be used to test alternative models to see
if they generalize well to unseen data.

\paragraph{over-training}
See over-fitting.

\paragraph{parameter}
A value which determines the specific form of a probabilistic mo\-del.

\paragraph{polymers}
Chemical compounds which consist of a large number of repeated,
nearly identical units.  See $n$-mer.

\paragraph{priming}
A psychological phenomenon in which the time it takes for a subject to
recognize something decreases dramatically when preceded by some other
related item.  For instance, when given the task of distinguishing
words from non-words, a subject will tend to recognize the word DOCTOR
as a word and minor corruptions of DOCTOR as non-words more quickly if
primed by the appearance of NURSE in the previous trial.  The degree
of priming correlates very closely with subjective estimates of the
degree to which words are related as well as proximity in associative
recall experiments.  Many other forms of priming have been explored.

\paragraph{principle}
A guiding heuristic for the analysis of observed data and drawing
conclusions from that data about the parameters of a model.  One
commonly used principle in statistics is to choose model parameters
which maximize the probability of the observed data.  Another
principle used is to choose parameters which maximize the posteriori
probability of the observed data subject to a prior distribution of
parameters.  The justification of these principles is a difficult
philosophical undertaking which draws on many areas of modern
philosophy. 

\paragraph{probability}
The primitive concept in statistics and probability theory.
Informally, probability can be identified with the intuitive concept
of the likelihood that an event will occur.  In probability theory,
probability is equated to the measure of a set of events relative to
the measure of all possible events.

\paragraph{prokaryotes}
A major taxonomic group of organisms whose cells have no nucleus.  See
also eukaryotes.  Bacteria are prokaryotes.

\paragraph{protein}
A macromolecule (polymer) composed of a linear structure of amino
acids (the monomers).  Proteins have key structural and functional
roles in living systems.

\paragraph{query}
The restatement of a user's information need in a form usable by a
machine. 

\paragraph{random}
Lacking predictability.  Mathematically speaking, a random sequence of
symbols cannot be constructed using any computer program which is
substantially simpler than the original sequence.  Various degrees of
randomness are possible, and the discovery of simple programs which
generate partially random strings can be viewed as the fundamental
description of the process of discovery of the structure in such
strings.

\paragraph{rare}
Events with probability low enough so that the number of times the
events are observed is small.  Typically, small is interpreted
relative to the constraints of a particular statistical test such as
Pearson's $\chi^2$ test.

\paragraph{relevant}
A document is considered to be relevant if its content satisfies a
user's needs.  In information retrieval research, user needs are
necessarily replaced by codified statements of what a document must
contain in order to be relevant.  For simplicity, properties of a
document other than content are not considered in research on
information retrieval.  Related research areas such as
authorship identification or collaborative filtering consider
other properties of documents.

\paragraph{ribosome}
The sub-cellular molecular machinery which translates mature messenger
RNA into protein sequences.

\paragraph{RNA}
Abbreviation for {\underline r}ibo{\underline n}ucleic {\underline a}cid.  See also DNA.

\paragraph{{\em Saccharomyces cerevisiae}}
Brewer's yeast.  This simple eukaryote is often used in genetic
research since it provides a very simple experimental model for many
of the genetic mechanisms present in higher (eukaryotic) organisms.
Generally written as {\em S. cerevisiae}.  The name comes from the
Latin word for beer from which the modern Spanish word {\em cerveza}
is derived.

\paragraph{sample}
A sample is a subset of the theoretically complete set of all possible
events.  Often the word is used in a very empirical context, there is
often a strong connotation of taking a subset of a larger theoretical
universe.  See also ``observations''.

\paragraph{{\em Homo sapiens}}
Modern humans.

\paragraph{sequence}
A mapping from the positive integers to a countable set of symbols.  A
finite sequence is a mapping of positive integers less than some bound
to a countable set of symbols.  This bound is called the length of the
sequence.  Stated more informally, a sequence is like a set except that
duplicates are allowed and order matters.

\paragraph{significance}
Informally speaking, the significance of a set of observations is the
degree to which observations contradict a null hypothesis.  This
degree is generally measured by the probability that an observation
would have had if the null hypothesis were true.

\paragraph{SMART}
A research software program from Cornell University which does
document retrieval.

\paragraph{smoothing}
A statistic process which is effectively a form of induction.  The
term smoothing refers to the practice of combining the predictions of
simpler and more complex models so as to produce more detailed
predictions than are possible with the simpler model and to avoid some
of the problems of over-training or over-fitting inherent in the use of
more complex models.  The combination is hopefully done in such a way
that the predictions of the more complex model are emphasized when it
is able to produce valid predictions, but otherwise, the predictions
of the simpler model are used.

\paragraph{species}
A somewhat imprecise term which refers to populations of organisms
within which, under natural conditions, there is significant gene flow
and which are more or less genetically isolated from other
populations.

\paragraph{splice site}
Either end of an intron.

\paragraph{spliceosomes}
Enzyme-RNA complexes which facilitate splicing of introns.

\paragraph{splicing}
The process of removing introns from messenger RNA.

\paragraph{statistic}
A quantity which summarizes some aspect of an observation or set of
observations.  The average of a set of numerical observations is a
statistic, as are the largest and smallest of the observations.  Some
statistics have interesting mathematical properties which allow
conclusions to be drawn about the characteristics of the process being
observed.  These conclusions are subject to various assumptions.

\paragraph{statistical}
Having to do with statistics.  Often used, somewhat erroneously, to
imply a random or incomplete quality.

\paragraph{structure}
The degree to which a symbolic sequence is more ordered than random.
See random.

\paragraph{symbols}
A member of a countable set called an alphabet.  Often, the set to
which a symbol belongs is taken to be finite.  Symbols are often of
interest because they occur in ordered sequences (strings).  The term
symbol can have the connotation that there is a meaning associated
with each symbol, but this dissertation does not use the term
``symbol'' in this way.

\paragraph{test}
See training.

\paragraph{text retrieval}
A synonym for document retrieval.

\paragraph{tokens}
Units of text which are convenient for processing.  Tokens may or may
not correspond to words.  The term ``token'' is often used since it
has fewer connotations than the term ``word'' and allows a discussion
to be more abstract.

\paragraph{training}
Evaluations of supervised learning systems are normally done by
providing a set of known correct examples.  This set is divided into
two portions.  The first and generally larger portion is called the
training set and is presented to the system so that the
characteristics of the examples can be (hopefully) learned.  The
second portion is called the test set and is held out during the
learning process.  The system is evaluated by having examining its
responses on the test set and comparing to the correct answers. 

\paragraph{translation}
The process of reading mature messenger RNA and assembling the linear
protein structure encoded by the RNA.  Translation is done by
ribosomes with the assistance of transfer RNA.  This is in contrast to
the process of transcription which is the process of synthesizing
messenger RNA that corresponds to a DNA template.

\paragraph{TREC}
Text Retrieval and Evaluation Conference.  An annual conference whose
aim is to stimulate text retrieval research primarily by providing a
small set of standardized retrieval tasks on a database of realistic
size. 

\paragraph{trimer}
A 3-mer.  See $n$-mer.

\paragraph{tuples}
See $n$-gram.

\paragraph{unigram}
A 1-gram.  The trivial case of an $n$-gram which consists of a
single symbol.

\paragraph{variance}
The mean squared deviation from the mean.

\paragraph{vocabulary}
See alphabet.

\paragraph{weighting}
The numerical process of assigning more importance to some cases than
others. 

\paragraph{whitespace}
A character which, when printed or represented on a computer screen,
normally produces no visible mark.  The set of whitespace characters
includes the tab character, the space as well as characters which mark
the ends of lines or pages.  The \verb*| | symbol is often used to
represent whitespace in figures and diagrams.

\paragraph{window}
A short contiguous sequence of symbols taken from a much larger
sequence.  Coocurrence statistics for a particular word can be
gathered by counting the words which appear within windows centered
around the word of interest.

\paragraph{word}
A primitive unit of linguistic analysis which is unfortunately
difficult to define algorithmically. 

\paragraph{worm}
Here, an informal reference to the nematode species {\em
Caenorhabditis elegans}.  See the glossary entry for {\em
Caenorhabditis elegans} for more information.

\paragraph{YAC}
Abbreviation for Yeast Artificial Chromosome, an experimental
tool for introducing foreign DNA into yeast cells in order to
replicate this foreign DNA.  YAC's can be larger than BAC's, but yeast
cells have odd constraints about which DNA sequences they will carry
without modification.

\paragraph{yeast}
A kind of fungus.  In genetic research, the term yeast is generally
used to refer to {\em Saccharomyces cerevisiae} under which entry
more information can be found.

\bibliographystyle{alpha}
\bibliography{refs}

\end{document}